# AUTODISCOVER: A REINFORCEMENT LEARNING RECOMMENDATION SYSTEM FOR THE COLD-START IMBALANCE CHALLENGE IN ACTIVE LEARNING, POWERED BY GRAPH-AWARE THOMPSON SAMPLING

Author

Parsa Vares

Supervised by

Prof. Jun Pang, University of Luxembourg

Dr Éloi Durant, Luxembourg Institute of Science and Technology

Reviewed by

Dr Nicolas Médoc, Luxembourg Institute of Science and Technology

Examined by

Prof. Jun Pang, University of Luxembourg

Dr Éloi Durant, Luxembourg Institute of Science and Technology

Dr Nicolas Médoc, Luxembourg Institute of Science and Technology

# Declaration of authorship

**For single-authored work:**

UNIVERSITÉ DU LUXEMBOURG

*I hereby declare that I am the sole author of the work entitled.*

**Autodiscover: A Reinforcement Learning Recommendation System for The Cold-start Imbalance Challenge in Active Learning, Powered by Graph-aware Thompson Sampling**

*and here enclosed, and that I have compiled it in my own words, that I have not used any other than the cited sources and aids, and that all parts of this work, which I have adopted from other sources, are acknowledged and designated as such. I also confirm that this work has not been submitted previously or elsewhere.*

11.07.2025

Date

Signature

Parsa VARES

First name and surname

# Abstract

Systematic literature reviews (SLRs) are fundamental to evidence-based research, but the manual screening process presents a bottleneck and is increasingly difficult due to the growing volume of scientific literature. This process is challenged by the scarcity of relevant studies and the limited availability of expert screening decisions, where relevant documents are rare within a large corpus, and the high cost of expert screening and adjudication. Traditional active learning (AL) systems, used to mitigate this, often employ fixed (static) query strategies, i.e., a rule for selecting which unlabeled documents to present next for expert screening. Such fixed strategies fail to adapt to the evolving learning process and overlook the relational structure inherent in scientific literature networks.

This thesis introduces **AutoDiscover**, a novel framework that reframes active learning as an online decision-making problem driven by an adaptive agent. Scientific literature is modeled as a heterogeneous graph to capture the structural relationships among documents, authors, and metadata. A Heterogeneous graph Attention Network (HAN) learns node representations, which are used by a Discounted Thompson Sampling (DTS) agent to dynamically manage a portfolio of query strategies. By receiving real-time feedback from human-in-the-loop labels, the agent learns to balance exploration and exploitation decisions, adapting to the non-stationary dynamics of the review process, in which the utility of different strategies evolves over time.

Evaluated on the 26-dataset SYNERGY benchmark [13], AutoDiscover demonstrates higher screening efficiency compared to static active learning baselines. A key contribution is showing that the adaptive agent addresses the cold-start challenge, successfully bootstrapping discovery from minimal initial labels where static approaches fail. This work also introduces **TS-Insight**, an open-source visual analytics dashboard designed to interpret the agent's decisions, supporting verification, explanation, and diagnosis of algorithm behavior. Together, these contributions offer a method for accelerating systematic literature review under limited expert screening labels and low prevalence of relevant studies.

## Plain Language Summary

Consider the need to identify all relevant scientific studies related to a new disease for public health guidelines, or engineering papers in response to a climate-related event. Although this information exists, locating it resembles identifying relevant items within a large volume of unrelated data. For scientists and policymakers, manually searching through millions of research papers can take a team of experts months or even years, delaying life-saving discoveries and critical decisions.

To speed up this process, new computer-assisted methods are being explored. However, many of these emerging tools face a key challenge: they rely on a single, fixed screening prioritization strategy. This rigid approach struggles to adapt to the screening process. As reviewers label more papers, the system's understanding of relevance evolves, and the best screening prioritization rule can change over time. A fixed method, effective at finding papers very similar to the first few examples, may later fail to uncover important work from different scientific communities, representing isolated clusters of research in the literature network. This limitation means that crucial studies can be missed.

To address these challenges, this thesis introduces AutoDiscover, an adaptive system. The system moves beyond analyzing text in isolation by first building a graph-based representation of interconnections among research papers. This map shows which papers cite each other, which scientists work together, and where the research was done. By using this rich network of relationships, AutoDiscover gains a more comprehensive understanding of the research landscape, allowing it to search more flexibly and effectively.

The intelligent core of AutoDiscover manages a toolkit of diverse search strategies. Some strategies prioritize papers similar to previously labeled items, while others identify atypical or underrepresented content that may be relevant. The system learns during the review process by continually assessing which strategy is currently most effective. It incorporates a form of short-term memory, assigning greater weight to recent outcomes and deactivating strategies with reduced utility. This ensures the system doesn't get stuck, but rather adapts its search to what is working best at any given moment.

We evaluated AutoDiscover's performance against other methods using 26 past research projects where the final list of relevant studies was already known, providing a known ground truth for performance evaluation. The system's efficiency was measured by how early it identified known relevant papers during the review process. On average, AutoDiscover found important studies at a significantly higher rate than both a random search and other common tools that rely on a single, fixed strategy. By prioritizing relevant documents, the system reduces the screening effort required to identify pertinent literature. This reduces the manual workload and allows reviewers to allocate more time to reading and synthesizing evidence, supporting efficient and complete literature reviews.

# Acknowledgments

It is with immense gratitude that I acknowledge the individuals and institutions whose support and guidance were instrumental in the completion of this thesis. This journey, though challenging, was made possible by their unwavering encouragement and wisdom.

First and foremost, I wish to express my gratitude to my supervisors. To Dr. Éloi Durant, thank you for your practical wisdom, your constant encouragement, for pushing me to think critically and rigorously, and for fostering an environment where intellectual curiosity could flourish. To Prof. Jun Pang, thank you for your insightful academic guidance and support.

My thanks also go to my reviewer. I am deeply grateful to Dr. Nicolas Médoc for his thorough review. I would also like to thank Dr. Mohammad Ghoniem for his time and valuable mentorship.

I am indebted to the University of Luxembourg, particularly the Department of High-Performance Computing (HPC), and the Luxembourg Institute of Science and Technology (LIST) for providing me with the opportunity, resources, and the exceptional academic environment to pursue this Master's degree and conduct my research. My appreciation extends to the faculty, staff, and fellow students in the High Performance Computing program, whose friendship and stimulating discussions have enriched my experience.

On a more personal note, I owe everything to my family: to my mother Sara and my father Fereidoon, who gave me opportunities they never had and supported me unconditionally, every milestone I have reached is because of your sacrifices. I miss you both more than words can say. And to my little sister Dorsa, whose bright smile and boundless curiosity turn every challenge into an adventure and remind me daily of the wonder in the world.

To my beloved grandparents, Iran Samimi, Hossein Akbari, and Shirin Abiyaneh, whose wisdom, warmth, and endless love have guided me, and to my dear Babayi, who may no longer be with us but whose spirit and encouragement continue to light my path.

To my dear friends in Iran, Amir Shirmohammadi, Arshia Hajarian, Arian Alaei, Parham Samimi, Rana Asadolahi, Saman Majid, Mohammad Javad Vaziri, Ali Maleki, Mana Samimi, Mona Alaei, Paniz Samimi, and Sina Eftekhari, whose friendship and support have been a profound source of inspiration and hope throughout this journey.

To the friends I've made here, Thalia, Erfun, Oliver, Ilia, Zalan, Bowen, Mika, Moses, Michael, Elmarashly, Thea, Ciara, Charlotta, Rike, Sophie, Caia, Saad, Clementine, Angelica, Enora, Thu Huong, Ben, and Jader, thank you all for filling my days with laughter, support, and unforgettable memories.

**To everyone who has been a part of this journey, thank you.**

# Contents

# Abbreviations

| Abbreviation | Full Term |
|---|---|
| **AL** | Active Learning |
| **ALBL** | Active Learning by Learning |
| **ANRMAB** | Active learning for Network Representation with Multi-Armed Bandits |
| **API** | Application Programming Interface |
| **ASReview** | Active learning for Systematic literature Reviews |
| **ATD** | Average Time to Discovery |
| **BALD** | Bayesian Active Learning by Disagreement |
| **CSV** | Comma-Separated Values |
| **CUDA** | Compute Unified Device Architecture |
| **D2V** | Document to Vector (Doc2Vec) |
| **DOI** | Digital Object Identifier |
| **DRE** | Discovery Rate Efficiency |
| **DTS** | Discounted Thompson Sampling |
| **GAT** | Graph Attention Network |
| **GCN** | Graph Convolutional Network |
| **GNN** | Graph Neural Network |
| **GNN-TS** | Graph Neural Network - Thompson Sampling |
| **GPU** | Graphics Processing Unit |
| **HDR** | Highest Density Region |
| **HAN** | Heterogeneous graph Attention Network |
| **HPC** | High-Performance Computing |
| **HPO** | Hyperparameter Optimization |
| **I/O** | Input/Output |
| **LCB** | Least-Confidence-with-Bias |
| **LGE** | Label Propagation Graph Exploit |
| **LP** | Label Propagation |
| **LR** | Logistic Regression |
| **MAB** | Multi-Armed Bandit |
| **MAG** | Microsoft Academic Graph |
| **MC** | Monte Carlo |
| **MeSH** | Medical Subject Headings |
| **NB** | Naive Bayes |
| **NLM** | National Library of Medicine |
| **OOM** | Out-of-Memory |
| **PICO** | Patient/Problem, Intervention, Comparison, and Outcome |
| **PMID** | PubMed Identifier |
| **PyG** | PyTorch Geometric |
| **R@k** | Recall at k |
| **RAM** | Random Access Memory |
| **RF** | Random Forest |
| **RL** | Reinforcement Learning |
| **RRF@p** | Relevant References Found at p% |
| **SciBERT** | Scientific Bidirectional Encoder Representations from Transformers |
| **SLR** | Systematic Literature Review |
| **SPECTER2** | Scientific Paper Embeddings with Citation-Enhanced Representations, v2 |
| **SVM** | Support Vector Machine |
| **TF-IDF** | Term Frequency-Inverse Document Frequency |
| **TPE** | Tree-structured Parzen Estimator |
| **TS** | Thompson Sampling |
| **UMAP** | Uniform Manifold Approximation and Projection |
| **VRAM** | Video Random Access Memory |
| **WSS@p** | Work Saved over Sampling at p% |
| **XAI** | Explainable Artificial Intelligence |

# 1 Introduction

Systematic literature reviews (SLRs) are essential for producing state-of-the-art reports and ensuring reproducibility in fields such as healthcare, policy, and science [35]. These tasks involve large datasets where only a small fraction of documents are relevant, and screening all records manually is prohibitively time-consuming and costly. Furthermore, reviews often begin with only a few known relevant documents, making early decisions particularly challenging. While the literature itself exhibits rich relationships (e.g., citations, co-authorship) that can be modeled as a graph, these are often ignored by conventional active learning systems that treat documents as independent text entries [42].

Active Learning (AL) provides a general workflow for reducing annotation effort in systematic reviews by iteratively selecting and labeling the most informative documents [42]. Tools like ASReview [48] demonstrate that AL reduces screening effort by prioritizing documents strategically. However, our preliminary research, detailed in Section 5.1, reveals that standard AL architectures face a cascade of challenges:

1. **Prohibitive Computational Cost:** The use of state-of-the-art Large Language Models as core classifiers in AL systems is computationally infeasible due to the cost of iterative retraining (§5.1.1).
2. **Semantic Insufficiency (in our setting):** While TF-IDF can support meaningful similarity structure in many text-mining pipelines, in our preliminary experiments it captured surface lexical overlap rather than conceptual similarity, leading to weak neighborhood quality and, in some cases, poorly connected graphs (§5.1.2).
3. **Data Sparsity:** Graph-based methods relying on explicit metadata like citation networks are unreliable, as such data is often too sparse to form a connected graph (§5.1.3).
4. **Inherent Strategic Duality:** The very structure of citation data reveals that both exploitation (following dense clusters) and exploration (finding "island" papers) are necessary, a dilemma that a static query strategy cannot resolve (§5.1.4).
5. **Non-Linear Separability:** Even with powerful semantic embeddings, the classification problem is non-linear, for which linear classifiers are inadequate (§5.1.5).
6. **The Cold-Start Trap:** A powerful Graph Neural Network (GNN), when guided by a simple greedy strategy, fails to learn from a single seed document and gets stuck (§5.1.6).
7. **Non-Stationarity of the Optimal Strategy:** No single, fixed query strategy is consistently optimal throughout the review process (§5.1.7).

Given these challenges, which render current automated tools unreliable or inflexible, many reviewers still revert to a fully manual, "by hand, one by one" screening approach. This underscores the urgent need for a more adaptive and reliable system.

These findings motivate our primary research goal:

*To develop a dynamic active learning system that is not only highly efficient at accelerating systematic literature reviews, but also transparent and verifiable from a model development perspective.*

To achieve this goal, our research is guided by the following central hypotheses, which form the foundation of the AutoDiscover framework developed in this thesis:

**Hypothesis 1 (Graph-Based Representation and Learning):** Modeling scientific literature as a heterogeneous graph and using a Heterogeneous graph Attention Network (HAN) to learn from it will be more effective than traditional methods. We argue this is for two primary reasons: (1) the graph structure captures non-linear relationships that text-based models miss [26, 56], and (2) the HAN's adaptive attention mechanism can dynamically learn the importance of different relationship types, making it robust to the sparse and varied metadata in real-world datasets [52].

**Hypothesis 2 (Adaptive Strategy Selection):** No single query strategy is optimal throughout a systematic review due to the non-stationary nature of the process. We hypothesize that an agent that dynamically selects from a diverse portfolio of strategies—balancing exploitation of currently effective strategies with exploration of alternatives—will accommodate variability across reviews. A single fixed strategy is unlikely to be uniformly effective. Furthermore, because the optimal strategy evolves as more data are labeled, an agent that prioritizes recent successes (using a "forgetting" mechanism such as Discounted Thompson Sampling) will improve adaptation to these changes.

**Hypothesis 3 (Trust through Visualization):** The probabilistic behavior of a strategy selection agent can be made transparent and verifiable, thereby enhancing trust [9, 31]. We hypothesize that a visual analytics tool can enhance trust by enabling verification, explanation, and diagnosis of the agent's behavior, specifically verifying its mechanics, explaining individual decisions (exploration vs. exploitation), and assessing decision reliability.

This thesis introduces and evaluates AutoDiscover, a framework designed to test these hypotheses. AutoDiscover first models the scientific literature as a heterogeneous graph, leveraging four distinct types of nodes: **papers**, **authors**, **institutions**, and **Medical Subject Headings (MeSH)**. These nodes are connected by six types of meaningful relationships: papers are linked by direct *citations* and high *semantic similarity*; papers are connected to their *authors* and assigned *MeSH terms*; and authors are linked to each other by *co-authorship* and to their respective *institutions.* A Heterogeneous graph Attention Network (HAN) is then used to learn which of these relationships are most useful for the classification task in any given scenario, enabling dynamic weighting of heterogeneous metadata sources to inform classification.
The core innovation is an adaptive agent, driven by Discounted Thompson Sampling (DTS), that manages a diverse portfolio of nine query strategies. This portfolio was designed to include strategies based on exploitation, exploration, uncertainty, diversity, and graph structure, enabling the DTS agent to select the most effective strategy at each iteration.
In contrast to static systems requiring per-dataset hyperparameter tuning to identify a single effective strategy, AutoDiscover's agent learns an effective querying policy *online* during the review itself. We validate this framework on the 26 diverse datasets of the SYNERGY benchmark, measuring performance with metrics such as Discovery Rate Efficiency (DRE) and Work Saved over Sampling (WSS). This architectural advantage allows the system to achieve robust performance across a wide spectrum of review scenarios, each with different topics, class imbalances, and metadata quality, thereby reducing the risk of being over-specialized to any single configuration.
To test this final hypothesis, we developed **TS-Insight** [49], the visual analytics dashboard introduced as a contribution of this work. TS-Insight is designed to answer XAI questions posed by such systems. As described by Vares et al. [49], the relevant questions are: *"Is the algorithm working correctly?" (verification), "Why did the algorithm make this specific choice?" (explanation), and "When was the algorithm's choice outside the certainty region?" (reliability).* We assess how this tool provides interpretability to address these questions in the detailed case study presented in Section 7.5.
These questions align with the trust levels (TLs) and categories in Chatzimparmpas et al.'s taxonomy [9]: *verification* targets TL3–TL4 (learning method/algorithms; concrete models) via the *debugging/diagnosis* category; *explanation* targets TL3–TL4 via *understanding/explanation*; and *reliability* targets TL2 and TL5 (processed data; evaluation/user expectations) via *uncertainty awareness* and *results/metrics validation.* In this sense, verification corresponds to mechanistic/causal checking, explanation is explicitly contrastive (why this arm rather than others), and reliability foregrounds abnormal/low-probability events [31].

**Contributions** The primary contributions of this thesis are:

- The design and implementation of **AutoDiscover**, a graph-aware active learning framework that integrates a **Heterogeneous graph Attention Network (HAN)** for modeling the multi-relational structure of scholarly data (including citations, co-authorships, and semantic links) with a **Discounted Thompson Sampling (DTS) agent** (Section 5).

- A comprehensive empirical evaluation on the 26-dataset **Synergy** benchmark [13], demonstrating that AutoDiscover achieves improvements in screening efficiency over non-adaptive baselines (Section 7).

- The development and public release of **TS-Insight** [49], an open-source visual analytics dashboard for verification and explainable AI (XAI), with precise targets: verification of posterior updates $(\alpha, \beta)$, discounting, and arm-selection logic (exploration vs. exploitation); step-level explanations and reliability assessment. It supports diagnosis and step-by-step inspection of exploration/exploitation dynamics in Thompson Sampling variants, facilitating debugging and deployment in sensitive decision-making scenarios (Section 7.5).
  `(github.com/parsavares/ts-insight )(github.com/LIST-LUXEMBOURG/ts-insight)`

## 2 Foundational Concepts

This section provides a concise reference for the key technical and domain-specific concepts central to this thesis. These terms are foundational to understanding the problem, the proposed methodology, and the results discussed in the subsequent chapters.

Table 1: Key Terminology and Foundational Concepts

| Term | Definition |
|---|---|
| ***A. Systematic literature Reviews and Information Science Domain*** | |
| **Class Imbalance** | A dataset characteristic in which class labels are distributed unequally. In systematic literature reviews, this is often extreme, with relevant documents forming a very small minority. |
| **Cold-Start Problem** | The challenge of making effective classifications when the system has little to no initial labeled data to learn from, representing an extreme case of label scarcity. |
| **MeSH** | **M**edical **S**ubject **H**eadings. A comprehensive controlled vocabulary maintained by the U.S. National Library of Medicine (NLM), used to index and retrieve biomedical literature. |
| ***B. Graph Theory and Graph Neural Networks*** | |
| **Heterogeneous Graph** | A graph composed of multiple node types (e.g., 'paper', 'author') and edge types (e.g., 'cites', 'written_by'). |
| **Message Passing** | The core operational principle of most Graph Neural Networks (GNNs), where nodes iteratively aggregate feature information (messages) from their neighbors to update their own representations. |
| **Label Propagation** | A semi-supervised algorithm for graphs that infers labels for unlabeled nodes by propagating information from a limited set of labeled nodes via graph connections. |
| ***C. Active Learning and Bandit Theory*** | |
| **Stationary Environment** | An environment in which the underlying rules or reward distributions remain constant over time. |
| **Non-stationary Environment** | An environment in which the underlying rules or reward distributions evolve over time. The active learning process is non-stationary because the utility of different query strategies (i.e., acquisition rules used to select the next items for labeling, such as uncertainty or diversity sampling) evolves as the model learns. |
| **Oracle** | In active learning, the source of ground-truth labels. This is typically a human expert who provides the correct classification for a queried data point. |
| **Online Learning** | A machine learning paradigm in which data points arrive in a sequential manner, and the model updates its parameters incrementally with each new example. This is essential for dynamic environments, such as active learning in systematic literature reviews, where labeled data accumulates over time and retraining the model from scratch at every step is impractical. |
| **Discount Factor ($\gamma$)** | A parameter in the range (0, 1] used in Discounted Thompson Sampling to control the agent's memory. It systematically reduces the weight of past observations, forcing the agent to prioritize recent feedback and adapt to a non-stationary environment. |

# 3 Background

This section outlines the foundational principles required to contextualize the proposed system. It introduces Active Learning as the core task, Graph Neural Networks as the modeling framework, and Thompson Sampling as the control mechanism.

## 3.1 Active Learning for Systematic Literature Reviews

The screening phase is central to a Systematic Literature Review (SLR). In this stage, researchers examine titles and abstracts to identify studies meeting predefined inclusion criteria [35]. This process is resource-intensive and susceptible to human error and fatigue [34]. Technology-assisted screening aims to reduce manual effort while maintaining review rigor and completeness.

Active Learning (AL) is a machine learning paradigm designed to maximize model performance while minimizing the amount of labeled data required [42]. It does so by allowing a model to iteratively query an oracle, typically a human expert, for labels. In the context of SLRs, AL seeks to reduce expert annotation workload while preserving classification effectiveness.

The AL process follows an iterative cycle:

1. **Train Model:** A classifier is trained on an initial set of labeled instances.
2. **Query Strategy (Acquisition Rule):** The model evaluates the unlabeled pool, and an acquisition rule selects the next instance(s) for labeling.
3. **Oracle Annotation:** A human expert provides labels for the selected instance(s).
4. **Update and Retrain:** The labeled data are added to the training set, and the model is retrained.

This cycle repeats until a stopping criterion is met, such as reaching a performance threshold or exhausting the labeling budget[1] or achieving a target level of model performance. The performance of an AL system depends on its query strategy. Common strategies include **uncertainty sampling**, which selects instances with the lowest model confidence [27], and **diversity sampling**, which prioritizes instances dissimilar to previously labeled examples to enhance coverage of the feature space [6].

## 3.2 Graph Neural Networks for Literature Analysis

Scientific literature does not exist in isolation; it forms a complex network of relationships. Papers share common concepts, and cite other papers, are written by authors, who are affiliated with institutions. This rich structure is naturally represented as a **heterogeneous graph**, which contains multiple types of nodes (e.g., `paper`, `author`) and multiple types of edges (e.g., `cites`, `written_by`)[2].

**Graph Neural Networks (GNNs)** are a class of deep learning models designed to perform inference on graph-structured data [56]. They operate on the principle of **message passing**, where information flows between connected entities in the graph. In this process, each node iteratively updates its feature representation by first **aggregating** feature vectors from its immediate neighbors, and then **combining** this aggregated information with its own vector to compute an updated representation for the next layer. By repeating this process across multiple layers, a GNN learns rich representations (embeddings) that capture not only the intrinsic features of each entity but also its contextual role within the graph's overall structure.

While Graph Neural Networks like the Graph Convolutional Network (GCN) [26] are effective on simple graphs, they are limited because they treat all neighboring nodes with equal importance. A more sophisticated approach is the Graph Attention Network (GAT) [51], which introduces a self-attention mechanism. This mechanism allows the model to learn to assign different weights, or levels of "attention[3]", to different neighbors during message passing, giving more influence to more relevant nodes. However, the standard GAT architecture is designed for homogeneous graphs and does not differentiate between varying relationship types.

To address this, AutoDiscover employs a Heterogeneous graph Attention Network (HAN) [52]. This architecture extends the GAT concept by learning a separate attention mechanism for each edge type, allowing it to differentiate the importance of a 'citation' link from that of a 'co-author' link, for example.

[1] The labeling budget refers to the total amount of resources (e.g., time, money, or number of queries) allocated for manual annotation by the expert oracle. An AL system aims to maximize performance within this fixed budget.

[2] Figure 8

[3] "Attention is all you need" [50].

This design is a direct response to real-world data quality challenges. As our data quality audit reveals (Figure 4), the completeness of relational attributes like citations varies dramatically across datasets. A HAN can adapt to this inconsistent data quality by learning to assign lower attention weights to signals from sparse or unreliable relationship types, while assigning higher weights to more complete and informative ones, thereby tailoring its logic to the specific data landscape of each review.

## 3.3 The Multi-Armed Bandit Problem

The selection of a query strategy in active learning is an instance of the classic reinforcement learning problem known as the **Multi-Armed Bandit (MAB)**. The MAB problem is a metaphor for a gambler facing a row of slot machines (one-armed bandits), each with a different, unknown probability of paying out a reward. The gambler's goal is to maximize their total reward over a sequence of plays.
This scenario presents the fundamental **exploration-exploitation dilemma**:

- **Exploitation:** Selecting the action believed to yield the highest immediate reward based on current knowledge.
- **Exploration:** Selecting a less-certain action to gather information that may lead to higher long-term rewards.

In the context of AutoDiscover, each "arm" is a different AL query strategy [58] (e.g., uncertainty sampling, diversity sampling) [4].

## 3.4 Thompson Sampling: A Bayesian Approach to Bandit Problems

**Thompson Sampling (TS)**, also known as posterior sampling, is a popular Bayesian strategy for solving the MAB problem [39, 46]. It is used in applications ranging from adaptive clinical trials and personalized recommendation systems to dynamic pricing and computational advertising [8, 41]. Instead of just tracking the average reward of each arm, TS maintains a full probability distribution (or *belief*) over its reward probability, aiming to select the most promising arm while injecting randomness to explore uncertain options and avoid getting stuck in local maxima[5].

### 3.4.1 Standard Thompson Sampling

For binary rewards (success/failure, or relevant/irrelevant in our case), the **Beta distribution** is the standard choice for modeling the reward probability, $\theta_k$, of each arm $k$. The Beta distribution is the *conjugate prior*[6] for the Bernoulli distribution that governs a binary outcome. This property is computationally convenient: if the prior belief about an arm's success rate is a Beta distribution, the updated belief after observing a success or failure (the posterior) is also a Beta distribution, simplifying the update step to incrementing counters.
The Beta distribution is defined by two positive shape parameters, $\alpha_k$ (successes) and $\beta_k$ (failures). These are interpreted as pseudo-counts of past outcomes that shape the distribution in two ways:

- **Mean (Belief):** The mean of the distribution, $\mu = \frac{\alpha_k}{\alpha_k+\beta_k}$, represents the agent's current best estimate of the arm's true success rate.
- **Variance (Uncertainty):** The sum of the parameters, $\alpha_k + \beta_k$, is inversely related to the variance. A low sum indicates high uncertainty (a wide distribution), while a high sum indicates high confidence (a narrow distribution).

Figure 1 illustrates how these parameters influence the agent's belief, showing different scenarios an arm might be in during the learning process.

[4] A "pull" of an arm corresponds to using that strategy to select a paper for labeling. A "reward" is given if the selected paper is relevant. The goal is to learn a policy for choosing arms that maximizes the discovery of relevant papers over time.

[5] A local maximum is a point where the outcome seems best within a small neighborhood, but there may be better options elsewhere. Algorithms that don't explore enough can get stuck in these suboptimal solutions.

[6] In Bayesian statistics, a prior distribution is **conjugate** to a likelihood function (the model of the data) if the resulting posterior distribution is in the same family as the prior. Here, 'Beta (prior) + Bernoulli data = new Beta (posterior)', which simplifies calculations.

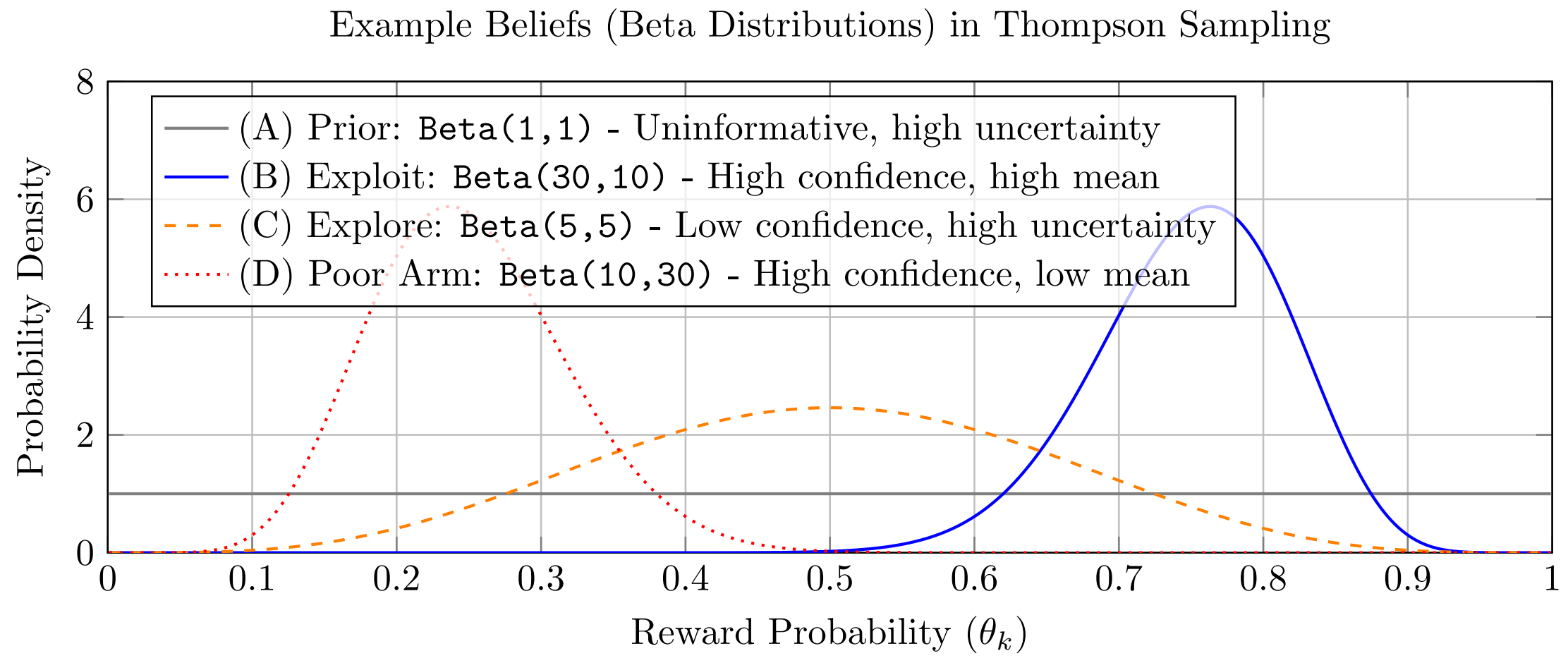

Figure 1: Example Beta distributions illustrating the agent's belief about an arm's reward probability. **(A)** The initial flat prior represents maximum uncertainty. **(B)** A highly successful arm develops a narrow distribution with a high mean, making it a prime candidate for exploitation. **(C)** An arm with few, mixed results has a wide distribution, indicating high uncertainty and making it a candidate for exploration. **(D)** An arm that consistently fails develops a narrow distribution with a low mean.

The TS algorithm proceeds as follows at each decision-making step $t$ of the active learning loop:

1. **Sample from Posteriors:**[7] For each arm $k$, a reward probability sample[8] $\hat{\theta}_{k,t}$ is drawn from its current Beta posterior distribution: $\hat{\theta}_{k,t} \sim \text{Beta}(\alpha_k, \beta_k)$.

2. **Select Arm with Highest Sampled Probability:** Choose the arm $a_t$ that yielded the highest sample: $a_t = \arg\max_k \hat{\theta}_{k,t}$.

3. **Observe Reward:** Apply the chosen strategy $a_t$ and observe the resulting binary reward $r_t \in \{0, 1\}$, where $r_t = 1$ corresponds to a relevant label from the oracle and $r_t = 0$ corresponds to an irrelevant one.

4. **Update Posterior:** Update the parameters of the chosen arm $a_t$. If the reward was a success ($r_t = 1$), increment its $\alpha$ parameter: $\alpha_{a_t} \leftarrow \alpha_{a_t} + 1$. If it was a failure ($r_t = 0$), increment its $\beta$ parameter: $\beta_{a_t} \leftarrow \beta_{a_t} + 1$.

Thompson Sampling provides a probabilistic mechanism to balance exploration and exploitation. An arm with a high mean but high uncertainty (a wide distribution, like Curve (C) in Figure 1) has a reasonable chance of producing a very high sample, leading to exploration. Conversely, an arm with a high mean and low uncertainty (a narrow, peaked distribution like Curve (B)) will reliably produce high-valued samples and thus be chosen frequently, leading to exploitation. This simple but powerful strategy is well-studied and has strong theoretical performance guarantees [39].

### 3.4.2 Discounted Thompson Sampling for Non-Stationary Environments

Standard TS assumes that the reward probabilities of the arms are *stationary*[9], that is, they do not change over time. While this assumption holds in many classic problems, it can be violated in dynamic scenarios where the probability of positive outcomes shifts. In active learning for systematic literature reviews, this violation is fundamental. The effectiveness of a query strategy (an arm) is inherently non-stationary because it changes as the underlying GNN model is retrained and more labels are acquired.

[7] A **posterior distribution** represents our updated belief about an unknown parameter (here, the arm's true success rate $\theta_k$) after we have observed new data (a reward). The update rule is conceptually 'Prior Belief + Data = Posterior Belief'. This posterior then serves as the prior for the next decision.

[8] The **sample** $\hat{\theta}_{k,t}$ is a single numeric value (e.g., 0.75) drawn from the arm's current Beta distribution. It represents one plausible guess for the arm's true, unknown reward probability, according to the agent's current belief.

[9] A **stationary** environment is one where the underlying rules or probabilities are constant. A slot machine with a fixed 20% payout rate is stationary. A **non-stationary** environment is one where these rules change over time, like an opponent in a game who learns and adapts their strategy.

For example, an uncertainty-based arm may be highly effective in the early **cold-start phase**[10], when the model has seen few examples and is highly uncertain. However, its utility may diminish later in the review when the model has become more confident and a purely exploitative strategy becomes more productive.

To address this, AutoDiscover uses **Discounted Thompson Sampling (DTS)** [36,37]. DTS adapts to non-stationary environments by introducing a **discount factor**, $\gamma \in (0, 1]$, which systematically de-weights historical evidence. Before the posterior of a chosen arm is updated with a new reward, its existing $\alpha$ and $\beta$ parameters are first decayed:

$$\alpha_k \leftarrow \gamma \cdot \alpha_k \tag{1}$$

$$\beta_k \leftarrow \gamma \cdot \beta_k \tag{2}$$

This multiplicative decay mechanism implements a form of exponential forgetting, giving greater weight to recent rewards. By preventing beliefs from becoming overly confident based on outdated information, the agent remains adaptive to changes in the utility of its query strategies throughout the review process. While alternative discounting formulations exist, such as the convex combination approach described by Russo et al. [39], we selected the direct multiplicative decay formulation (Eq. 1-2) for three reasons. First, its implementation is straightforward and directly models exponential memory decay. Second, it is a well-established method used in foundational DTS literature [20, 37]. Third, this choice aligns with the formal regret bounds presented in our theoretical grounding (Appendix C, Theorem C.1), directly linking our implementation to its theoretical justification.

## 3.5 Related Work

The work presented in this thesis synthesizes and extends research from three areas: active learning (AL) for systematic literature reviews (SLRs), graph-based representations of scholarly metadata, and adaptive agent-based systems for decision-making. This section reviews both foundational and recent contributions in these domains to establish the motivation for AutoDiscover's architecture. It begins with established AL pipelines, then discusses the evolution toward adaptive strategy selection and the integration of graph structures.

To accelerate the screening phase of SLRs, researchers have increasingly adopted active learning. Numerous tools have been developed, including **ASReview** [48], **SysRev** [5], and **Rayyan** [33], which provide standardized pipelines for screening large text corpora. Studies have shown that these AL systems can significantly reduce the screening workload compared to a fully manual review [34, 48].

These established systems rely on a *static* query strategy, such as uncertainty sampling. This strategy is paired with a fixed text representation model and a classifier. Common configurations include:

- **Text Representation:** Term Frequency-Inverse Document Frequency (TF-IDF), which represents documents as sparse vectors of word counts. This method can struggle with synonymy and the semantic nuances of scientific text.
- **Classifier:** Simple probabilistic models like Naive Bayes or linear models like Logistic Regression. These classifiers operate on the TF-IDF features.

This static approach has two primary limitations: the query strategy cannot adapt during the screening process, and the models do not inherently leverage the rich relational information available in scholarly metadata, such as citation networks. AutoDiscover is designed to address both of these shortcomings.

#### Adaptive Strategy Selection with Multi-Armed Bandits

The insight that no single query strategy is likely optimal across all datasets and all phases of a review has led researchers to explore adaptive methods using the Multi-Armed Bandit (MAB) framework. Hsu and Lin's **ALBL (Active Learning by Learning)** [23] framed AL strategy selection as a contextual bandit problem[11], representing an early and influential work in this area. More recently, the **TAILOR** framework [58] employed Thompson Sampling (TS) to manage a portfolio of AL algorithms, an agent-based approach that is conceptually aligned with the design of AutoDiscover.
A limitation of these systems is their assumption of a stationary environment, where the reward probability of each strategy is fixed. This assumption does not hold in AL for SLRs, as the utility of a query strategy evolves as the underlying model is retrained. To address this non-stationarity, AutoDiscover incorporates **Discounted Thompson Sampling (DTS)** [36, 37], a modification of TS designed for changing environments. By applying a discount factor, the DTS agent gives more weight to recent rewards, allowing it to adapt to the evolving utility of its arms. While DTS is an established technique, its application to manage a portfolio of query strategies within a graph-aware active learning loop for SLRs is a novel contribution of this work.

[10] The **cold-start phase** refers to the initial stage of an active learning process where the model has been trained on very few (or no) labeled examples. Its predictions are often unreliable, making it difficult to "bootstrap" the learning process without effective exploration.

[11] A **contextual bandit** is an extension of the simple multi-armed bandit problem. At each step, before choosing an arm, the agent observes a "context" vector of features that provides side information about the arms or the environment. The agent's goal is to learn a policy that maps contexts to the optimal arm choice. In GNN-TS, the GNN-generated node embedding is the context.

### Graph-Based Active Learning

To move beyond flat text representations, some researchers have modeled scientific literature as a graph, often using citation or co-authorship links to structure the data. Several frameworks have explored active learning on these graphs. For instance, **ActiveHNE** [11] proposes a method for cost-effective node labeling in heterogeneous networks by selecting nodes that are predicted to maximize network representation quality. Similarly, **ANRMAB** [16] uses a multi-armed bandit to select informative nodes for representation learning. More recently, reinforcement learning (RL) has been used to guide the AL process, as seen in **SMARTQUERY** [28], which trains an RL agent to formulate a dynamic querying policy. A potential limitation of such deep RL approaches is that they can be *sample-intensive*[12], often requiring many interactions to train the policy-formulating agent. In contrast, AutoDiscover's DTS agent addresses a simpler meta-problem of selecting from a fixed portfolio of strategies, making it more sample-efficient and better suited to the rapid feedback loop of an SLR [7].
The work most closely related to our framework is **Graph Neural Thompson Sampling (GNN-TS)** [55], which provided the direct inspiration for integrating GNNs with a Thompson Sampling agent. The GNN-TS framework conceptualizes active learning as a large-scale contextual bandit problem in which each node of the graph is treated as an individual arm. In this approach, a GNN learns a feature embedding for each node. These embeddings then provide the context for a linear bandit model that estimates a reward function over the entire set of nodes. At each query step, Thompson Sampling is employed on the posterior belief over this linear model's parameters to select the single node with the highest estimated reward.
While sharing the spirit of combining GNNs and TS, AutoDiscover differs in a crucial architectural choice that we hypothesize improves robustness, particularly in the cold-start phase. Instead of using the GNN to parameterize a reward model for every node, AutoDiscover uses a DTS agent to select from a *portfolio of diverse, pre-defined query strategies.* This portfolio includes GNN-dependent strategies (e.g., uncertainty sampling), model-agnostic strategies (e.g., diversity sampling), and methods based purely on graph structure (e.g., label propagation). This hybrid design ensures that even when the GNN is under-trained, the DTS agent can select other effective strategies, thereby leveraging both the power of the GNN once trained and the principles of established AL strategies when the GNN is still learning.

### Visualization and Explainability in Active Learning

As AL systems become more complex, understanding their internal decision-making processes becomes critical for debugging, tuning, building trust, and iterative development. Several platforms provide interactive dashboards to inspect model internals[13]. TensorBoard's Embedding Projector lets users explore learned feature spaces, and Google Vizier [19] offers a customizable UI for tracing training metrics. However, these general-purpose tools are not designed to analyze the specific probabilistic dynamics of Thompson Sampling based bandit agent within an AL loop. To our knowledge, there is no visualization dedicated to the explainability of TS-based algorithms.
To address this gap, our work contributes **TS-Insight** [49], an open-source dashboard tailored to Thompson Sampling agents. As detailed in Section 5.4, it provides targeted visualizations for the specific probabilistic dynamics of TS-based algorithms that general-purpose tools do not offer.

[12]**Sample-intensive** refers to learning algorithms that require a large amount of labeled data or environmental interactions to converge to a good solution. In the context of RL for AL, this could mean requiring many query-label-retrain cycles before the RL agent itself learns an effective querying policy.
[13]Model internals may include embedding trajectories, reward signals, uncertainty estimates, or selection histories relevant to system debugging.

# 4 Dataset Characterization & Data Analysis

Before detailing the architecture of AutoDiscover, it is essential to characterize the problem landscape it is designed to navigate. This section analyzes the benchmark datasets used for evaluation to demonstrate that they are representative of real-world systematic literature reviews, which are defined by challenges of scale, disciplinary diversity, extreme class imbalance, and imperfect metadata. As the overview in Table 2 illustrates, the variance in these datasets provides a rigorous testbed for the methods developed in this thesis.

All experiments in this thesis are conducted on the 26 datasets from the SYNERGY benchmark collection [13]. As summarized in Table 2, this benchmark is a strong proxy for real-world conditions as it comprises a diverse set of SLRs from multiple disciplines, with significant variance in both corpus size (from 258 to 38,114 records) and the prevalence of relevant documents (from 0.2% to 21.9%).

Table 2: Overview of the 26 SYNERGY Datasets Used for Evaluation. The table summarizes the characteristics of the benchmark, highlighting the wide variance in corpus size and prevalence of relevant documents.

| **Dataset Name** | **Total Records** | **Relevant** | **% Relevant** | **Primary Topic(s)** |
|---|---|---|---|---|
| Appenzeller-Herzog_2019 | 2873.0 | 26.0 | 0.9 | Medicine |
| Bos_2018 | 4878.0 | 10.0 | 0.2 | Medicine |
| Brouwer_2019 | 38,114.0 | 62.0 | 0.2 | Psychology, Medicine |
| Chou_2003 | 1908.0 | 15.0 | 0.8 | Medicine |
| Chou_2004 | 1630.0 | 9.0 | 0.6 | Medicine |
| Donners_2021 | 258.0 | 15.0 | 5.8 | Medicine |
| Hall_2012 | 8793.0 | 104.0 | 1.2 | Computer science |
| Jeyaraman_2020 | 1175.0 | 96.0 | 8.2 | Medicine |
| Leenaars_2019 | 5812.0 | 17.0 | 0.3 | Psychology, Chemistry, Medicine |
| Leenaars_2020 | 7216.0 | 583.0 | 8.1 | Medicine |
| Meijboom_2021 | 882.0 | 37.0 | 4.2 | Medicine |
| Menon_2022 | 975.0 | 74.0 | 7.6 | Medicine |
| Moran_2021 | 5214.0 | 111.0 | 2.1 | Biology, Medicine |
| Muthu_2021 | 2719.0 | 336.0 | 12.4 | Medicine |
| Nelson_2002 | 366.0 | 80.0 | 21.9 | Medicine |
| Oud_2018 | 952.0 | 20.0 | 2.1 | Psychology, Medicine |
| Radjenovic_2013 | 5935.0 | 48.0 | 0.8 | Computer science |
| Sep_2021 | 271.0 | 40.0 | 14.8 | Psychology |
| Smid_2020 | 2627.0 | 27.0 | 1.0 | Computer science, Mathematics |
| van_de_Schoot_2018 | 4544.0 | 38.0 | 0.8 | Psychology, Medicine |
| van_der_Valk_2021 | 725.0 | 89.0 | 12.3 | Medicine, Psychology |
| van_der_Waal_2022 | 1970.0 | 33.0 | 1.7 | Medicine |
| van_Dis_2020 | 9128.0 | 72.0 | 0.8 | Psychology, Medicine |
| Walker_2018 | 4837.0 | 76.0 | 1.6 | Biology, Medicine |
| Wassenaar_2017 | 7668.0 | 111.0 | 1.4 | Medicine, Biology, Chemistry |
| Wolters_2018 | 4280.0 | 19.0 | 0.4 | Medicine |
| **Mean** | 5334.0 | 87.0 | 5.3 | |
| **Median** | 3576.0 | 55.0 | 1.9 | |
| **Min** | 258.0 | 9.0 | 0.2 | |
| **Max** | 38,114.0 | 583.0 | 21.9 | |

### 4.0.1 Characteristics Analysis

A comprehensive analysis of the benchmark reveals several key characteristics that define the problem landscape. The datasets are not uniform; they exhibit significant diversity in size, class balance, and data distribution, providing a rigorous testbed for evaluating model robustness. Figure 2 visually summarizes these characteristics.

The benchmark presents several simultaneous challenges:

- **Scale Diversity:** As shown in Panel (A), the datasets range from small (Donners_2021, 258 records) to large (Brouwer_2019, 38,114 records), testing a model's ability to perform in both data-scarce and high-volume scenarios.
- **Extreme Class Imbalance:** Panel (B) illustrates the core challenge of SLRs. Across the aggregated corpus of over 169,000 documents, only 1.67% are labeled as relevant.
- **Data Distribution Skew:** The metadata features are also highly skewed. Panel (C) shows that the literature is concentrated in recent years, while Panel (D) reveals a long-tail distribution for citations, where most papers have few citations and a small number are highly cited.

An effective AL system must therefore be robust to these concurrent conditions.

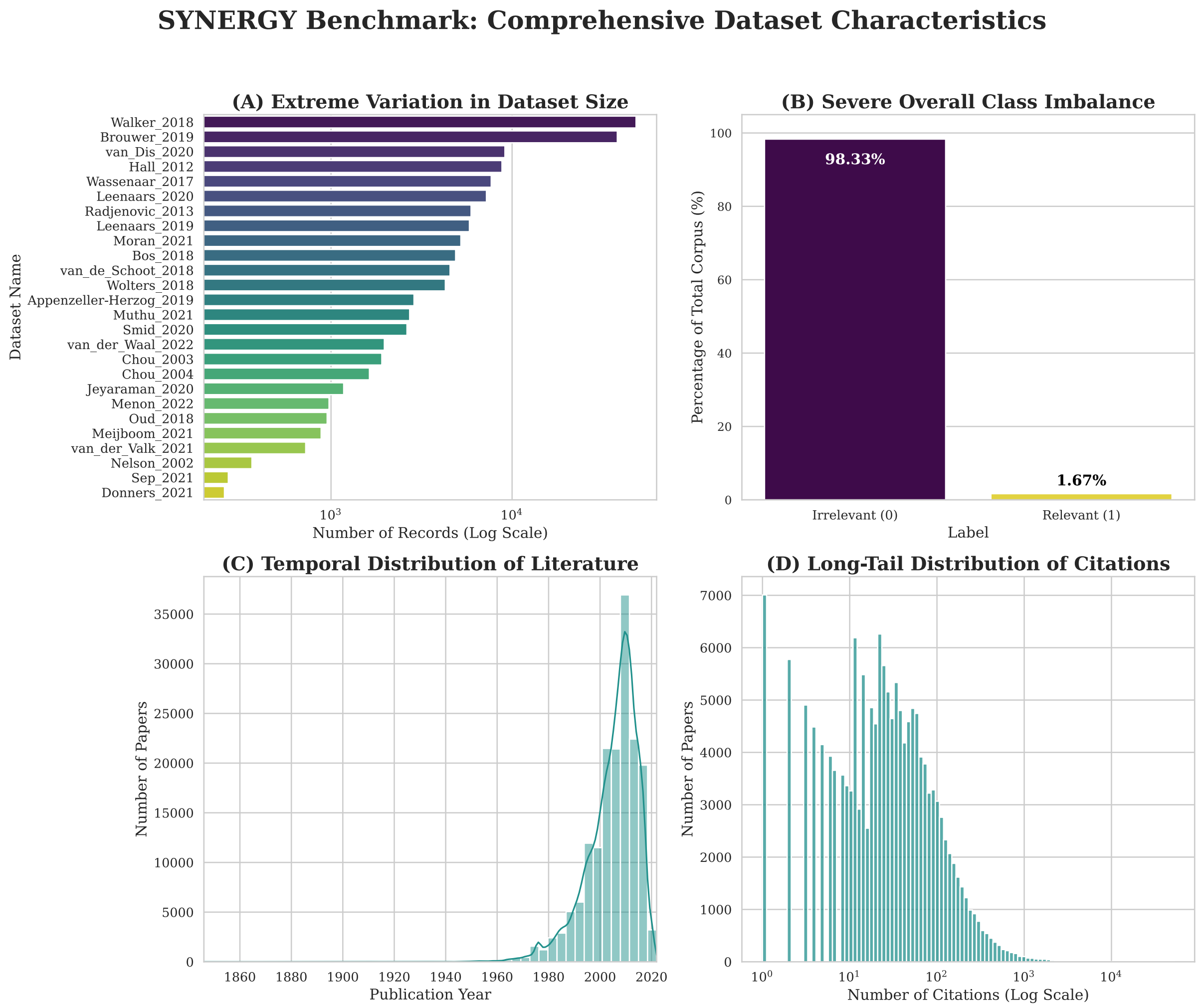

Figure 2: A comprehensive visualization of the SYNERGY benchmark characteristics, aggregating all 26 datasets. **(A)** shows the vast range in the number of records per dataset, plotted on a log scale. **(B)** illustrates the severe class imbalance across the entire corpus. **(C)** displays the temporal distribution of publication years. **(D)** reveals the long-tail distribution of citation counts on a log-log scale. These characteristics underscore the difficulty of the Systematic literature reviews task.

### 4.0.2 Distribution Analysis

The scale and diversity of the benchmark are further summarized by visualizing the distribution of key characteristics across the 26 datasets, as shown in Figure 3. To provide a complete picture of the data's properties, we present the distributions on both a linear scale (Figure 3a) and a logarithmic scale (Figure 3b).

The linear-scale view in Figure 3a highlights the data's extreme right-skew. The distributions for "Total Records" and "Relevant Records" (Panels A and B) are compressed near zero, with numerous outliers extending to the right. This visual compression reflects the underlying statistics: the mean number of total records (5,334) is substantially larger than the median (3,576), a classic indicator of a right-skewed distribution.

Replotting these metrics on a logarithmic scale (Figure 3b) resolves the structure within the compressed data. The log transformation makes the interquartile range clearly visible and the distributions more symmetric, allowing for a better inspection of the spread. The number of relevant documents per dataset spans nearly two orders of magnitude, from a minimum of 9 to a maximum of 583.

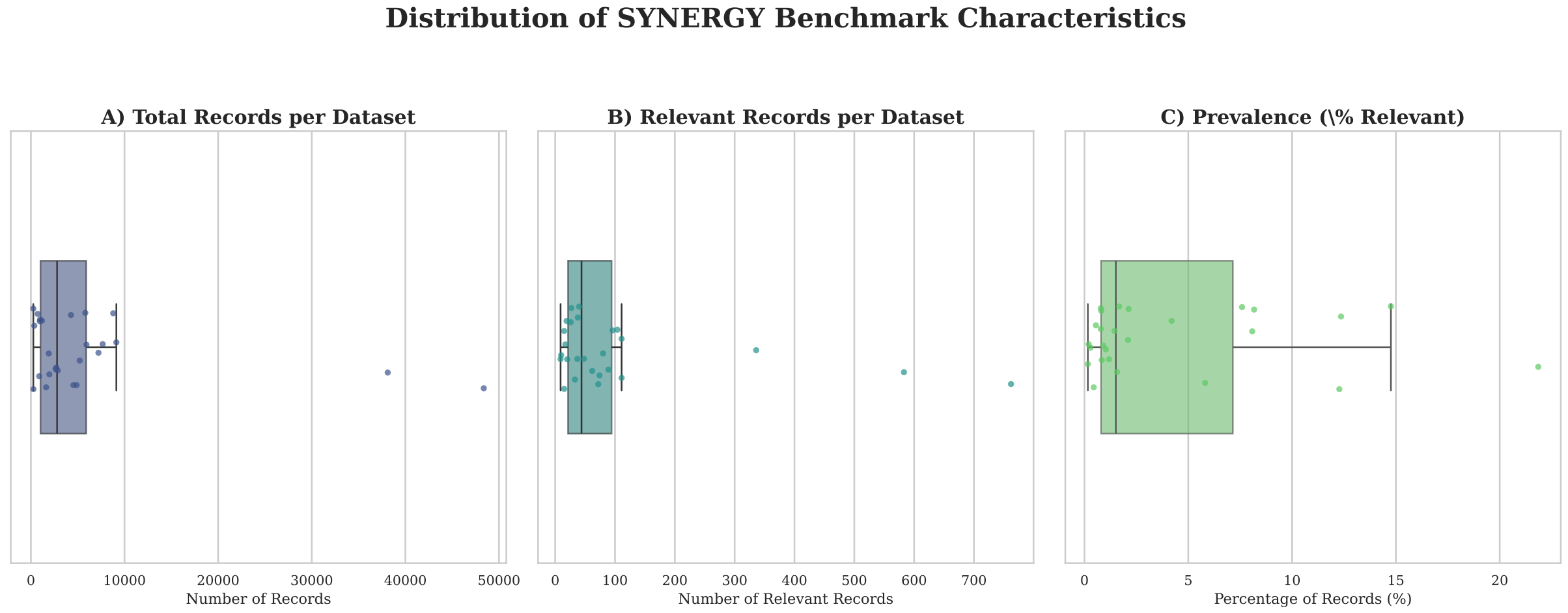

(a) Distribution on Linear Scale

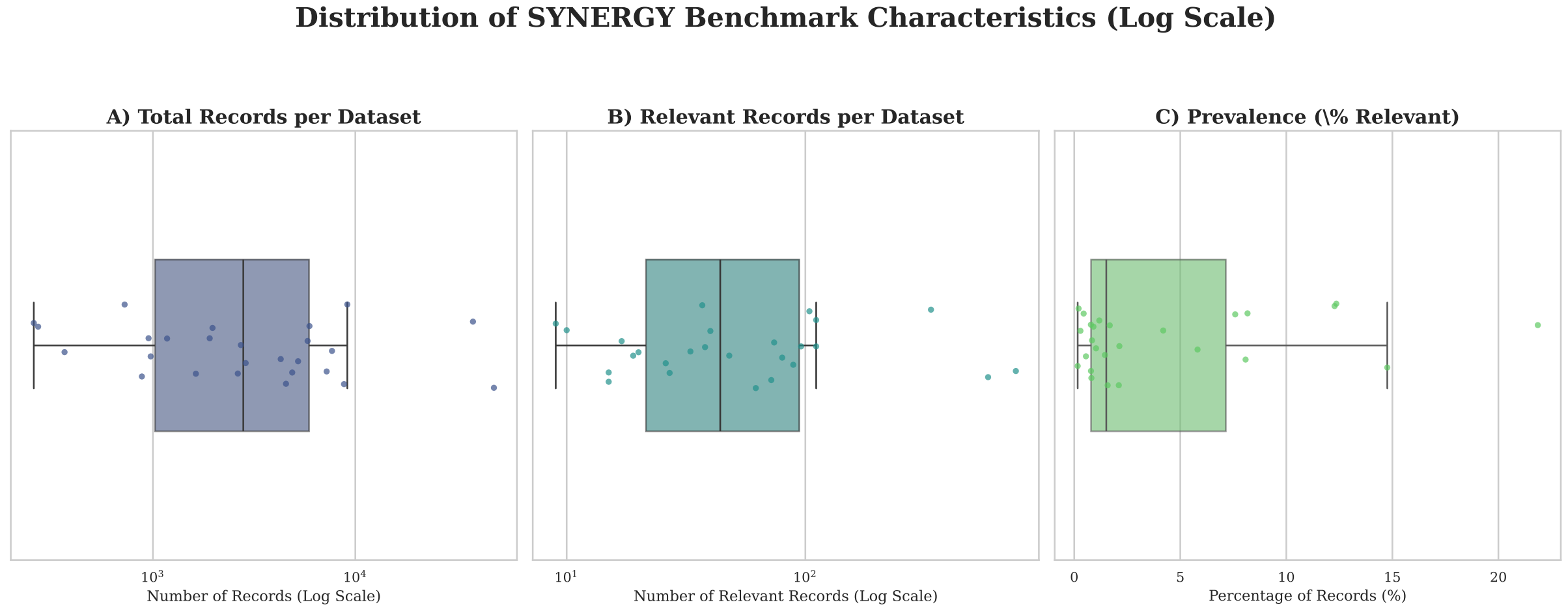

(b) Distribution on Logarithmic Scale

Figure 3: Comparison of dataset characteristics on linear and logarithmic scales. **(a)** The linear scale view highlights the extreme clustering of most datasets near zero, a classic sign of right-skewed data. **(b)** The logarithmic scale transforms the data to reveal the internal structure of the distributions and the full range of outliers.

The most critical insight comes from the distribution of prevalence (% Relevant) in Panel (C). On a linear scale, there is an extreme concentration of datasets at the low end of the spectrum: the median relevance is a mere 1.9%, and 75% of all datasets have a prevalence below 8% (Q3 = 7.98%). This graphically confirms that the "needle-in-a-haystack" problem is the default condition, not an exception, in this domain. This analysis illustrates the simultaneous challenges of scale, rarity, and heterogeneity that a robust discovery system must overcome.

#### 4.0.3 Data Quality Audit

A standard null-check analysis of data completeness can be misleading. Scholarly metadata, such as that in the SYNERGY benchmark, often contains entries that are not technically null but are **semantically empty**; for example, an empty list ('[]') for an author attribute or placeholder text for a title. We therefore performed a comprehensive data quality audit that accounts for both null values and these semantically empty entries.
Table 3 details the 33 available data attributes, highlighting those selected for graph construction, while Figure 4 visualizes the results of our audit.

Table 3: Complete List of Attributes Available in the Pre-Processed SYNERGY Benchmark Data. Rows highlighted in green represent the attributes selected during the data parsing stage for use in graph construction.

| # | `Attribute Name` | **Description** |
|---|---|---|
| **1** | `id` | Unique OpenAlex identifier for the work. |
| **2** | `doi` | Digital Object Identifier URL. |
| **3** | `title` | The title of the work (source for `title_clean`). |
| **4** | `display_name` | The standardized title for display purposes. |
| **5** | `publication_year` | The year the work was published. |
| **6** | `publication_date` | The full publication date (YYYY-MM-DD). |
| **7** | `ids` | A dictionary of other external identifiers (e.g., MAG, PMID). |
| **8** | `primary_location` | Information about the primary source/venue of the publication. |
| **9** | `host_venue` | Detailed information about the host journal or conference. |
| **10** | `type` | The type of the work (e.g., journal-article, book-chapter). |
| **11** | `open_access` | Details about the Open Access status of the work. |
| **12** | `authorships` | A list of author objects, including their affiliations. |
| **13** | `cited_by_count` | The total number of times the work has been cited. |
| **14** | `biblio` | Bibliographic information like volume, issue, and page numbers. |
| **15** | `is_retracted` | A boolean indicating if the work has been retracted. |
| **16** | `is_paratext` | A boolean indicating if the work is paratext. |
| **17** | `concepts` | A list of related concepts from the OpenAlex knowledge graph. |
| **18** | `mesh` | A list of Medical Subject Headings (MeSH) associated with the work. |
| **19** | `locations` | A list of all known locations where the work is hosted. |
| **20** | `best_oa_location` | Information about the best available Open Access location. |
| **21** | `alternate_host_venues` | A list of other venues where the work has appeared. |
| **22** | `referenced_works` | A list of OpenAlex IDs for works cited by this paper. |
| **23** | `related_works` | A list of OpenAlex IDs for works algorithmically related to this paper. |
| **24** | `ngrams_url` | A URL to the n-grams data for the work's full text. |
| **25** | `abstract_inverted_index` | An inverted index representation of the paper's abstract (source for `abstract_clean`). |
| **26** | `cited_by_api_url` | The API endpoint to retrieve works that cite this paper. |
| **27** | `counts_by_year` | A yearly breakdown of citation counts. |
| **28** | `updated_date` | The date the record was last updated in OpenAlex. |
| **29** | `created_date` | The date the record was created in OpenAlex. |
| **30** | `is_authors_truncated` | A boolean indicating if the author list is incomplete. |
| **31** | `label_included` | The ground-truth relevance label (1 for relevant, 0 for irrelevant). |
| **32** | `source_dataset` | The name of the original SYNERGY dataset from which the record originates. |
| **33** | `openalex_topics` | A list of topics assigned by the OpenAlex model. (Added via API call for analyses). |

This analysis reveals a more challenging data landscape than a simple null check would suggest. While some attributes are complete by design (e.g., `id`), many crucial relational attributes exhibit significant sparsity. For example, the heatmap shows that for some datasets, over 30% of entries for `referenced_works` are semantically empty lists, and the `doi` field has high rates of missingness. This audit establishes a critical requirement for AutoDiscover: the system must be robust to both overt and covert data sparsity, capable of leveraging whichever relational signals are present without failing when others are absent.

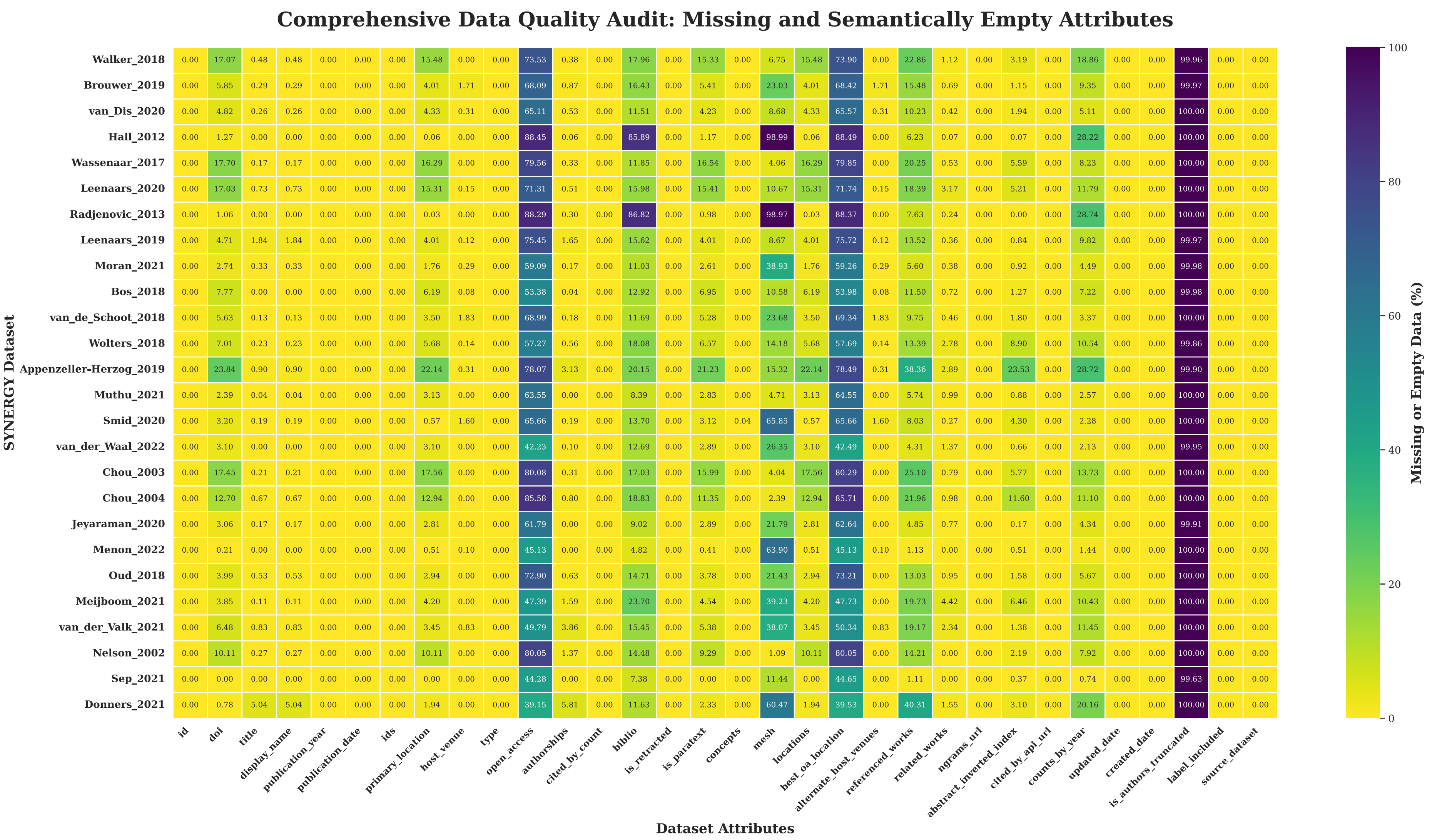

Figure 4: Comprehensive Data Quality Audit: A heatmap visualizing the percentage of missing or semantically empty entries for all available attributes across the 26 SYNERGY datasets. Darker colors indicate poorer data quality. This analysis goes beyond simple null checks to include empty lists and placeholder values, revealing the true extent of data sparsity that a robust system must handle.

# 5 Methodology

## 5.1 Design Rationale: Motivating Challenges

AutoDiscover is designed to support the core task of SLR screening: efficiently discovering the rare relevant studies in a large corpus under limited expert labeling time, while preserving transparency for model development and debugging. Building on the problem and data characterization in Chapters 1 and 4, This section details the design rationale by presenting an argumentative funnel that justifies each component of our final system.

Each challenge, labeled C1 through C7, directly motivates a key design decision, progressively building the case for the three core components of our final architecture: the use of pre-computed semantic embeddings, the necessity of a Graph Neural Network, and the critical role of an adaptive agent. A comprehensive summary of this logical progression is provided in Table 15 in the Appendix.

### 5.1.1 C1: Prohibitive Retraining Cost of Large Language Models

A common approach for text classification is to fine-tune a large language model (LLM) like SciBERT [2], which is pre-trained on scientific corpora. To assess the viability of this method for an interactive active learning loop, we conducted a benchmark experiment by performing a fine-tuning of SciBERT on the `Jeyaraman_2020` dataset. The model was trained for three epochs on a static 80% training split (940 documents) using a batch size of 8.

While the fine-tuned model demonstrated strong classification performance, achieving an F1-score[14] of **0.821** for the relevant class, the computational cost makes this approach impractical for an interactive system. The training process alone took **2 hours and 58 minutes** on a high-performance V100 GPU.

Table 4: Performance and computational cost of a single SciBERT fine-tuning cycle. The high training time makes this approach impractical for iterative retraining in an active learning loop.

| Model | Key Performance Metric | Computational Cost |
|---|---|---|
| SciBERT | F1-Score (Relevant Class): **0.821**<br>Recall (Relevant Class): **84.21%** | **Training Time (3 Epochs): ~2.97 hours**<br>Eval Time (235 docs): ~3.8 Min |

In an active learning system, the model must be retrained after each batch of user labels. A retraining cycle lasting hours renders the system non-interactive and unusable for a reviewer in a continuous session. Therefore, we conclude that while fine-tuning of LLMs is effective for static classification tasks, it is not a feasible strategy for the core engine of a responsive, human-in-the-loop active learning system due to prohibitive retraining latency. This challenge motivated the architecture of AutoDiscover: a one-time, upfront computation of high-quality static embeddings (using SPECTER2), followed by iterative, rapid retraining of a lightweight Graph Neural Network that takes seconds, not hours.

### 5.1.2 C2: Limitations of TF-IDF Similarity Graphs in Our Screening Setting

Before committing to a deep learning architecture, we evaluated the efficacy of the traditional Term Frequency-Inverse Document Frequency (TF-IDF) method for representing scientific text. TF-IDF is a strong and widely used baseline for document retrieval and clustering; however, our analysis shows that constructing a useful *similarity graph* from TF-IDF vectors can be highly sensitive to preprocessing and thresholding choices, and may yield a fragmented topology in our screening setting.

We generated TF-IDF vectors for all 1,175 documents in the `Jeyaraman_2020` dataset and constructed a similarity graph by connecting papers with a cosine similarity score above a permissive threshold of 0.2.[15] The resulting graph, shown in Figure 6, is extremely sparse, containing only 44 edges. This fragmentation is a critical structural problem, as a graph-based model cannot propagate label information between the numerous disconnected components.

[14] The F1-score is the harmonic mean of precision and recall, calculated as $2 \cdot \frac{\text{Precision}\cdot\text{Recall}}{\text{Precision}+\text{Recall}}$. It provides a single score that balances the concerns of precision (accuracy of positive predictions) and recall (completeness of positive predictions).

[15] The similarity threshold is a cutoff value. A high threshold is strict, while a low threshold is permissive. For this dataset, we found that $\tau = 0.2$ already produces very few edges (Figure 5); lowering $\tau$ increases connectivity but quickly introduces many weak links driven by generic terms, reducing interpretability of the graph structure. Lowering the threshold further would not create meaningful connections, but would instead introduce noise by linking papers based on generic terms (e.g., `group`, `score`) rather than true conceptual similarity.

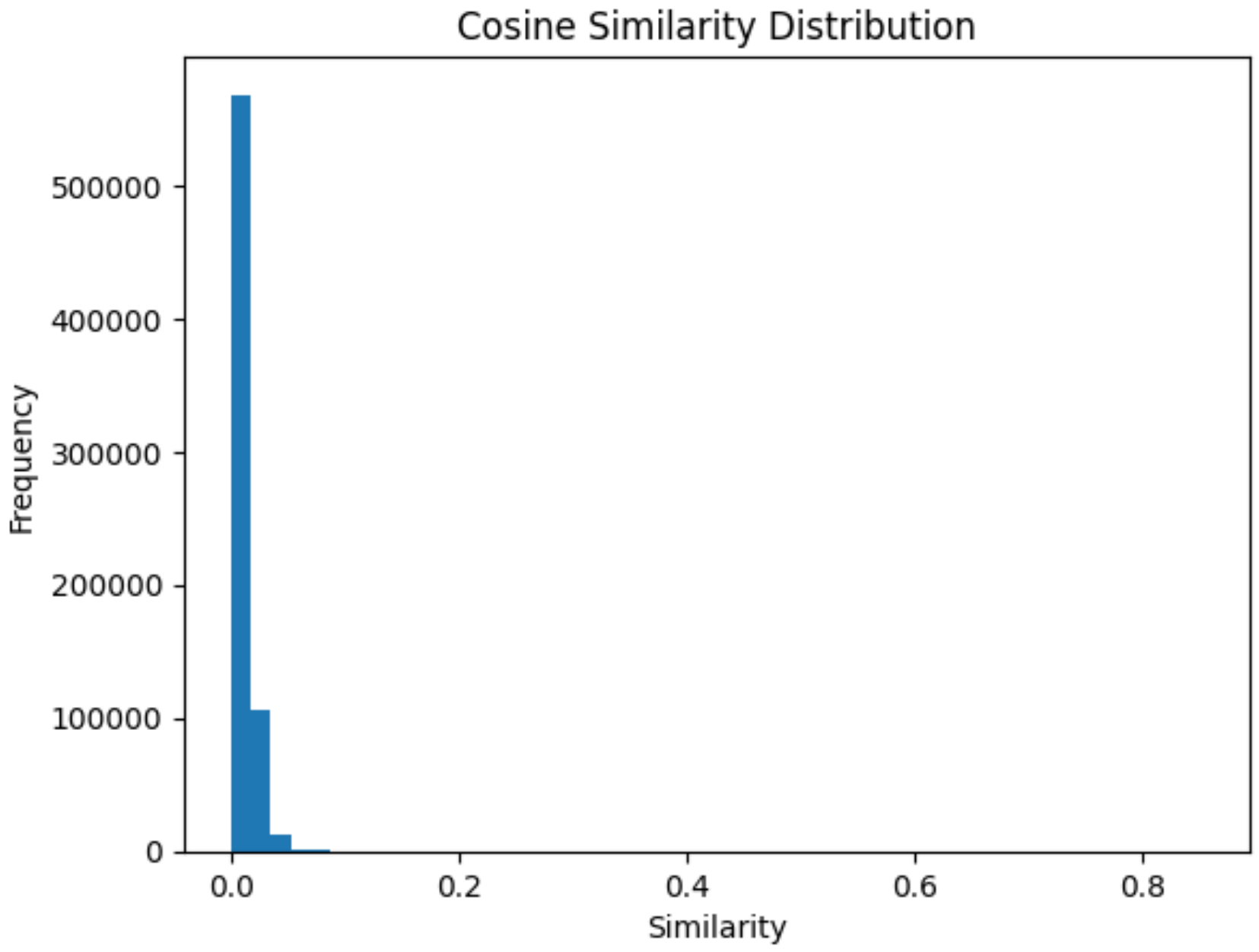

Figure 5: Distribution of cosine similarity scores between all document pairs using TF-IDF. The overwhelming concentration of scores near zero indicates that most documents share very few specific keywords, leading to a structurally disconnected graph even at low similarity thresholds.

More importantly, an analysis of the top TF-IDF terms for the known relevant documents reveals a reliance on specific, non-generalizable keywords, as listed in Table 5. Terms such as `microfracture`, `ikdc`, and `koos` are highly specific procedural or scoring terms. While useful, they create a "vocabulary mismatch" problem: a relevant paper discussing a different surgical technique or using an alternative evaluation score would have a low similarity score and would likely be missed. This keyword dependency fails to capture the underlying conceptual relationships between studies.

Table 5: Top 30 most significant TF-IDF n-grams for relevant documents in the `Jeyaraman_2020` dataset. The terms are highly specific to particular procedures, patient groups, and scoring systems.

| Rank | Term | Score | Rank | Term | Score | Rank | Term | Score |
|---|---|---|---|---|---|---|---|---|
| 1 | group | 2.5619 | 11 | cartilage | 1.4859 | 21 | outcome | 1.1955 |
| 2 | patients | 2.0887 | 12 | outcomes | 1.4558 | 22 | results | 1.1833 |
| 3 | groups | 1.8458 | 13 | months | 1.4517 | 23 | ikdc | 1.1735 |
| 4 | randomized | 1.7241 | 14 | follow | 1.3569 | 24 | koos | 1.1638 |
| 5 | microfracture | 1.6758 | 15 | scores | 1.3338 | 25 | mean | 1.1619 |
| 6 | score | 1.6529 | 16 | significant | 1.2713 | 26 | repair | 1.1485 |
| 7 | years | 1.6370 | 17 | significantly | 1.2697 | 27 | treatment | 1.1466 |
| 8 | clinical | 1.5949 | 18 | study | 1.2645 | 28 | implantation | 1.1385 |
| 9 | knee | 1.5754 | 19 | compared | 1.2170 | 29 | mri | 1.1286 |
| 10 | trial | 1.5082 | 20 | 12 | 1.2087 | 30 | autologous | 1.0866 |

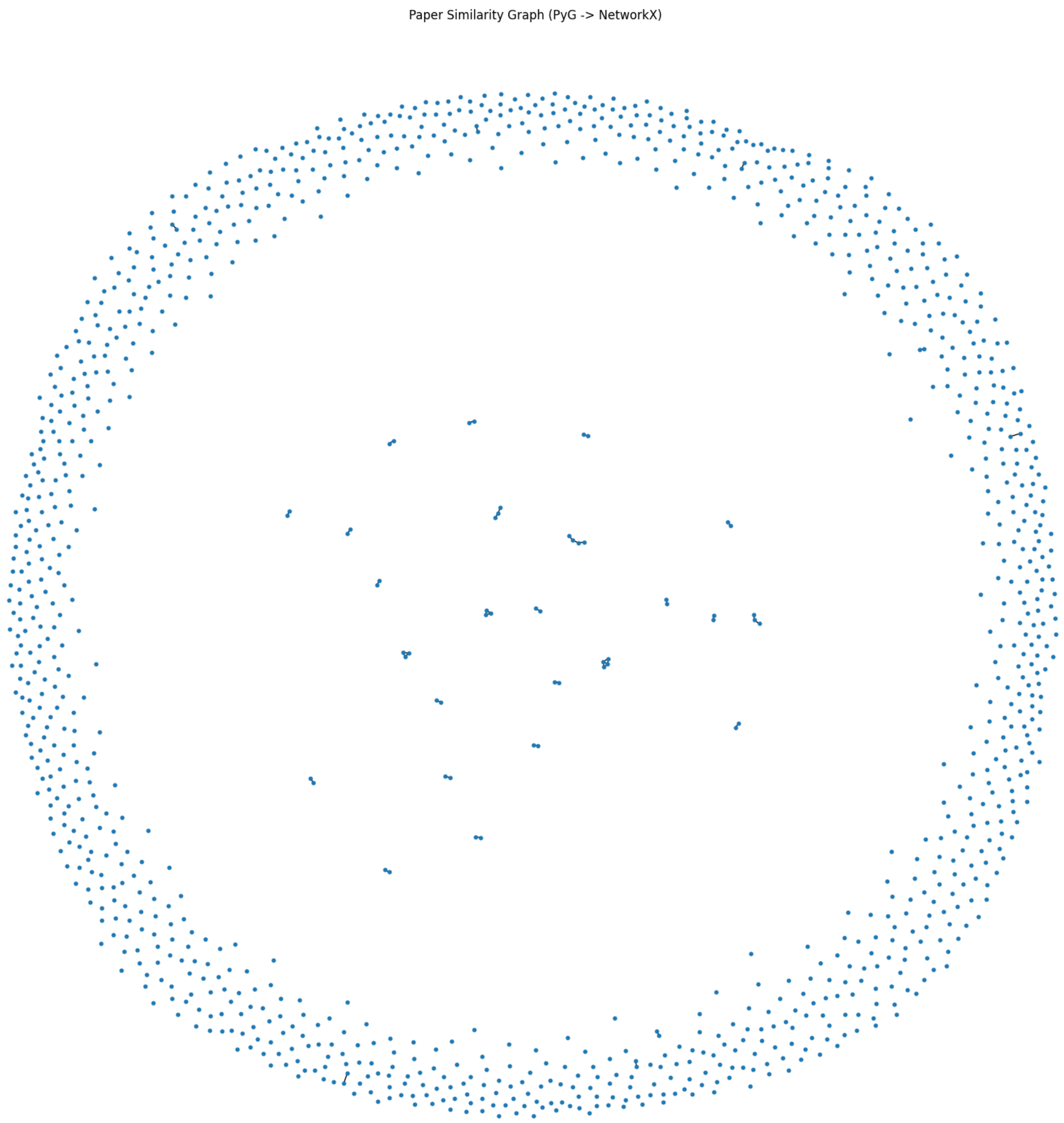

Figure 6: A visualization of the paper-paper similarity graph constructed from the Jeyaraman_2020 dataset. The graph's nodes represent 1,175 documents, featurized using TF-IDF vectors derived from their titles and abstracts. An edge is created between two nodes if the cosine similarity of their vectors exceeds a threshold of **0.2**. This specific parameterization results in a highly fragmented graph with only 44 edges, illustrating the failure of keyword-based methods to capture broader semantic relationships. The graph layout and rendering were generated using the **NetworkX** [21] and **Matplotlib** [25] libraries.

This analysis demonstrates that TF-IDF struggles to connect documents that are conceptually similar but use different terminology, a challenge in systematic reviews where diverse phrasing is common. This evidence confirmed the need for a semantically-aware model. We therefore chose transformer-based embeddings like SPECTER2, which understand the meaning behind the words, not just the keywords themselves. We note that alternative bag-of-words similarity formulations (e.g., Jaccard-style overlap on n-grams) and topic-model-based representations can also induce meaningful document structure; we focus here on SPECTER2 because it provides a robust semantic neighborhood signal with less parameter sensitivity in our experiments.

### 5.1.3 C3: Sparsity of Explicit Graph Structures

Recognizing the importance of relational data, we next evaluated several static graph structures to determine if they could provide a robust backbone for active learning.

**Citation Networks.** First, we investigated if a direct citation network could form a sufficiently connected graph for effective information propagation. To test this, we constructed a directed citation graph from the 1,175 papers in the `Jeyaraman_2020` dataset, where an edge exists from a citing paper to a cited paper only if both are within the corpus.[16] The results of our network analysis, summarized in Table 6, revealed a structurally deficient graph. With a network density of only 0.48%, the graph is extremely sparse. Critically, the analysis of strongly connected components showed that the graph is a directed acyclic graph (a forest of small trees), with no cyclical citation patterns and a largest component size of only a single node. This fragmentation means that relevant documents exist on disconnected "islands," making it impossible for label information to propagate across the entire set of relevant literature.
This finding motivated our use of **SPECTER2 embeddings** [43]. SPECTER2 is pre-trained on a massive citation graph, not just text, using a method that explicitly encodes citation links into the embedding space.[17] This pre-training allows SPECTER2 to place semantically related papers near each other in the embedding space even if they do not directly cite one another within our specific dataset, thus creating a dense, implicit graph that overcomes the sparsity of the explicit citation network.

Table 6: Topological Analysis of the Explicit Citation Graph for the `Jeyaraman_2020` Dataset. The metrics, calculated using the NetworkX library, reveal a structurally sparse and fragmented graph, which is unsuitable for effective information propagation.

| **Metric** | **Value** | **Interpretation (What this means for the graph)** |
|---|---|---|
| Total Nodes | 1,175 | The total number of unique papers in the dataset corpus. |
| Total Edges | 6,645 | The number of citation links where both the citing and cited paper are present within this specific dataset. |
| Average Out-Degree | 5.66 | On average, each paper in the corpus cites approximately 5-6 other papers that are also in the corpus. |
| **Network Density** | **0.0048** | Calculated as $\frac{E}{N(N-1)}$, this shows that only **0.48%** of all possible directed citation links actually exist. This is a quantitative indicator of an extremely sparse graph. |
| No. of Strongly Connected Components | 1,175 | A strongly connected component is a group of nodes where every node is reachable from every other. Since this number equals the total number of nodes, it proves there are no cyclical citation patterns. |
| **Largest SCC Size** | **1** | This is the most critical finding. It confirms that the graph is a **directed acyclic graph** (a "forest") completely fragmented into individual nodes and small, non-reciprocal citation chains. There are no interconnected communities for information to circulate within. |

**Concept and Topic-Based Networks.** Next, we investigated if shared metadata tags could provide better connectivity by building two bipartite graphs connecting papers to their OpenAlex ‘concept‘ and ‘topic‘ tags. The results, summarized in Table 7, show that the fine-grained ‘concept‘ tags create a significantly denser and more informative graph. This is a direct result of the metadata's granularity; the corpus contains 804 unique ‘concept‘ nodes but only 98 ‘topic‘ nodes, resulting in five times as many edges for the concept-based graph (17,550 vs. 3,495).

[16]This is a critical detail: the graph only includes *internal* citations between documents that are both part of the specific review corpus. The vast majority of a paper's references will naturally fall outside the narrow scope of the dataset and are therefore discarded. This methodological constraint is the primary reason for the extreme sparsity observed in the network, as it only captures a small fraction of the true citation landscape.

[17]The SPECTER2 paper states its training objective is to generate embeddings where "the document is closer to the embedding of a paper it cites than to a randomly sampled paper from the corpus." This is achieved via a triplet loss function, forcing the model to learn what makes two papers semantically related in the context of scientific citations.

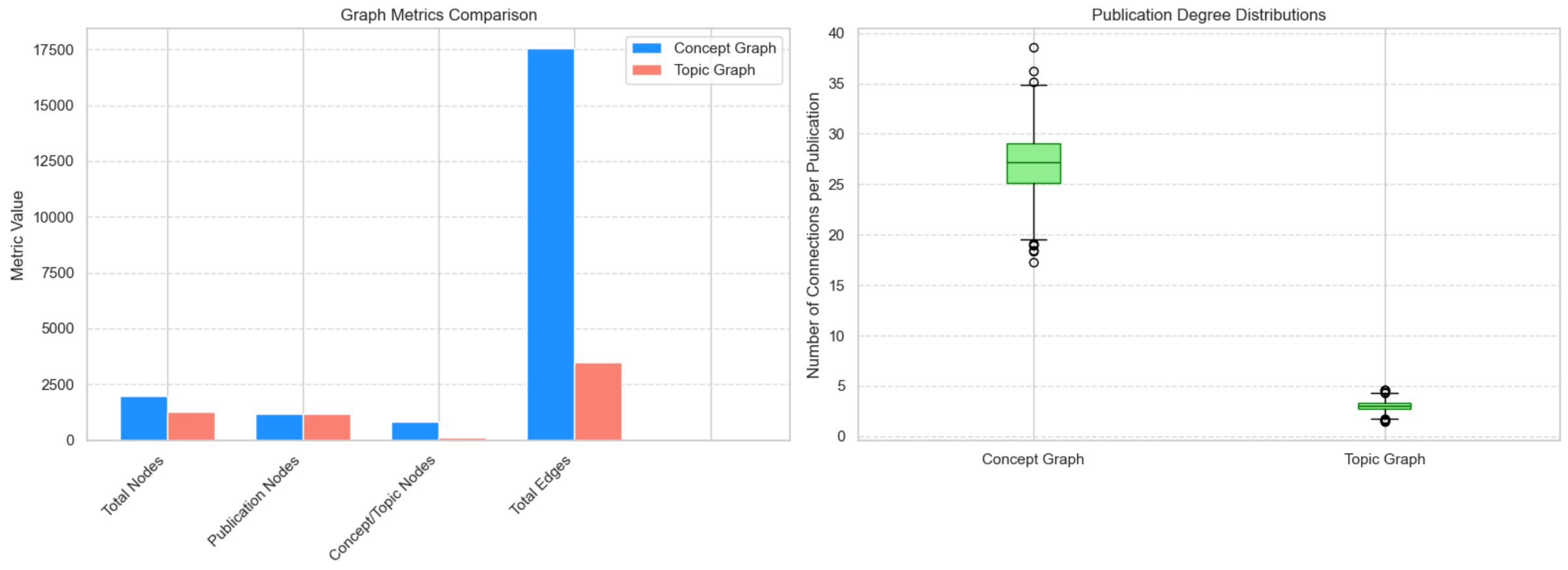

Figure 7: A comparison of graph metrics derived from connecting papers to their OpenAlex ‘concept‘ versus ‘topic‘ attributes on the Jeyaraman_2020 dataset. The concept-based graph (left) is far denser and offers richer connectivity, justifying the use of granular metadata to build informative graphs.

Table 7: Comparison of Bipartite Graphs Constructed Using OpenAlex Concepts vs. Topics for the `Jeyaraman_2020` Dataset. The statistics clearly show that concepts provide a much richer and denser network structure.

| **Metric** | **Concept-Based Graph** | **Topic-Based Graph** |
|---|---|---|
| Total Nodes | 1,979 | 1,273 |
| *(Paper Nodes)* | *1,175* | *1,175* |
| *(Concept or Topic Nodes)* | *804* | 98 |
| **Total Edges** | **17,550** | **3,495** |
| Average Degree | 17.74 | 5.49 |
| Relevant Papers Avg. Degree | 13.70 | 3.00 |
| Non-Relevant Papers Avg. Degree | 15.05 | 2.97 |
| **T-test p-value (Relevant vs. Non-Relevant)** | **7.79e-05** | 1.39e-05 |

To determine if this connectivity was meaningful for classification, we conducted a t-test[18] to compare the average number of connections (degree) for relevant versus non-relevant papers. For both graph types, we found a statistically significant difference ($p \ll 0.01$), confirming that these metadata links provide a useful signal for distinguishing relevant documents.
Our analysis concluded that while the concept-based graph was structurally superior, its reliance on a proprietary, algorithmically generated taxonomy (OpenAlex) presented a trade-off. In contrast, Medical Subject Headings (MeSH) are a manually curated and universally accepted standard for indexing in the biomedical and health sciences [29]. Despite our data quality audit (Figure 4) showing that MeSH metadata was often sparser than concepts, we made a pragmatic design choice. We prioritized the use of

[18] A Welch’s t-test was used to determine if the average number of metadata connections (node degree) for relevant papers was statistically different from that of non-relevant papers.

- **What it is:** A t-test is a statistical hypothesis test that compares the means (averages) of two groups to see if they are likely different from each other.
- **Why we used it:** To scientifically validate that the graph structures we built are meaningful. If relevant papers connect to concepts or topics differently than non-relevant ones, the structure contains a useful signal for classification.
- **How it works:** The test calculates a **p-value**, which is the probability of observing the data we have if there were truly no difference between the groups (the ”null hypothesis”). A p-value below a chosen significance level (e.g., $\alpha = 0.01$) allows us to reject the null hypothesis.
- **Our Result:** The tests yielded p-values of $7.79 \times 10^{-5}$ for the concept graph and $1.39 \times 10^{-5}$ for the topic graph. Since both are far below 0.01, we conclude that the difference in connectivity is statistically significant, confirming that these graph structures are indeed informative.

**MeSH** terms in our final architecture to ground the system in this established, controlled vocabulary, favoring domain-specific authority and standardization over the denser but less standardized connectivity provided by OpenAlex concepts.

#### 5.1.4 C4: Justify an Adaptive Exploration-Exploitation Strategy

We analyzed the structure of the citation network to understand if a simple, static discovery strategy would be sufficient. A common assumption is that relevant papers tend to cite other relevant papers, a principle known as "citation chasing." If this were universally true, a purely exploitative strategy of following citations from known relevant documents would likely be optimal.

To test this, we constructed a specific subgraph from the `Jeyaraman_2020` dataset. This differs from the graph in C3 as it is not the full corpus network; instead, it contains only the known relevant papers and any other papers they directly cite within the corpus. Visualizing this "ground-truth" citation landscape, as shown in Figure 8, reveals two distinct structural patterns.

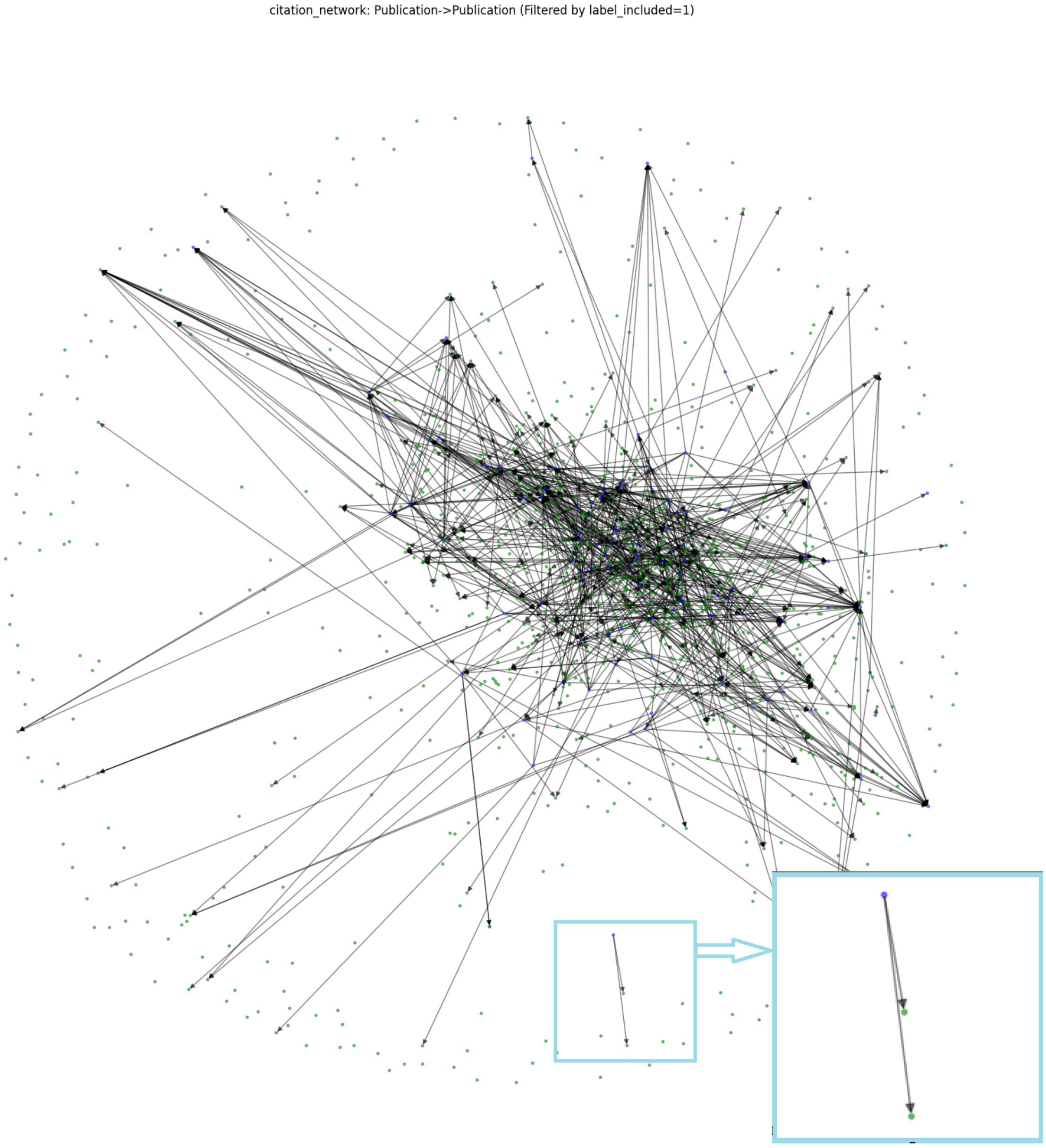

Figure 8: Visualization of the citation subgraph originating from known relevant documents in the `Jeyaraman_2020` dataset. Relevant papers (blue) form a large, highly interconnected component, suggesting the utility of an exploitation strategy. However, the presence of links to irrelevant papers (green) that act as bridges to other relevant "islands" highlights the necessity of an exploration strategy to ensure a comprehensive review.

First, the figure shows a large, highly interconnected component where the majority of relevant papers (blue nodes) cite other relevant papers. This tightly-knit community provides a clear rationale for an **exploitation** strategy. The existence of this cluster suggests that following citation links from a known relevant paper is a high-probability method for discovering more relevant documents within that same community.

However, the visualization also exposes a critical second pattern: relevant papers exist as smaller, peripheral "islands." These are connected to the main group only by citing an irrelevant paper (green node). A purely exploitative model, by definition, would not query these irrelevant "bridge" documents, making the relevant papers behind them undiscoverable.[19] This pattern demonstrates the necessity of an **exploration** strategy, one that is willing to query uncertain or structurally diverse documents, to discover relevant papers outside the main cluster.

This analysis illustrates the fundamental exploration-exploitation dilemma in systematic reviews. The presence of both a highly interconnected relevant component and peripheral, bridge-linked relevant papers provides the final justification for our core architectural decision: an effective system requires an adaptive agent that can dynamically balance these competing discovery patterns.

#### 5.1.5 C5: Case Study Evidence of Non-Linear Separability in Semantic Space

Having established the need for semantic representations, we next investigated if SPECTER2 embeddings alone were sufficient for classification. To test this, we analyzed the structure of the embedding space by generating a 2D UMAP [30] projection of the SPECTER2 embeddings for the `Jeyaraman_2020` dataset, shown in Figure 9. This visualization is presented as an illustrative case study; the broader impact of non-linear modeling is assessed empirically in the cross-dataset evaluation in Section 7.

The visualization reveals that the relevant papers (red) are not randomly scattered; they are concentrated in specific regions of the embedding space, indicating that SPECTER2 successfully captures a semantic signal. However, these relevant papers do not form a single, cleanly separable cluster. Instead, they appear as multiple distinct groups and isolated points, interspersed with irrelevant documents (blue). This proximity highlights a key challenge: an irrelevant paper may share topical similarity with a relevant one but fail to meet a specific review criterion (e.g., wrong study design), causing it to land nearby in the embedding space.

This complex, non-linear separability is a challenge. A simple linear classifier,[20] which can only separate data with a straight line or hyperplane, would fail to draw an effective decision boundary in this space. This directly motivates the use of a more powerful, non-linear model.

Our conclusion is that while SPECTER2 provides a powerful feature representation, these features are not sufficient for a simple classifier. This led to our decision to use a Graph Neural Network. By constructing a graph that connects semantically similar nodes (i.e., those that are close in the embedding space), a GNN can exploit these local neighborhood structures. It can propagate label information between nearby relevant papers while learning the nuanced, non-linear boundaries needed to distinguish them from their topically similar but irrelevant neighbors.

[19] This can be shown by proof by contradiction. Assume a purely exploitative strategy queries the node with the highest predicted relevance. In the early stages, an irrelevant "bridge" node will have a lower predicted relevance than other unqueried nodes within the main relevant cluster. Therefore, the strategy will never select the bridge node, and the relevant "island" node will never be reached.

[20] A linear classifier, such as Logistic Regression or a Support Vector Machine with a linear kernel, learns a single hyperplane to separate classes. Such a model would be unable to draw the complex, disjoint boundaries needed to isolate the multiple "islands of relevance" seen in the UMAP plot.

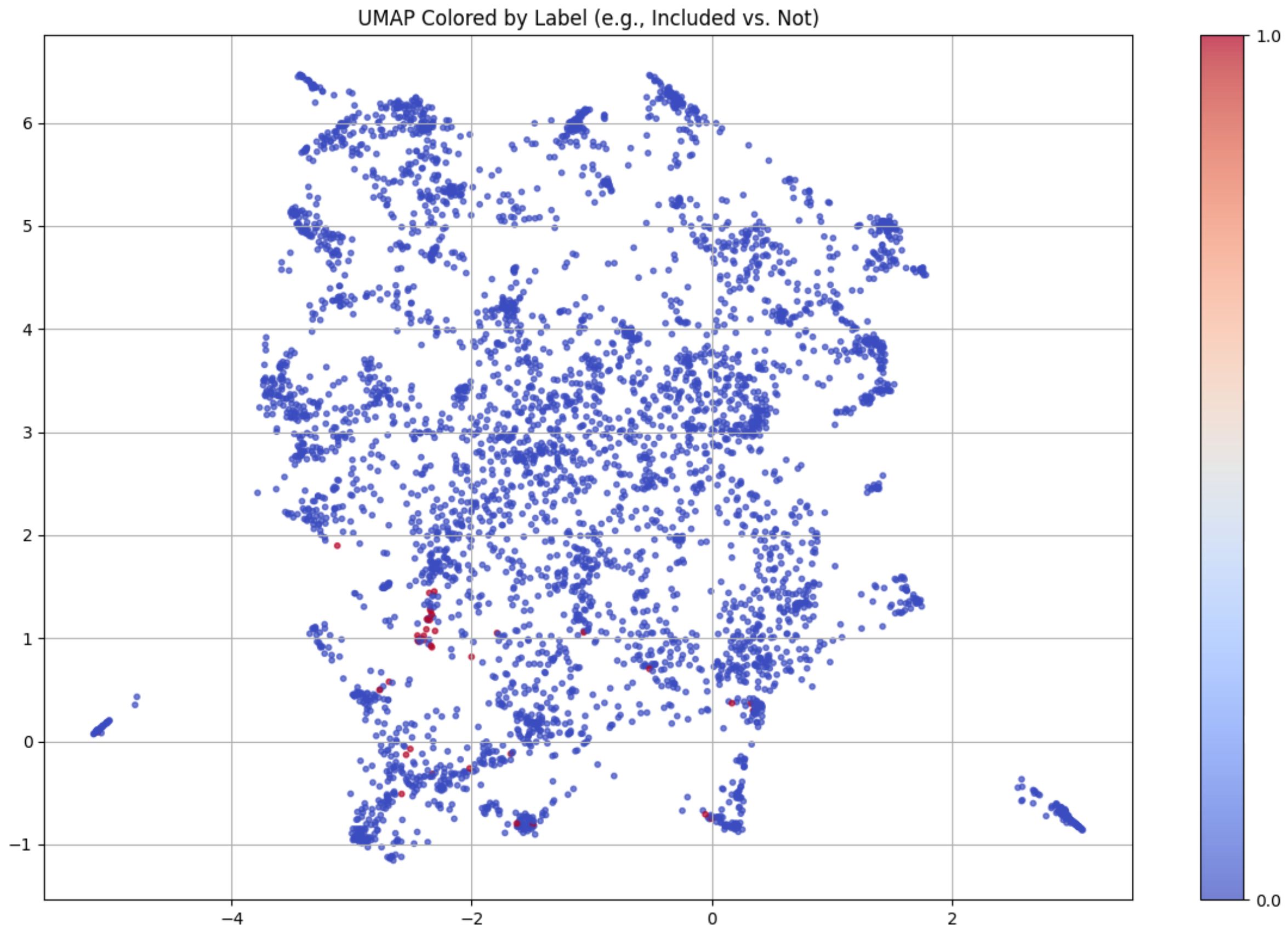

Figure 9: UMAP projection of SPECTER2 embeddings for the `Jeyaraman_2020` dataset, colored by relevance. Relevant papers (red) are largely concentrated in specific regions of the feature space, primarily within the coordinate range of approximately **x=[-3, 0.5]** and **y=[-1, 2]**. However, they do not form a single, linearly separable cluster. Instead, they are interspersed with irrelevant papers (blue), highlighting a complex, non-linear classification problem and showing that SPECTER2 embeddings alone are not sufficient for a simple classifier.

#### 5.1.6 C6: A Greedy GNN Fails to Overcome the Cold-Start Problem

While a GNN operating on SPECTER2 embeddings can model non-linear semantic relationships, its effectiveness depends on sufficient training data. In the critical cold-start phase of a systematic review, which begins with only a single known relevant document, a supervised model like a GNN is highly susceptible to overfitting to that single example or failing to generalize entirely [48].
To test the severity of this challenge, we conducted a targeted ablation study on the `Hall_2012` dataset, comparing the full AutoDiscover system against a version where the adaptive DTS agent was disabled. This forced the system to use a purely greedy strategy, relying only on the GNN's top-ranked prediction (the "GNN Exploit" arm) at every step.
The results are shown in Figure 10. In the greedy run (Figure 10a), the GNN fails to learn effectively from the single positive seed; its predictions are insufficient to identify other relevant documents, and the system discovers no new positive instances. In contrast, the full AutoDiscover system with the DTS agent enabled (Figure 10b) deploys exploratory strategies that provide the GNN with diverse examples. This allows the system to escape the cold-start phase and achieve 95% recall after screening only 4.9% of the dataset.
This experiment demonstrates that a GNN, even when built on powerful semantic embeddings, is not sufficient on its own. The model's representational power must be paired with an effective exploration-exploitation mechanism to overcome initial label scarcity. This finding provides the primary justification for the agent-based architecture of AutoDiscover.

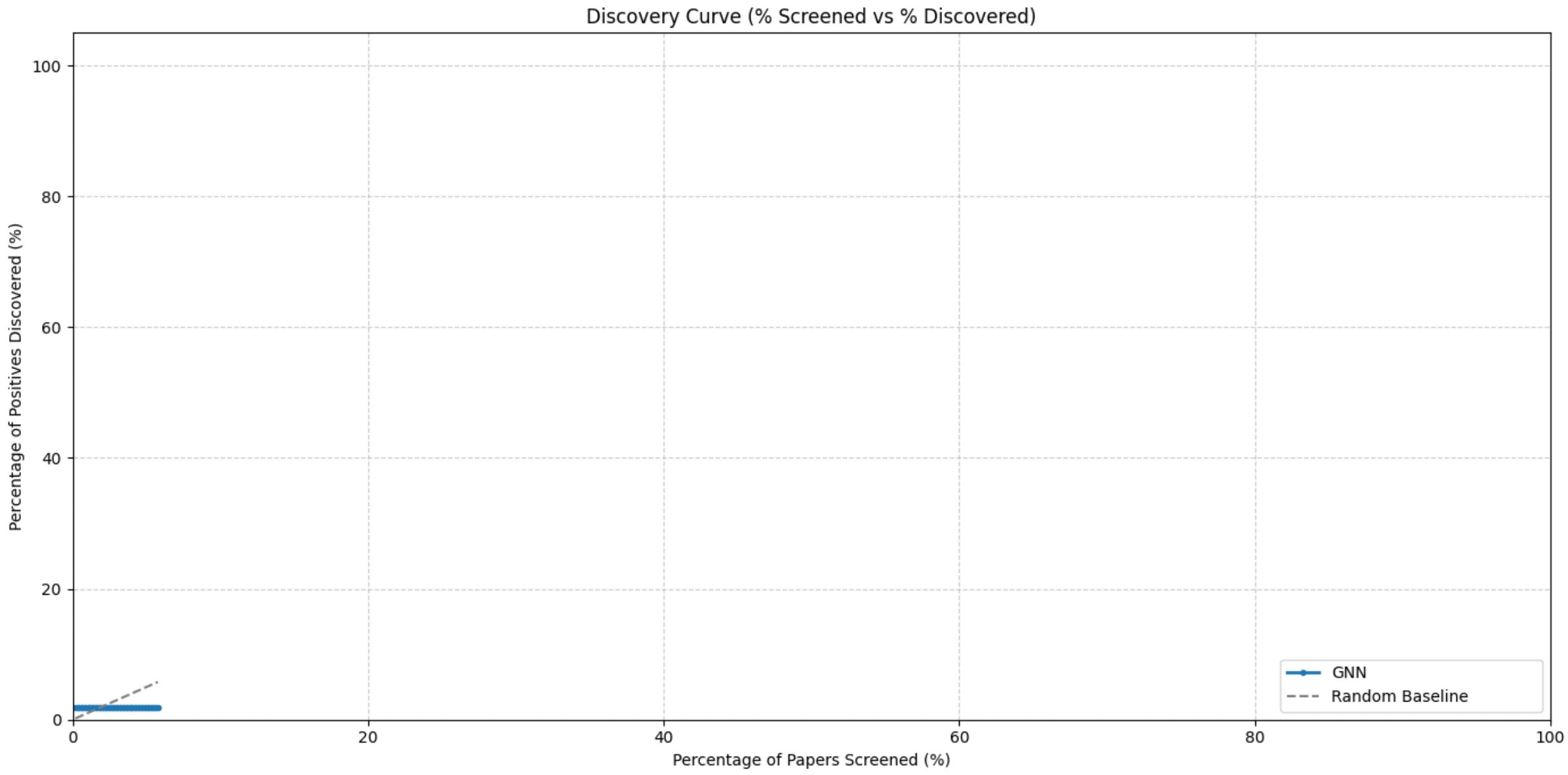

(a) AutoDiscover with a Greedy Strategy (GNN Exploit Only)

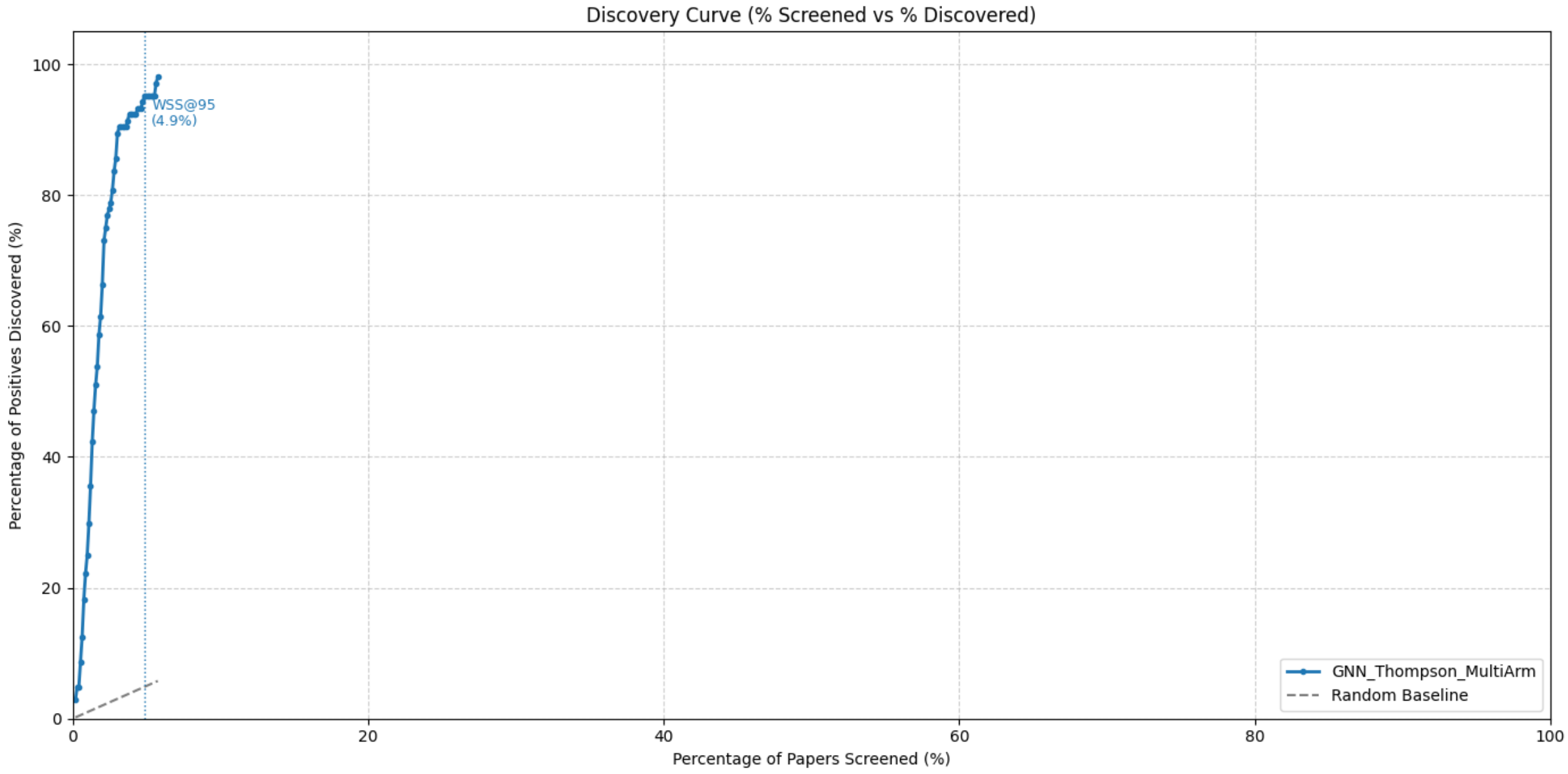

(b) AutoDiscover with Adaptive DTS Agent Enabled

Figure 10: An ablation study on the `Hall_2012` dataset. (a) Without the adaptive agent, a purely greedy GNN-based strategy fails to escape the cold-start phase and discovers no additional relevant documents. (b) With the DTS agent enabled, the system successfully balances exploration and exploitation, bootstrapping an effective discovery process.

#### 5.1.7 C7: The Inadequacy of a Fixed Active Learning Strategy

Having established the need for a graph-based model, the final question was whether a single, well-chosen but static query strategy would be sufficient. If one strategy (e.g., uncertainty sampling) consistently outperformed all others, an adaptive agent would be unnecessary.

To test this, we conducted an experiment applying several common active learning strategies to a static graph with a Label Propagation model. The discovery curves for each strategy are plotted in Figure 11. The results show that the optimal strategy is non-stationary, as the intersecting lines demonstrate that no single strategy is dominant throughout the screening process. A strategy that excels in early, exploratory iterations may be outperformed by a more exploitative one later on.

This finding provides the final evidence motivating our architecture. A successful system requires not only an adaptive model but also an adaptive controller that can learn the best query strategy for the current phase of the review. This directly justifies the core innovation of AutoDiscover: a Discounted Thompson Sampling agent that manages a dynamic portfolio of strategies. The necessity of this portfolio is empirically substantiated by our final results (Appendix E.2), where the agent utilizes nearly every available strategy across the 26-dataset benchmark, providing a powerful proof by contradiction[21] to the idea of a single optimal strategy.

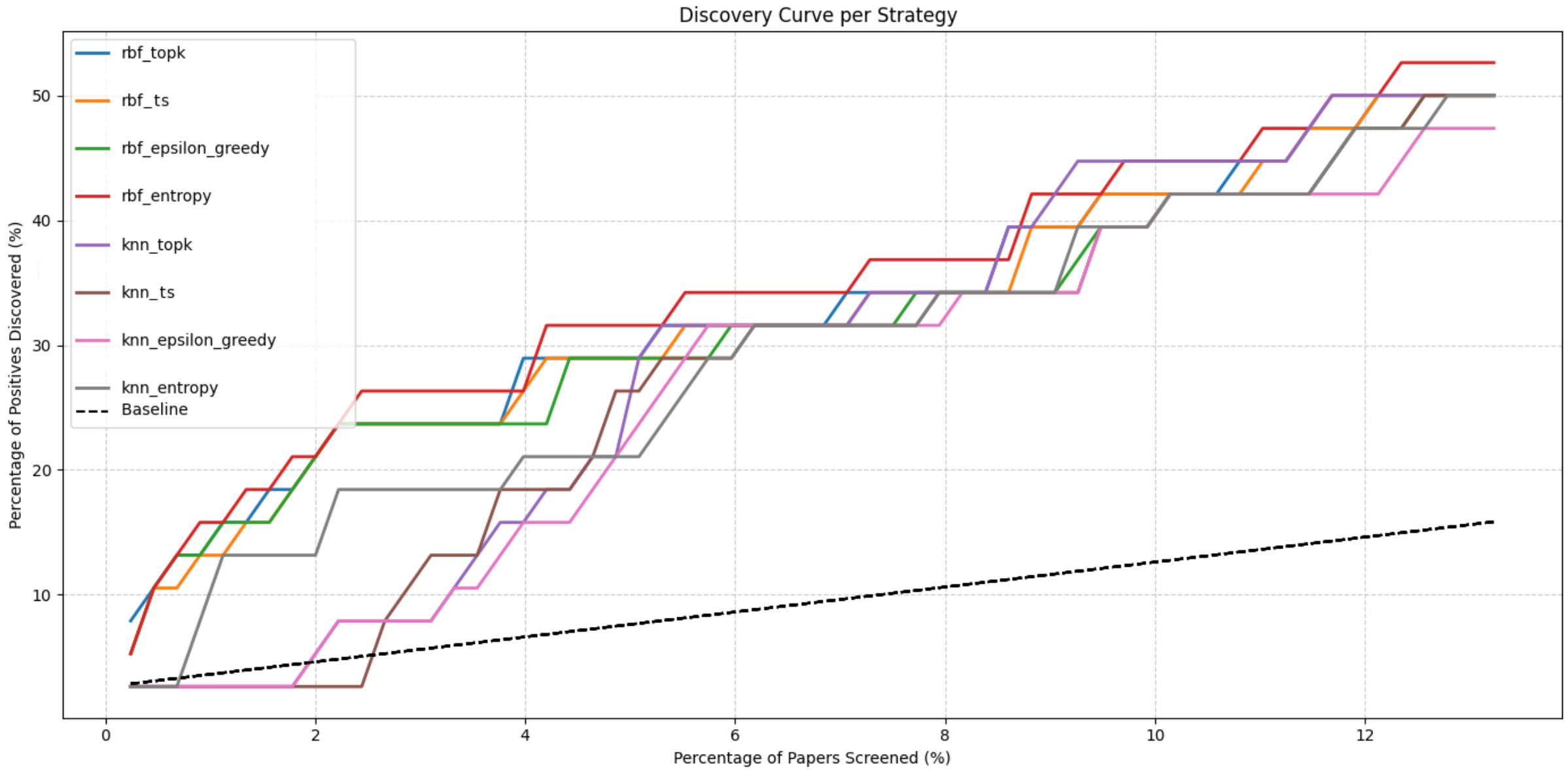

Figure 11: A comparison of discovery curves for various active learning strategies applied to a static graph with a Label Propagation model. The varied performance, where different strategies excel at different stages, demonstrates that no single strategy is dominant, motivating an adaptive approach.

#### 5.1.8 The AutoDiscover Rationale: A Synergy of Semantic Embeddings, Graph Learning, and an Adaptive Agent

These preliminary findings collectively demonstrated that no single, static approach is sufficient. They motivated the three core design principles of AutoDiscover: (1) using pre-trained semantic embeddings to overcome data sparsity and vocabulary mismatch; (2) modeling the literature as a heterogeneous graph to capture complex, non-linear relationships; and, most critically, (3) employing an adaptive agent to dynamically select the optimal query strategy. This synergy led to the final architecture: a Heterogeneous graph Attention Network (HAN) managed by a Discounted Thompson Sampling (DTS) agent.

Our hypothesis is that a dynamic, adaptive attention mechanism is critical for achieving robust performance across diverse and imperfect datasets. A model that relies on a single, fixed relationship type would be vulnerable to failure on datasets where that attribute is sparse. For instance, a model dependent on citation links would under-perform on a dataset with incomplete citation metadata, a problem observed in our data quality audit (Figure 4).

By learning separate attention weights for each relation, a Heterogeneous graph Attention Network (HAN) [52] can mitigate this risk. The architecture allows the model to dynamically up-weight the importance of well-populated and currently informative relations while down-weighting relations from sparse or less useful attributes in different phases of SLRs. This adaptivity is not only important for handling data quality issues but also for navigating the non-stationary dynamics of the active learning loop itself. As the model is retrained with new labels, its understanding of the feature space evolves. A relationship type that was highly predictive in the cold-start phase might become less so as more nuanced criteria for relevance are learned. The HAN's ability to continuously re-weigh the influence of different relationship types allows it to adapt to the specific data landscape and the current state of the learning process. This makes the HAN architecture well-suited for modeling the noisy and evolving relational patterns in real-world SLRs. The overall effectiveness of this approach is substantiated by the strong performance of AutoDiscover across the diverse SYNERGY benchmark (Section 7).

[21] The argument can be framed as a proof by contradiction. **Assume**, for contradiction, that a single static strategy is universally optimal. **If this were true**, a rational agent designed to maximize rewards, like DTS, would have learned to exclusively select that one arm across all datasets. **However**, our empirical evidence (Appendix E.2) shows the opposite: the agent dynamically selects from the entire portfolio, demonstrating that the utility of each arm is context-dependent. This contradicts the initial assumption, thereby validating the need for a diverse, managed portfolio of strategies.

While the HAN architecture provides a mechanism for adapting to data sparsity and evolving model knowledge, its effectiveness is still contingent on being sufficiently trained. In the critical cold-start phase of an SLR, the process often begins with only a single known relevant document. A supervised deep learning model like a GNN has an insufficient data to learn and for generalization; it is mathematically prone to learn a trivial solution, such as classifying all documents as irrelevant, or to severely overfit to the specific features of that single positive instance [10,48]. Consequently, the GNN's predictions and its learned attention weights are inherently unreliable. A strategy based solely on the GNN's output (e.g., GNN Exploit) would therefore risk failing to discover any new relevant documents. This exact failure mode is empirically demonstrated in our ablation study (Section 5.1.6), which motivates the use of a higher-level adaptive controller.

A hypothesis of this thesis is that, given the infinite variety of systematic literature reviews, spanning diverse topics, class imbalances, and varying metadata quality across datasets, no single static algorithm can be universally optimal. An adaptive approach is therefore hypothesized to offer greater robustness. This hypothesis is supported by our comparative analysis (Table 13), where the adaptive AutoDiscover framework achieves a higher median Discovery Rate Efficiency than the static baselines across the 26-dataset benchmark.

Our framework therefore employs a Discounted Thompson Sampling (DTS) agent to manage a diverse portfolio of query strategies. This approach decouples the strategy selection from sole reliance on the under-trained GNN. Crucially, the portfolio includes model-agnostic strategies, such as a strategy based on Label Propagation (LP) [59][22]. The LP arm operates directly on the graph structure, providing a powerful discovery mechanism when the GNN is still learning. We apply LP specifically to the subgraph of semantic similarity links. This choice is deliberate: this subgraph, built from SPECTER2 embeddings, is the densest and most richly connected. This aligns with our initial hypothesis, as SPECTER2 itself is pre-trained on a massive citation graph and is therefore designed to produce embeddings that implicitly capture citation-based relatedness [43]. Using these embeddings to build a dense semantic graph thus serves to bridge the information gaps left by sparse, explicit citation data. By including this LP-based arm, the DTS agent is equipped with a robust, semi-supervised tool that is particularly effective for bootstrapping discovery[23] from a minimal set of initial labels. The necessity and efficacy of maintaining this diverse portfolio is visually substantiated in Appendix E.2; the arm usage charts demonstrate that the agent dynamically alters its strategy based on the dataset and that nearly all arms contribute to discovering relevant documents at some point, confirming their situational utility.

[22]**Label Propagation (LP)** is a semi-supervised algorithm that infers labels for unlabeled nodes by iteratively "propagating" or "spreading" label information from the small set of known labeled nodes to their neighbors. It is computationally efficient and can be effective with very few initial labels if the graph structure is meaningful.

[23]**Bootstrapping discovery** refers to the process by which an active learning system uses the information from a very small set of initial labeled examples to find more relevant examples, which in turn improves the model, allowing it to find even more, and so on. It is the crucial initial phase of "pulling itself up by its own bootstraps" to escape the cold-start problem.

## 5.2 AutoDiscover: A Modular and Adaptive Framework

AutoDiscover is an end-to-end framework designed to accelerate systematic literature reviews. It takes a standardized CSV file of scholarly metadata as input, processes it through a configurable pipeline to build a heterogeneous graph, and then executes an active learning loop to intelligently recommend papers for manual annotation. The final output is a ranked list of documents and a set of performance logs for analysis.

The framework's modularity is managed by a central configuration file, which allows each stage of the pipeline to be customized without altering the source code. This makes the pipeline dynamic, as its behavior can be adapted for different datasets and research goals. The entire process, depicted in Figure 12, is composed of five stages.

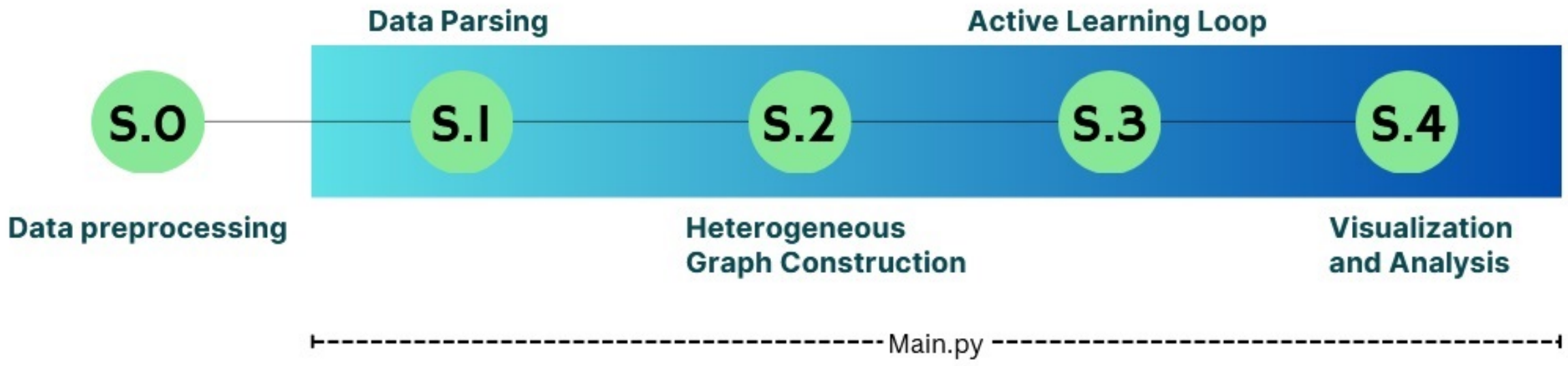

Figure 12: A visual overview of the AutoDiscover pipeline, which acts as a table of contents for the methodology. The process begins with an offline data preprocessing step (S.0). The main framework, orchestrated by `main.py`, then executes four stages: Data Parsing (S.1), Heterogeneous Graph Construction (S.2), the core Active Learning Loop (S.3), and final Visualization and Analysis (S.4).

**Stage 0: Data preprocessing.** Before training, input data must be standardized into a consistent format. Although the SYNERGY benchmark datasets [13] share a general structure, they contain nested fields (e.g., lists of authors, tokenized abstracts) that require careful parsing and transformation. The preprocessing routine extracts the relevant fields, resolves nested or complex attributes, merges publication records with their associated relevance labels, and outputs a structured CSV file for each dataset. This process ensures a unified schema across all datasets, allowing the core pipeline to operate independently of the original data structure.

While the current implementation is tailored to the SYNERGY format, the preprocessing interface can be adapted to other corpora with similar metadata. However, doing so may require modifications to handle dataset-specific nesting and attribute naming. In particular, the interface assumes consistent naming conventions for key fields (e.g., `id`, `abstract`, `label_included`) in the resulting CSVs. If another dataset diverges from these assumptions, those mappings must be updated to maintain compatibility across downstream components.

**Stage 1: Data parsing.** Following preprocessing, the input data is parsed to retain only the fields essential for downstream graph construction and representation learning. This includes document identifiers, binary relevance labels, and core textual fields such as titles and abstracts. In addition, relational metadata such as authorship, referenced works, MeSH terms, concepts, and topical descriptors are preserved. Non-essential fields are discarded at this stage to reduce memory usage and streamline processing in subsequent stages.

**Stage 2: Heterogeneous Graph Construction.** In the second stage of the pipeline, the graph builder transforms the parsed tabular data into a structured heterogeneous graph using PyTorch Geometric's `HeteroData` representation (Figure 14). The graph includes nodes of various types, papers, authors, institutions, MeSH terms, connected by typed edges based on citation, authorship, co-authorship, semantic similarity, and affiliation. This representation enables multi-relational reasoning across scholarly metadata.

The construction process is modular and configurable:

- **Node Construction.** Each enabled entity type (e.g., paper, author, institution, mesh term) is instantiated as a node set.

- **Feature Initialization.** Paper nodes are embedded using a pre-trained language model (e.g., SPECTER2 [43]), with embeddings cached optionally to reduce redundancy.
- **Edge Creation.** Edges are added for relationships toggled in the configuration. Semantic similarity is governed by a cosine threshold; co-authorship links require a minimum collaboration count; citation and topical links are instantiated based on parsed metadata.
- **Graph Assembly.** All nodes, features, and edge types are combined into a unified `HeteroData` object. This object is then serialized for downstream learning tasks.

The framework's core data structure is a heterogeneous graph, whose meta-schema is depicted in Figure 13. This structure is designed to capture the multi-relational nature of scholarly data.

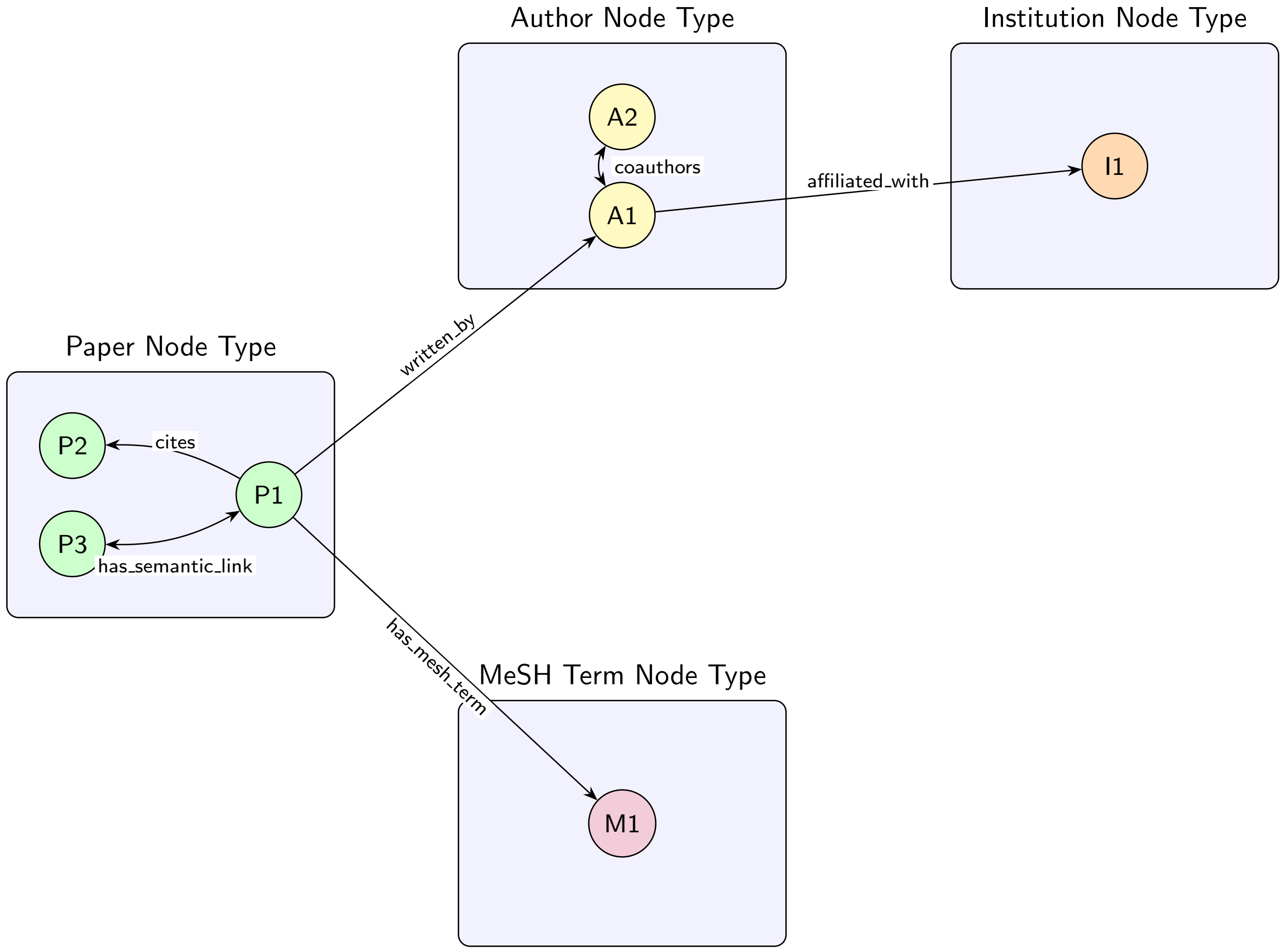

Figure 13: The meta-schema of the heterogeneous graph used in AutoDiscover. Each node type is shown as a container with example instances (e.g., **P1**, **A1**). The diagram illustrates the multi-relational structure, showing internal relationships (e.g., `cites`, `coauthors`) and external connections between different node types (e.g., `written_by`, `has_mesh_term`).

The node types include 'paper' as the primary entity, along with 'author', 'institution', and 'mesh' terms. Paper nodes are initialized with dense semantic embeddings[24] derived from their title and abstract using SPECTERv2. The other node types ('author', 'institution', 'mesh') are initialized with lower-dimensional, learnable embeddings. This is done to provide the GNN with trainable parameters for these entities, allowing it to learn meaningful representations for them from the graph structure and the few labeled examples, even when they lack initial descriptive features.

The edges between these nodes were chosen to represent well-understood signals in scholarly communication, as detailed in Table 8. These relationships allow the GNN to learn from direct citations, semantic content similarity, shared medical terminologies, authorship patterns, co-author collaborations, and institutional affiliations.

[24] A **dense embedding** is a low-dimensional vector where most values are non-zero, designed to capture semantic meaning. This contrasts with high-dimensional, **sparse** vectors like TF-IDF, where most values are zero and meaning is derived from specific keyword matches.

Table 8: Heterogeneous Graph Edge Types and Rationale.

| Relation Type (Source → Target) | Weight | Rationale and Supporting Literature |
|---|---|---|
| paper $\xrightarrow{\text{cites}}$ paper | 1.0 | Direct citation links are strong indicators of topical relatedness and influence [17]. |
| paper $\xleftrightarrow{\text{has_semantic_link}}$ paper | Norm. CosSim | Connects papers via high cosine similarity of SPECTERv2 embeddings, capturing content similarity beyond explicit citations [43]. Weights are the normalized scores. |
| paper $\xrightarrow{\text{has_mesh_term}}$ mesh | 1.0 | Links papers to their assigned MeSH terms, indicating shared biomedical topics. Used in biomedical GNNs [57]. |
| paper $\xrightarrow{\text{written_by}}$ author | 1.0 | Connects authors to the papers they have written. |
| author $\xleftrightarrow{\text{coauthors}}$ author | 1.0 | Indicates collaboration strength. An edge is formed if the co-publication count meets a threshold ($\tau_{co} \geq 2$), a practice to denote collaborations [32]. |
| author $\xrightarrow{\text{affiliated_with}}$ institution | 1.0 | Links authors to their institutions. |

**Semantic Similarity Edges.** To capture implicit topical connections between papers, we introduce edges between pairs of papers whose titles and abstracts exhibit high cosine similarity in the SPECTERv2 embedding space [43]. Given a pair of papers $(p_i, p_j)$ with vector embeddings $\vec{e}_i$ and $\vec{e}_j$, we compute the cosine similarity as:

$$\mathrm{sim}(p_i, p_j) = \frac{\vec{e}_i \cdot \vec{e}_j}{\|\vec{e}_i\| \cdot \|\vec{e}_j\|}$$

If this score exceeds a configurable threshold $\tau_{\text{sem}}$ an edge is formed from $p_i$ to $p_j$, with the similarity score used as the edge weight. Since cosine similarity is symmetric, we add both $(p_i \to p_j)$ and $(p_j \to p_i)$ edges, treating the relation as bidirectional. This allows the model to leverage content-level relatedness between papers, even in the absence of explicit citations, providing richer supervision for downstream node classification or link prediction tasks.

The division by the product of the vector magnitudes, $\|\vec{e}_i\| \cdot \|\vec{e}_j\|$, is a critical normalization step. Its purpose is to make the similarity score independent of the length of the document's text or the magnitude of its embedding, ensuring the comparison is based purely on semantic direction. This provides two key benefits:

- **Focus on Orientation, Not Magnitude:** In the embedding space, the "direction" of a vector represents its semantic meaning (e.g., its topics and concepts), while its "magnitude" can be influenced by factors like the length of the abstract. A very long, comprehensive paper and a short, concise paper about the same topic should be considered highly similar. Without normalization, the dot product alone would be larger for the longer paper, creating a misleadingly high similarity score. By dividing by the magnitudes, we isolate the angle between the vectors, which is a pure measure of semantic orientation.
- **Creation of a Standardized Score:** This normalization constrains the similarity score to a consistent range of [-1, 1] (or [0, 1] for non-negative embeddings like those from SPECTERv2). This makes the scores comparable across all pairs of documents in the corpus and allows for the application of a single, meaningful threshold ($\tau_{\text{sem}}$). Without this, a threshold would be arbitrary as the raw dot product scores could vary over an unbounded range.

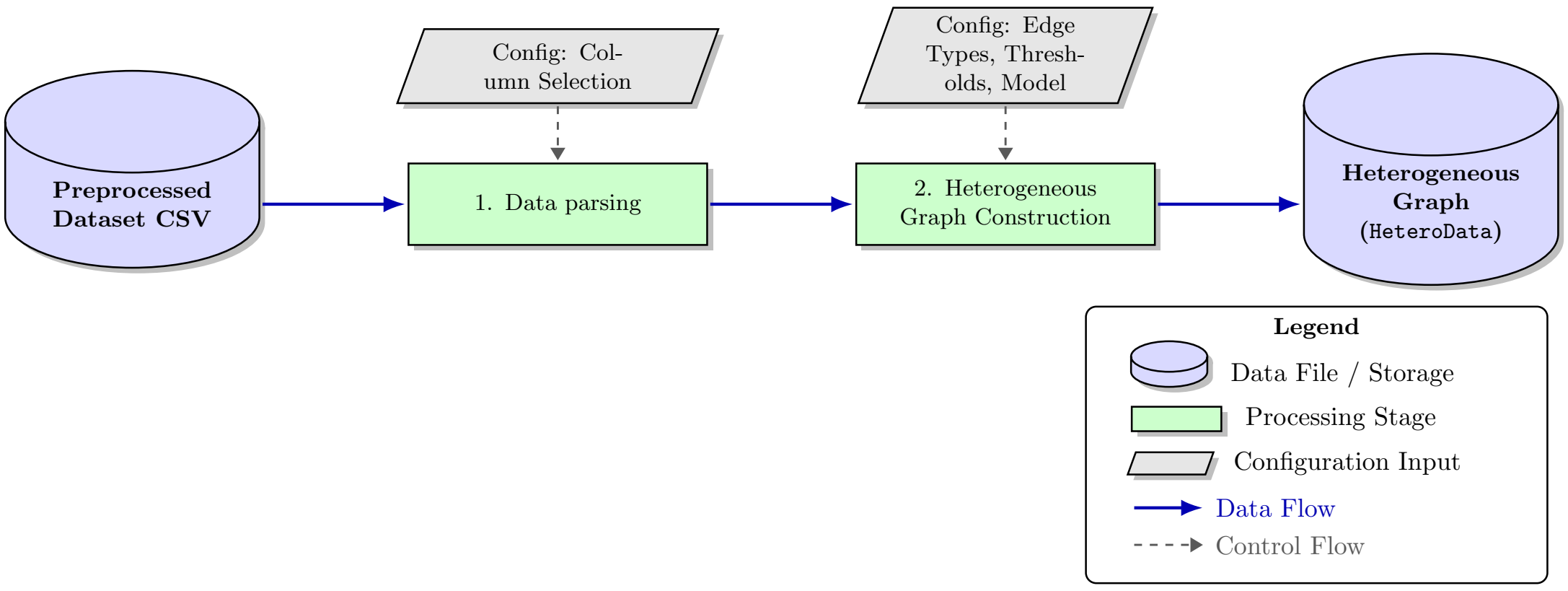

Figure 14: High-level overview of Data parsing & Heterogeneous Graph Construction stages. This two-stage transforms the preprocessed CSV into a Heterogeneous graph object ready for the active learning loop.

**Stage 3: Active Learning Loop.** This is the core stage where the iterative discovery process occurs. The system enters a loop, depicted in Figure 15, that intelligently recommends papers for annotation and retrains itself based on the user's feedback. This loop is managed by three primary components, which are described in detail in Section 5.3.

- **HAN Model:** A Heterogeneous graph Attention Network (HAN) [52], a type of GNN discussed in Section 3, serves as the core predictive engine. It learns to represent papers as feature vectors (embeddings) by passing information across the different types of relationships in the graph (e.g., citations, co-authorships). At each layer, it uses an attention mechanism[25]to weigh the importance of a paper's neighbors differently, allowing it to learn which relations are most predictive. The model is retrained from scratch on all known labels after every batch of 10 newly annotated papers.
- **Query Arm Portfolio:** The system includes a portfolio of nine distinct query strategies, or "arms." These strategies, detailed in Table 9, are drawn from established active learning literature and include methods based on model exploitation, uncertainty, diversity, and graph structure. Several of these arms, such as Embedding Diversity, directly use the learned HAN embeddings to make their selections.
- **Discounted Thompson Sampling (DTS) Agent:** The central innovation is the DTS agent, which acts as the decision-maker. At each step, this agent selects which of the nine query strategies to use next. By observing the rewards from the user's feedback, the agent learns which strategies are most effective for the current phase of the review, dynamically balancing exploration and exploitation.

**Stage 4: Visualization and Analysis.** The final module of the pipeline facilitates post-run analysis and interpretation. Its purpose is to process the detailed performance logs from the active learning loop and generate a suite of visualizations to help researchers assess the system's performance and behavior. The module can generate a variety of plots to answer research questions. For instance, a user can produce **discovery curves** to evaluate the overall recall and efficiency of a run (as shown in Appendix 25 and 26), and **diverging bar charts** to inspect the yield of each strategy at every iteration (as shown in Appendix E.2). These plots enable evaluation of the adaptive agent's effectiveness for developers and provide the quantitative backing for the results presented in this thesis. For a deeper, interactive analysis of the agent's decision-making process, the log files can be loaded into our explainability dashboard, **TS-Insight**, which is detailed in Section 5.4.

[25]The HAN contains a stack of `HeteroConv` layers from PyTorch Geometric, where each layer is composed of multiple GATConv [51] modules, one for each edge type. This allows the model to learn a separate set of attention weights for citations, semantic links, co-authorships, etc., and aggregate them to produce the final node embedding.

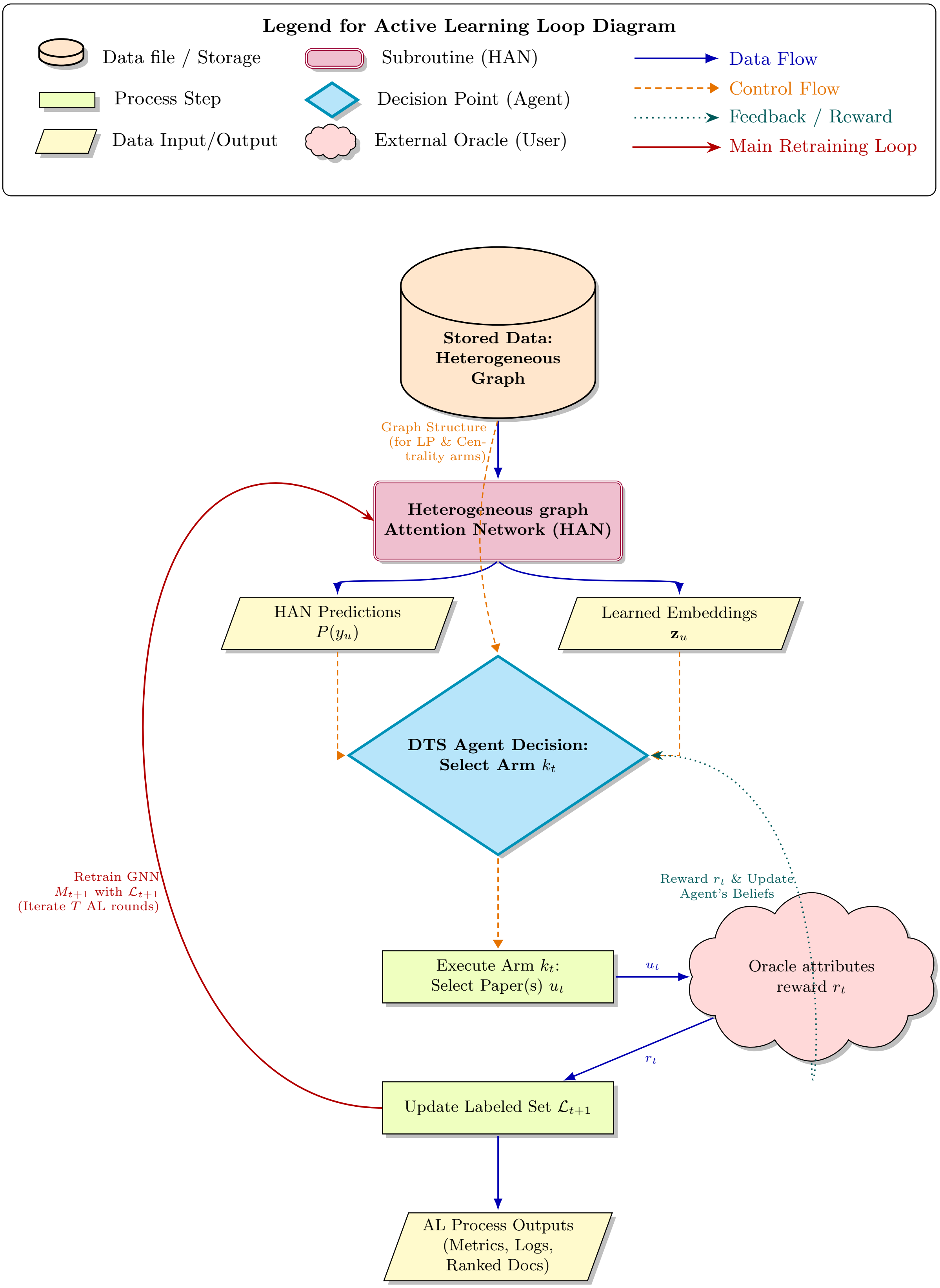

Figure 15: The synergy of the AutoDiscover Active Learning Loop. A HAN provides predictive power, while a DTS agent provides adaptive intelligence, using oracle feedback to learn the optimal query strategy and guide the iterative retraining of the graph model.

## 5.3 Discounted Thompson Sampling for Dynamic Recommendations

### 5.3.1 The Multi-Arm Bandit Framework & Query Arm Portfolio

AutoDiscover frames active learning strategy selection as an agentic decision-making problem. A Thompson Sampling agent dynamically selects query strategies (”arms”) from the following portfolio:

- **GNN Exploit:** Exploit the highest-scoring prediction of our GNN. This concept is foundational to bandit problems, as formulated by Robbins (1952), which established the basis for balancing exploration and exploitation in sequential choices [38].
- **Entropy Uncertainty:** Select the node of maximum Shannon entropy. Lewis and Gale (1994) detailed uncertainty sampling methods where class labels are requested for instances with high class uncertainty, often quantified by selecting the example with maximal Shannon entropy in its predicted distribution [27].
- **Margin Uncertainty:** Pick the node with smallest margin between the top two class scores. Tong and Koller (2001) employed a similar idea for SVMs, querying batches of instances closest to the current decision hyperplane [47].
- **BALD (Bayesian Active Learning by Disagreement):** Maximize mutual information under posterior. Houlsby et al. (2011) defined BALD as seeking the input for which model parameters under the posterior show the most disagreement about the outcome [22].
- **Embedding Diversity:** Query the unlabeled node most dissimilar in embedding space. Brinker (2003) introduced a batch-mode active learning approach incorporating a diversity measure, designed for low computational cost and suitability for large-scale problems [6].
- **Bias-Aware Uncertainty:** A hybrid strategy that prioritizes uncertain nodes that are also likely to be in the rare positive class, adapted from [12].
- **Centrality + Uncertainty (QUIRE):** Combine node-centrality with entropy. Huang, Jin, and Zhou (2010) proposed QUIRE, a principled min-max strategy that integrates informativeness (uncertainty) with representativeness (centrality), addressing a common trade-off in active learning where methods typically focus on one or the other [24].
- **Label Propagation:** Exploit graph structure via fast LP. Zhou et al. (2003) discussed semi-supervised learning through label propagation on graphs, presenting a simple algorithm that enforces local and global consistency to effectively leverage unlabeled data for smooth solutions [59].

Each strategy leverages different information sources to select the next node(s) for labeling. Table 9 provides an overview of the motivation and context for each arm, while Section B[26] details the specific selection strategies derived from our implementation. The Thompson Sampling agent adaptively learns[27] which arm provides the most utility at different stages of the active learning process. The selection of these specific query strategies was guided by a principle of balancing diversity, simplicity, and established effectiveness. Our goal was to create a comprehensive yet manageable portfolio that covered the fundamental axes of active learning: exploitation, uncertainty, diversity, and graph-structural information. Each arm represents a well-understood and computationally tractable strategy from the literature.

[26]The detailed operational principles, formal strategies, and algorithmic implementations for each of these nine arms are provided in Appendix A for completeness and reproducibility.

[27]A key component of AutoDiscover is the use of **Discounted Bayesian Updates** (Eqs. 3–4) within the Thompson Sampling framework to handle the non-stationary reward dynamics inherent in active learning, ensuring continued adaptivity [20, 37, 39].

Table 9: AutoDiscover Query Arms: Motivation and Context

| Arm Name | Potential Benefit / Motivation | References | Type |
|---|---|---|---|
| **GNN Exploit** | Effective late-stage refinement when model is well-calibrated. Risky early. | [38] | Greedy Exploitation |
| **Entropy Uncertainty** | General uncertainty reduction, useful early/mid-stage. Not biased to rare class. | [27] | Uncertainty |
| **Margin Uncertainty** | Targets samples near decision boundary, useful mid-stage. Not biased to rare class. | [47] | Uncertainty |
| **BALD** | Theoretically robust uncertainty sampling, effective early/mid-stage. Not biased to rare class. | [22] | Uncertainty |
| **Embedding Diversity** | Promotes feature space exploration, useful early/mid-stage to avoid redundancy. | [6] | Diversity |
| **Bias-Aware Uncertainty** | Particularly suited for imbalanced datasets (like SR screening) mid-stage. | [12]* | Hybrid (Unc./Bias) |
| **Centrality + Uncertainty** | Focuses uncertainty sampling on potentially influential nodes in the graph. | [24]* | Hybrid (Struct./Unc.) |
| **Label Propagation** | Effective early-stage (few labels) if structure (citations, semantic) is informative. | [59]* | Greedy Exploitation |

*Note:* * = Adapted from the original work.

#### 5.3.2 Active Learning Loop

A key contribution is the synergistic integration of graph learning with an adaptive agent for query strategy selection. We achieve this by incorporating discounted Bayesian updates within a Thompson Sampling (TS) framework, a mechanism designed to handle the inherent non-stationarity of the active learning process. As the GNN learns and the set of labeled nodes expands, the relative utility of different query strategies changes. Standard TS assumes stationary rewards, which can lead to premature convergence on arms that were effective early but become suboptimal later. To address this, our discounted approach incorporates a mechanism similar to that proposed by Raj and Kalyani (2017), who introduced a discount factor to systematically diminish the influence of past observations [37]. Likewise, the DTS agent from Qi et al. (2023) was specifically designed to adapt to changing environments by using a decay factor in its posterior updates [36].

Our AL loop operates as follows, with each step illustrated in Figure 15:

1. At each step $t$, the Thompson Sampling (TS) agent updates its belief about the expected reward of each arm $k$. As established in the Background section, this belief is maintained as a Beta distribution, where $\alpha_k$ and $\beta_k$ are pseudo-counters for the observed successes (relevant documents found) and failures (irrelevant documents found) for that arm, respectively. We initialize these parameters as $\alpha_k = 1$ and $\beta_k = 1$ for all arms, which corresponds to a uniform Beta(1,1) prior. This standard, uninformative prior is chosen because it ensures the agent begins with no preconceived bias towards any particular strategy, allowing the observed evidence to guide its decisions entirely. The agent's belief about an arm's reward probability[28] is therefore represented by $\theta_k \sim \text{Beta}(\alpha_k, \beta_k)$.

2. To decide the next action, the agent samples a probability of reward from each arm's posterior:

$$\theta_k \sim \text{Beta}(\alpha_k, \beta_k)$$

and selects the arm with the highest probability:

$$k_t = \arg\max_k \theta_k$$

3. The selected query strategy, $k_t$, is then applied to the pool of unlabeled data to recommend a paper node, $u_t$, for annotation by the oracle.

[28] As defined in Section 3, $\theta_k$ represents the true, unknown probability of success (i.e., finding a relevant document) for a given query strategy (arm) $k$.

4. Upon receiving the true label for the queried paper node, the agent observes a binary reward, $r_t \in \{0, 1\}$, where $r_t = 1$ if the paper is relevant and $r_t = 0$ otherwise. This reward is then used to update the posterior parameters of the selected arm via **discounted Bayesian updates**:

$$\alpha_{k_t}^{\text{post}} = \gamma\, \alpha_{k_t}^{\text{prior}} + r_t, \tag{3}$$

$$\beta_{k_t}^{\text{post}} = \gamma\, \beta_{k_t}^{\text{prior}} + (1 - r_t). \tag{4}$$

Here, $\gamma \in (0, 1]$ is a **discount factor** that governs the extent of memory decay. A value of $\gamma < 1$ introduces a form of **exponential forgetting**, allowing the agent to adapt to evolving utility of arms throughout the non-stationary active learning process[29].

5. The heterogeneous GNN is then retrained on the expanded labeled set, propagating information throughout the graph structure to update both predictions and embeddings.

This agentic design, particularly the discounted TS mechanism, allows the system to autonomously adopt, navigating the exploration–exploitation trade-off without manual intervention.[30]
The theoretical justification for this DTS-driven adaptive selection is rooted in regret minimization frameworks for non-stationary multi-armed bandits (e.g., [4, 18, 37]). As formalized in Section C (Theorem C.1), these frameworks establish that under assumptions of bounded variation in arm reward means, a condition reflective of the evolving AL environment where model accuracy and unlabeled pool composition change, DTS can achieve an expected cumulative regret that grows sub-linearly with the number of queries. This implies that the agent efficiently learns to adapt to changes in arm utilities, minimizing the selection of suboptimal query strategies over time. Consequently, this adaptivity is hypothesized to provide superior average performance and robustness across a diverse suite of datasets compared to any single static AL strategy, which lacks the mechanism to alter its approach when its fixed strategy becomes inefficient for a given dataset or AL phase.

## 5.4 TS-Insight: A Visual Analytics Dashboard for Explainable AI

Thompson Sampling (TS) and its discounted variant (DTS) are effective for sequential decision-making under uncertainty, but their stochasticity and non-stationary belief updates can make the controller difficult to verify, debug, and trust in practice [9,31]. To address this opacity, we developed **TS-Insight**[31], an open-source visual analytics dashboard that exposes the internal mechanics of TS-based agents at the level of individual sampling steps [49].

### 5.4.1 Design framing: users, data, and task abstraction

We structure the design rationale following standard visualization design principles: (i) characterize the *users* and their problems; (ii) formalize the *data types* produced by the algorithm; (iii) derive *low-level analytic tasks* required to accomplish higher-level XAI goals; and (iv) justify view designs and interactions relative to alternatives.

**Target users and usage context.** TS-Insight is designed primarily for **model developers and researchers** who build, tune, and validate TS/DTS controllers (e.g., for active learning strategy selection). Their core needs are to (a) confirm implementation correctness (e.g., discounting and reward updates), (b) explain specific arm choices, and (c) diagnose pathological behaviors such as premature convergence, arm starvation, or failure to adapt to non-stationarity [9].

**Data characterization ("bandit trace" data).** The dashboard consumes a *bandit execution trace* consisting of: (i) a discrete time axis $t = 1..T$ (sampling steps), (ii) a categorical arm index $k \in \{1..K\}$, (iii) per-arm **Beta posterior state** parameters $(\alpha_{k,t}, \beta_{k,t})$, (iv) derived statistics such as posterior mean $\mu_{k,t} = \alpha_{k,t}/(\alpha_{k,t} + \beta_{k,t})$ and uncertainty, (v) a sampled posterior draw $\hat{\theta}_{k,t} \sim \text{Beta}(\alpha_{k,t}, \beta_{k,t})$, (vi) the selected arm $a_t = \arg\max_k \hat{\theta}_{k,t}$, and (vii) a binary reward $r_t \in \{0, 1\}$.

[29]As noted by Raj and Kalyani, the discounting factor in their algorithm, managing the balance between remembering and forgetting, significantly influences performance [37]. This mechanism prevents overconfidence in arms that performed well in early iterations but may become less effective as the model and label distribution shift.

[30]Our full algorithmic specification of the Discounted Thompson Sampling agent, along with its formal theoretical grounding in regret minimization theory for non-stationary bandits, is provided in Appendix C.

[31]`https://github.com/LIST-LUXEMBOURG/ts-insight`

**From high-level XAI goals to low-level tasks.** TS-Insight operationalizes four analysis goals that recur in our debugging workflow and are also reflected in the tool's design summary: **Diagnosis, Verification, Explanation,** and **Reliability** [9, 31, 49]. Each goal decomposes into low-level tasks the user must execute on the trace:

- **Diagnosis ("What is the algorithm doing over time?")**: locate phases of persistent selection, switching, or starvation; identify behavioral shifts; compare arms' selection histories across time windows.
- **Verification ("Is it functioning as intended?")**: confirm that (a) $\alpha/\beta$ update matches observed rewards; (b) discounting produces expected decay when arms are idle; (c) the chosen arm is consistent with $\arg\max \hat{\theta}$.
- **Explanation ("Why did it choose arm $a_t$ at step $t$?")**: at a given step, compare (i) posterior means (exploit signal) vs. (ii) posterior draws (explore signal); determine whether the choice was exploitation- or exploration-driven.
- **Reliability ("Was this choice confident or risky?")**: detect steps where the chosen draw is a low-probability event under the arm's posterior; identify broad/unstable posteriors indicating high epistemic uncertainty.

### 5.4.2 Coordinated views and design rationale (with alternatives)

To support the above tasks, TS-Insight implements coordinated views organized as per-arm rows (small multiples) with synchronized time axes, plus a step-level comparison view (XAI Snapshot). This layout prioritizes: (i) cross-arm comparison without overplotting, (ii) temporal reasoning for non-stationarity, and (iii) direct access to the algorithm's internal state variables.

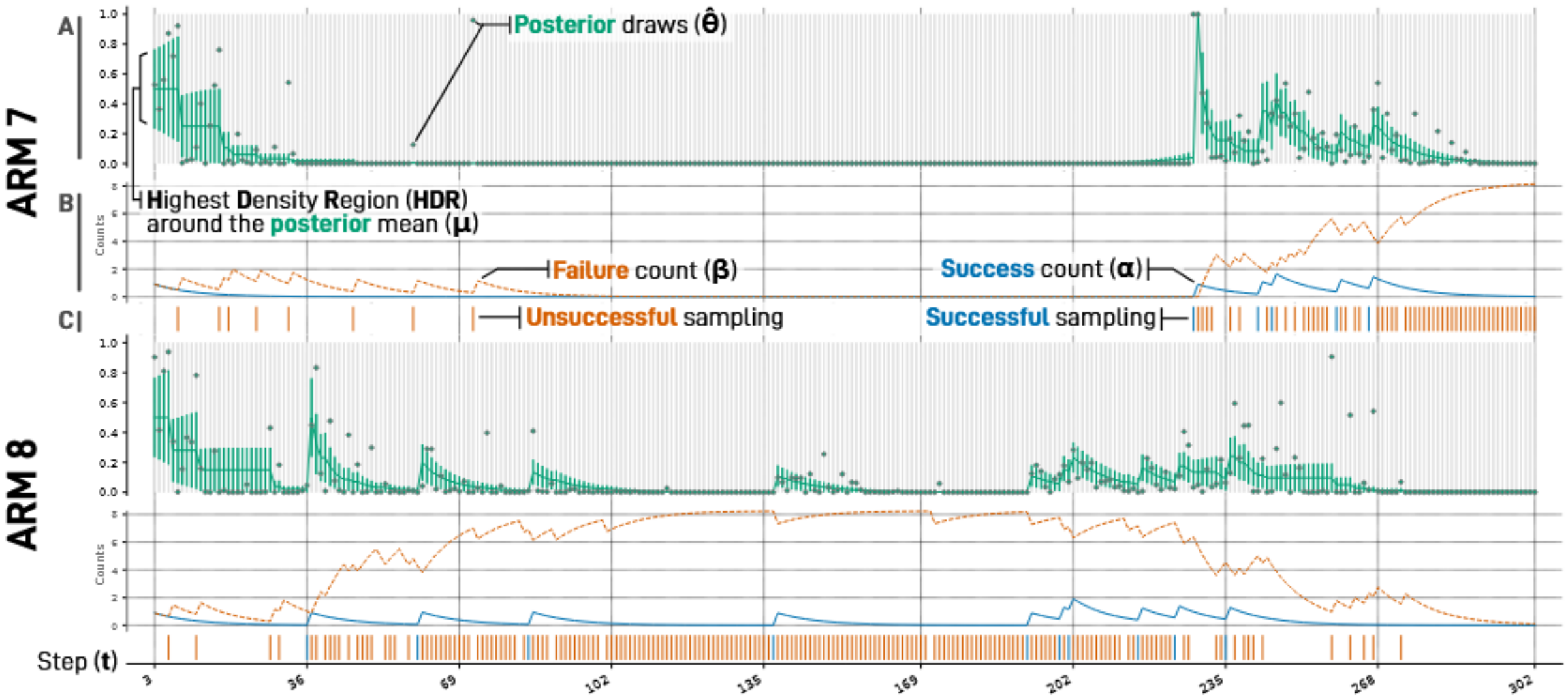

Figure 16: Excerpt from the TS-Insight dashboard, illustrating the behavior of multiple arms over time. Each arm's evolution is decomposed into three subplots: **(A)** reveals the algorithm's belief and uncertainty about an arm's success rate. **(B)** shows the raw evidence accumulation. **(C)** provides a compact history of when an arm was chosen and its outcome. Arms and subplots can be displayed or hidden at will.

**HDR Evolution plot (belief + uncertainty over time).** To visualize belief dynamics, TS-Insight uses a Highest Density Region (HDR) encoding of the Beta posterior at each step (Figure 16A). Concretely, the plot represents:

- **x-axis:** time step $t$,
- **y-axis:** reward probability in $[0, 1]$,

- **HDR band:** an interval containing a fixed probability mass of the posterior (e.g., a 95% HDR), i.e., the *most probable* region of $\theta$ under Beta$(\alpha_{k,t}, \beta_{k,t})$,
- **posterior mean:** a reference marker/curve at $\mu_{k,t}$,
- **posterior draw:** a point for $\hat{\theta}_{k,t}$ (and especially the chosen arm's draw).

This design exposes diagnostic patterns that directly support the stated goals: (i) **narrowing HDR bands** indicate increasing certainty as evidence accumulates, (ii) **HDR drift upward/downward** indicates changing expected reward, and (iii) a draw **outside the 95% HDR** is, by definition, a *low-posterior-density event* (rare under current belief), which visually flags a high-uncertainty/exploratory decision and motivates reliability inspection.
*Alternatives considered.* A standard alternative is a *credible interval* (equal-tailed quantiles) band or a ridgeline/violin density plot over time. We prefer HDR because it encodes the *highest-probability mass* region (not merely symmetric quantiles), making "out-of-region" samples more semantically aligned with the notion of *low-density, risky* decisions, which is central to our reliability goal.

**Alpha/Beta evidence plot (mechanics + discounting verification).** The Alpha/Beta subplot (Figure 16B) displays the internal evidence counters over time as two aligned time series: $\alpha_{k,t}$ (success evidence) and $\beta_{k,t}$ (failure evidence). This view is intentionally "white-box": rather than showing only outcomes, it reveals the controller's *state variables* that generate posterior means and variances. It supports verification tasks such as: (a) checking that a success increments $\alpha$ (and a failure increments $\beta$), (b) confirming DTS discounting as a **visible decay** of $\alpha/\beta$ during intervals where the arm is not pulled, and (c) diagnosing parameter regimes (e.g., too-aggressive discounting) when evidence decays faster than it can stabilize.
*Alternatives considered.* A common alternative is plotting only empirical success rate or moving-average reward. We use $\alpha/\beta$ because it is the direct object manipulated by the algorithm; therefore it minimizes inference distance for debugging (users verify the update logic itself, not only its aggregated effects).

**Barcode history plot (event sequence + regime shifts).** The Barcode view (Figure 16C) encodes each arm's selection events as a 1D timeline: at step $t$ a stroke is drawn if the arm was selected, and its color encodes reward (success/failure). This supports diagnosis by revealing patterns such as: (i) **bursts** of repeated selection (possible exploitation lock-in), (ii) **long gaps** (arm starvation), and (iii) **phase changes** where the selected arm distribution shifts, suggesting non-stationarity or controller adaptation.
*Alternatives considered.* Two typical alternatives are (a) stacked area charts of selection frequencies or (b) a heatmap of arm-by-time selections. The barcode design emphasizes *exact event timing and ordering* with minimal clutter, which is valuable for step-level linking to explanation views (below). This can be evaluated empirically (see proposed study below).

**XAI Snapshot view (step-level explanation of exploration vs. exploitation).** For "why did it choose this arm now?", TS-Insight provides an XAI Snapshot view (Figure 17) that compares all arms at a selected time step $t$. The view overlays two quantities per arm:

- **posterior mean** $\mu_{k,t}$ (exploitation signal; long-run belief),
- **posterior draw** $\hat{\theta}_{k,t}$ (exploration driver; instantaneous sample used for selection).

The chosen arm is the one with maximal $\hat{\theta}_{k,t}$. Hence, the view makes the exploration–exploitation mechanism visually explicit: if $a_t$ is also the argmax of $\mu_{k,t}$, the step is exploitation-driven; otherwise, it is exploration driven by variance (an unusually high draw). This direct decomposition supports local explanation without requiring users to mentally combine information across multiple plots.
*Alternatives considered.* Possible alternatives include a tabular readout, a slopegraph (mean-to-draw), or a paired-dot plot. We use bar+marker because it supports rapid rank comparison and clear separation between "belief" (mean) and "sample" (draw), which correspond to distinct causal factors in TS selection.

**Coordinated view (supporting trace-to-explanation workflows).** A key design choice is **coordination** across views: users can locate a suspicious event pattern in the barcode (e.g., a burst or failure streak), then inspect its belief trajectory (HDR plot), verify the mechanical updates ($\alpha/\beta$), and finally open a step-level explanation (XAI Snapshot). This supports a common debugging workflow: *overview − > locate − > explain − > verify.*

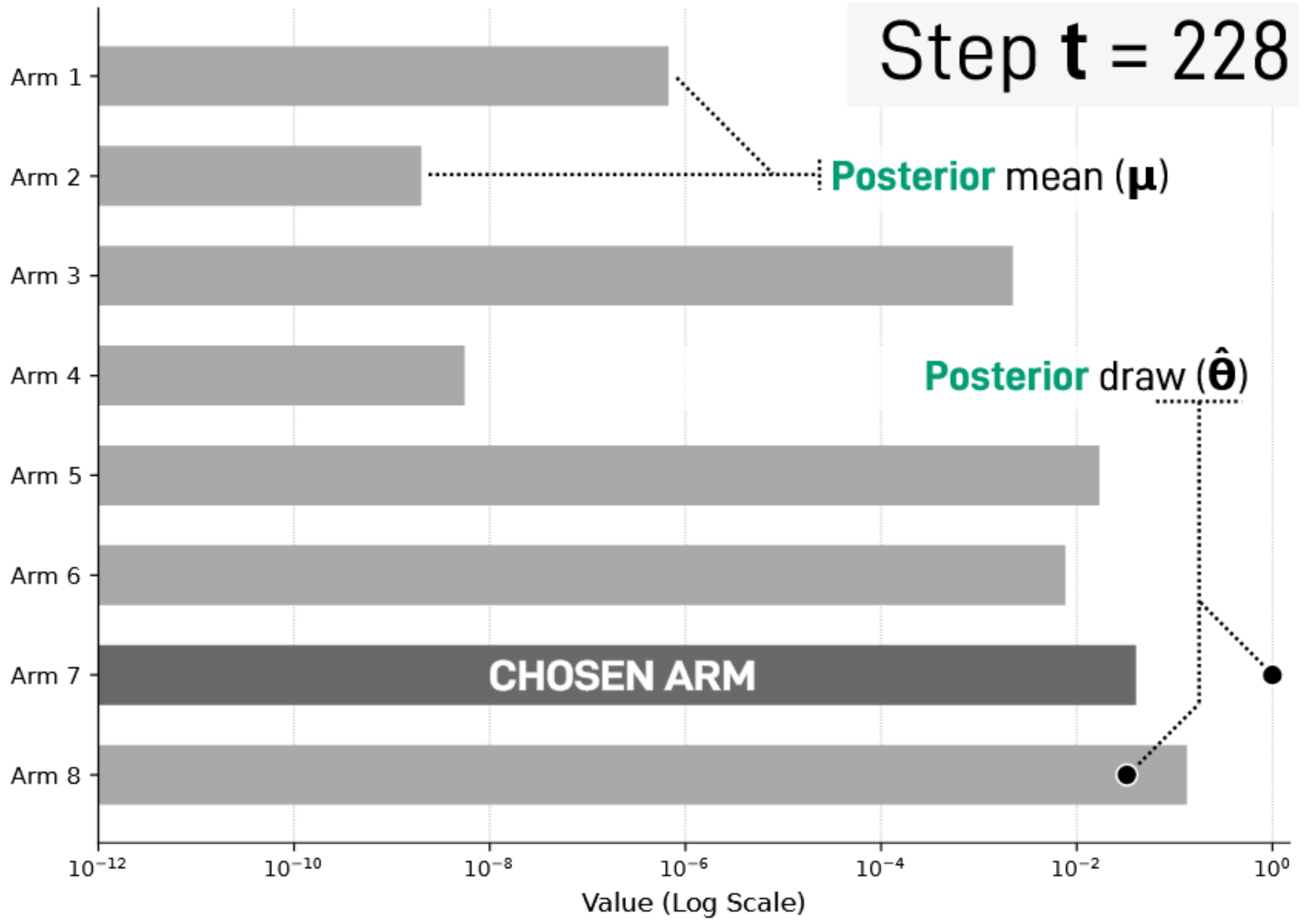

Figure 17: The XAI Snapshot view explains the algorithm's choice at a single sampling step. Here, one can see that although Arm 8 has the highest posterior mean, Arm 7 was chosen because its random posterior draw was the highest, clearly illustrating an **Exploration** step.

#### 5.4.3 Evaluation hypotheses (task efficiency claims) and study sketch

The above design choices imply measurable hypotheses about user performance for TS/DTS debugging tasks. We propose the following hypotheses for a controlled study against baselines such as numerical logs and aggregate regret curves:

- **H1 (Verification efficiency):** Users identify implementation errors (e.g., wrong discounting or incorrect $\alpha/\beta$ updates) faster and with higher accuracy using TS-Insight than with log-only baselines.
- **H2 (Explanation accuracy):** Users more accurately classify decisions as exploration vs. exploitation at a given step $t$ using the XAI Snapshot than with posterior-mean-only plots.
- **H3 (Reliability/diagnosis):** Users detect non-stationary regime shifts and arm starvation earlier using the Barcode+HDR combination than with aggregated frequency summaries.

A suitable experiment design is a within-subject study where participants complete timed tasks aligned to the goal taxonomy above (verification, explanation, diagnosis, reliability), with dependent variables including completion time, error rate, and subjective workload (e.g., NASA-TLX). This evaluation would quantify whether the proposed visual encodings improve trace comprehension and debugging effectiveness, thereby operationalizing the dashboard's intended contribution beyond qualitative design claims.

# 6 Evaluation

Having detailed the architecture of the AutoDiscover framework in the preceding chapter, this section outlines the rigorous protocol designed to evaluate its performance and validate our core hypotheses. To ensure a comprehensive and fair assessment, we define the specific metrics used to measure screening efficiency, the established baseline models against which AutoDiscover is compared, the hyperparameter optimization strategy employed to elicit the best performance from our system, and the computational environment that ensures the reproducibility of our results. This protocol provides the foundation for the empirical findings presented in Section 7.

## 6.1 Evaluation Metrics

To provide a multi-faceted assessment of model performance, we employ a suite of metrics that capture overall efficiency, high-recall workload reduction, and early-stage discovery effectiveness. The following metrics are used in our evaluation:

- **Discovery Rate Efficiency (DRE):** Measures overall screening efficiency relative to random.
- **Work Saved over Sampling (WSS@p):** Quantifies the workload reduction at 95% and 80% recall.
- **Recall at k (R@k):** Measures recall after screening a fixed number of documents (200 and 500).

For DRE, WSS, and R@k, higher values are better.

### 6.1.1 Discovery-Rate Efficiency (DRE)

Let

$$\begin{aligned} D &= \text{number of relevant items discovered (incl. seed positives)}, \\ P &= \text{total relevant items}, \\ Q &= \text{total items screened}, \\ N &= \text{total items in dataset}. \end{aligned}$$

Then

$$\text{DRE} = \frac{D/P}{Q/N} = \frac{D\,N}{P\,Q}. \tag{5}$$

Equivalently, defining percentages

$$R\% = 100 \cdot \frac{D}{P}, \qquad S\% = 100 \cdot \frac{Q}{N},$$

we have

$$\text{DRE} = \frac{R\%}{S\%}.$$

### 6.1.2 Recall@$k$

The recall after screening exactly $k$ items is

$$\text{Recall@}k = \frac{D(k)}{P}, \tag{6}$$

where $D(k)$ is the number of relevant items found in the first $k$ screened [32].

### 6.1.3 Work Saved over Sampling at $p\%$ (WSS@$p$)

To reach $p\%$ recall under random sampling, one would expect to screen $(p/100)\,N$ items. Let $q_p$ be the actual number of items screened by our system to achieve $p\%$ recall of the total $P$ relevant items (i.e., to find $(p/100)P$ relevant items). Then

$$\text{WSS@}p = 1 - \frac{q_p}{(p/100)\,N} = \frac{(p/100)\,N - q_p}{(p/100)\,N}. \tag{7}$$

[32] In our implementation, for a given $k$ (e.g., 200 or 500), we find the point on the discovery curve corresponding to $k$ items screened

## 6.2 Baseline Models

To rigorously evaluate the performance of AutoDiscover, we compare it against two categories of baselines: a theoretical random baseline and a suite of strong, static active learning models derived from the literature. These baselines were chosen to represent both a no-information reference point and the performance of established, non-adaptive AL techniques commonly used in systematic literature reviews.

1. **Random Sampling:** This serves as the fundamental baseline, representing the performance expected from screening documents in a completely random order, as if a reviewer were not using any prioritization tools and simply worked through every title and abstract one by one until their own study is encountered. Its theoretical performance is a Discovery Rate Efficiency (DRE) of 1.0 and a Work Saved over Sampling (WSS) of 0.0. The expected recall after screening $k$ documents is $k/N$, where $N$ is the total number of documents in the dataset. This baseline helps quantify the value added by any intelligent prioritization strategy.

2. **Recommended Static Active Learning Models:** We benchmark AutoDiscover against two high-performing static active learning models from the comprehensive simulation study by Teijema et al. (2025) [45]. Their work evaluated a wide range of models on the same SYNERGY benchmark datasets used in this thesis. Based on their findings and the models' established robustness, we selected the following two configurations as our primary baselines:

   - Naive Bayes classifier with TF-IDF features (**NB + TF-IDF**).
   - Logistic Regression classifier with TF-IDF features (**LR + TF-IDF**).

   Performance data for these baselines was derived from the publicly available raw simulation results of their "Simulation Study 2" on DataverseNL [44]. For our analysis, we processed their time-to-discovery data to reconstruct discovery curves and calculate our target metrics (DRE, WSS@P, R@k), averaging results across all simulation runs for each dataset-model pair.

   Crucially, the experimental setup in Teijema et al. used an initial training set of 5 relevant and 10 irrelevant records, providing a well-initialized but still challenging starting point for the models. This makes their results a strong and relevant benchmark for evaluating the performance of AutoDiscover.

## 6.3 Hyperparameter Optimization

To ensure a rigorous and fair comparison of AutoDiscover's potential against established baselines, and acknowledging that optimal configurations can be dataset-specific, we adopted a per-dataset hyperparameter optimization (HPO) strategy. This mirrors the methodology employed in comprehensive benchmarks of systems like ASReview, where, for instance, Ferdinands et al. [14] utilized HPO with the Tree-structured Parzen Estimator (TPE) [53] sampler to maximize model performance independently for each dataset in their large-scale simulations.

For each of the 26 SYNERGY datasets, we conducted 80 HPO trials for AutoDiscover using Optuna [1] with the TPE sampler. To select the single best-performing configuration from these trials in a deterministic and multi-faceted way, we defined a **hierarchical optimization objective**. The best trial was selected based on the following prioritized criteria:

1. Highest Overall Discovery Rate Efficiency (**DRE**).
2. *(Tie-breaker)* Highest Work Saved over Sampling at 95% recall (**WSS@95**).
3. *(Tie-breaker)* Highest Work Saved over Sampling at 80% recall (**WSS@80**).
4. *(Tie-breaker)* Highest Recall at 500 documents (**R@500**).
5. *(Tie-breaker)* Highest Recall at 200 documents (**R@200**).

This hierarchical approach ensures that we first select for the most efficient models overall (DRE), and then use progressively finer-grained recall and work-saved metrics to break any ties, providing a robust method for identifying the top-performing configuration. The hyperparameter search space, encompassing GNN architecture, graph construction parameters, and AL/DTS settings, is detailed in Table 10.

**Rationale for Number of HPO Trials.** The hyperparameter search space for AutoDiscover is exceptionally large. A conservative estimate based only on the discrete and stepped parameters (ignoring the continuous log-uniform ranges for learning rate, weight decay, and the discount factor) reveals approximately 278 million unique combinations. The true cardinality of the search space is effectively infinite due to the continuous parameters.
Given this vastness, an exhaustive search is computationally impossible. The choice of 80 trials per dataset for our HPO represents a pragmatic balance between the desire for thorough exploration and the computational cost of each trial. While 80 trials cannot guarantee finding the global optimum in such a large space, we rely on the sample efficiency of the Tree-structured Parzen Estimator (TPE) algorithm [53]. Unlike random search [3], TPE builds a probabilistic model of the objective function and uses it to intelligently select more promising hyperparameter configurations in subsequent trials. As demonstrated by our optimization history plots (Figures 20 and 21), the TPE sampler was able to identify high-performing regions, suggesting that our approach, while constrained by resources, was adequate to find robust and effective configurations for comparison.

Table 10: Hyperparameter Search Space for Optuna Tuning.

| **Component** | **Hyperparameter** | **Search Range / Values** |
|---|---|---|
| GNN Model | Hidden Channels (`gnn_hidden_channels`) | Integer: [32, 256], step 32 |
| | Number of Layers (`gnn_num_layers`) | Integer: [1, 5] |
| | Dropout (`gnn_dropout`) | Float: [0.1, 0.7], step 0.1 |
| | Learning Rate (`gnn_lr`) | LogUniform Float: [5e-4, 1e-2] |
| | Weight Decay (`gnn_weight_decay`) | LogUniform Float: [1e-6, 1e-3] |
| Graph Construction | Semantic Threshold (`semantic_threshold`) | Float: [0.88, 0.98], step 0.01 |
| | Collaboration Threshold (`collaboration_threshold`) | Fixed: 2 |
| | Embedding Model (`embedding_model_name`) | Fixed: `allenai/specter2_base` |
| Active Learning Core | Epochs per Iteration (`al_epochs_per_iter`) | Integer: [50, 200], step 50 |
| | BALD MC Samples (`al_bald_samples`) | Integer: [10, 30], step 5 |
| | LP Arm Layers (`al_lp_layers`) | Integer: [2, 8], step 2 |
| | LP Arm Alpha (`al_lp_alpha`) | Float: [0.5, 0.95], step 0.05 |
| Focal Loss (Conditional) | Use Focal Loss (`al_use_focal_loss`) | Categorical: [True, False] |
| | Gamma (`al_focal_gamma`) | Float: [0.5, 4.0], step 0.5 |
| | Alpha (`al_focal_alpha`) | Float: [0.1, 0.75], step 0.05 |
| Discounted TS (Conditional) | Use Discounted TS (`al_use_ts_forgetting`) | Fixed: [True] |
| | Discount Factor (`al_ts_discount`) | LogUniform Float: [0.88, 1.0] |

## 6.4 Computational Environment

The experiments in this thesis, particularly the training of Graph Neural Networks on large, dense scholarly graphs, are computationally demanding. They require significant memory to hold the graph structure and embeddings, as well as substantial parallel processing power for GNN message passing and hyperparameter optimization. To meet these demands, all experiments were conducted on a dedicated high-performance computing server.
To ensure a consistent and reproducible research environment, all code was executed within a Docker container, managed via Portainer and developed in Visual Studio Code. This containerized approach encapsulates all software dependencies and system configurations. The key specifications of the hardware and software environment are detailed in Table 11.

Table 11: Hardware and Software Specifications of the Experimental Environment.

| **Component** | **Specification** |
|---|---|
| **Hardware** | |
| GPU | NVIDIA Tesla V100 (PCIe) |
| GPU Memory (VRAM) | 32 GB HBM2 |
| GPU Power Limit | 250 W |
| System Memory (RAM) | ~512 GB |
| **Software & Execution Environment** | |
| Operating System | Linux (within a Docker container) |
| NVIDIA Driver | Version 470.256.02 |
| CUDA Toolkit Version | 12.4 |
| Core Libraries | Python, PyTorch, PyTorch Geometric [15] |
| Language Models | Hugging Face Transformers [54] |
| Visualization Dashboard | Streamlit |

The configuration of a high-memory GPU (32 GB VRAM) was crucial for accommodating the largest graphs from the SYNERGY benchmark, such as `Brouwer_2019`, without encountering out-of-memory errors during GNN training. The large amount of system RAM was similarly essential for pre-processing steps, including the initial loading and manipulation of the 169,000-record combined dataset. This robust computational foundation ensures that the experimental results presented are a true reflection of the models' capabilities, unconstrained by hardware limitations.

# 7 Results and Discussion

This section presents a comprehensive empirical evaluation of AutoDiscover's performance on the 26 datasets of the SYNERGY benchmark. Having established the characteristics of the datasets and our experimental methodology, we now turn to the results of our 70-trial per-dataset hyperparameter optimization. The findings are presented first from a high-level, visual perspective to identify macro-level trends, followed by a detailed quantitative comparison against state-of-the-art static baseline models.

## 7.1 Overall Performance Across the SYNERGY Benchmark

To provide a consolidated overview of AutoDiscover's peak performance, we first visualize the key metrics from the best-performing trial for each of the 26 datasets in a comprehensive heatmap (Figure 18). The datasets in this figure are sorted in descending order of their primary optimization metric, the Overall Discovery Rate Efficiency (DRE). This arrangement allows for an immediate visual assessment of performance patterns and their potential correlation with underlying dataset characteristics. A DRE score can be interpreted as a multiplier of efficiency compared to random screening; for instance, a DRE of 17.5 means the system was 17.5 times more efficient at finding relevant papers than a manual, random search.

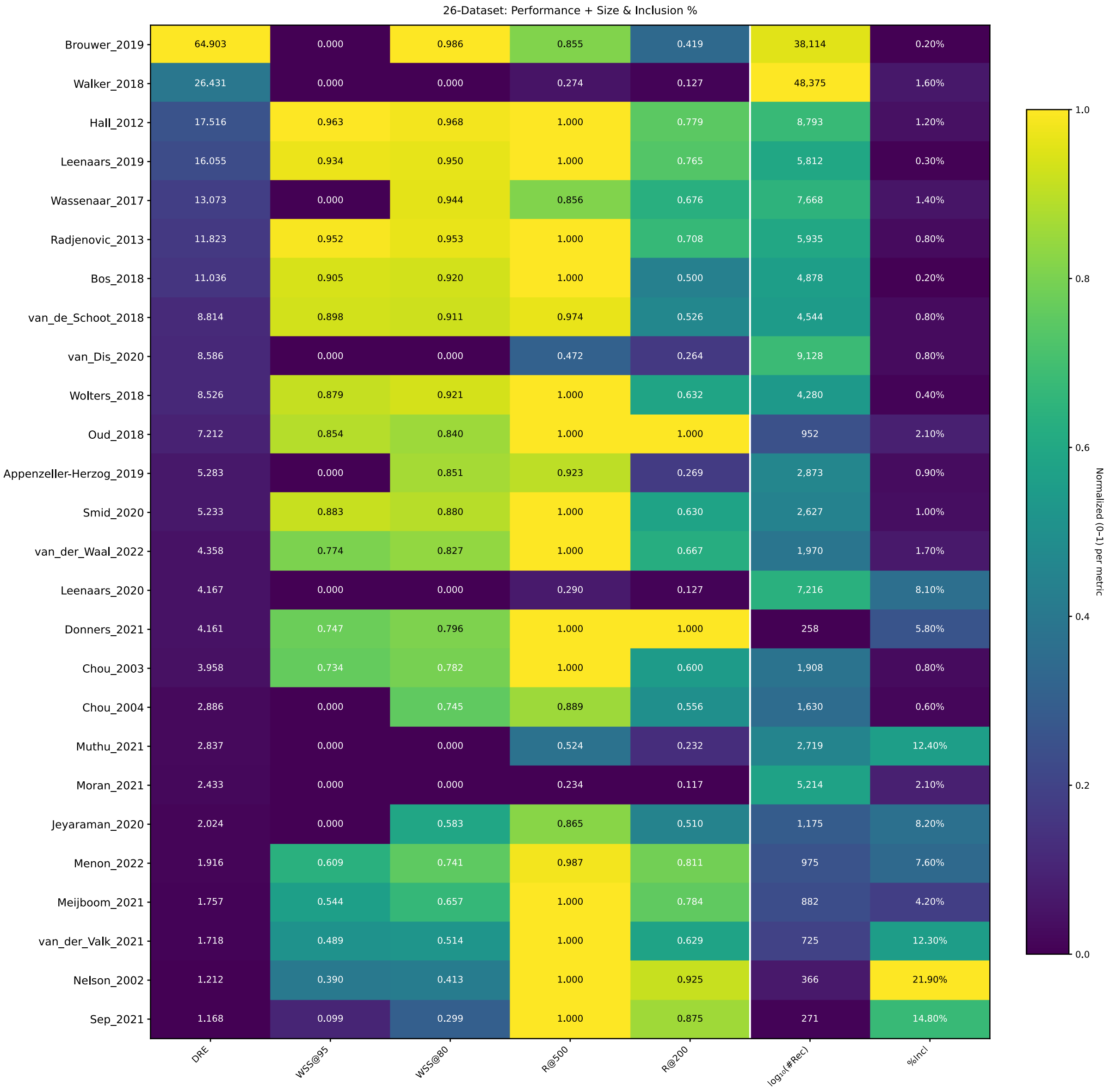

Figure 18: Performance of AutoDiscover (best HPO configuration per dataset) across 26 systematic literature review datasets, sorted by Discovery Rate Efficiency (DRE). The heatmap visualizes key performance metrics alongside dataset characteristics. Colors are normalized column-wise to show relative performance within each metric, while cell annotations provide the exact raw values. DRE: Discovery Rate Efficiency; WSS@$p$: Work Saved over Sampling at $p$% recall; R@$k$: Recall after $k$ papers screened; #Rec: total records; %Incl: inclusion rate.

## 7.2 Distributional Analysis of Performance Metrics

While the heatmap provides a per-dataset view, a distributional analysis summarizes the overall behavior of AutoDiscover. Figure 19 presents boxplots for our key metrics, illustrating the central tendency and variance in performance across the entire SYNERGY benchmark.

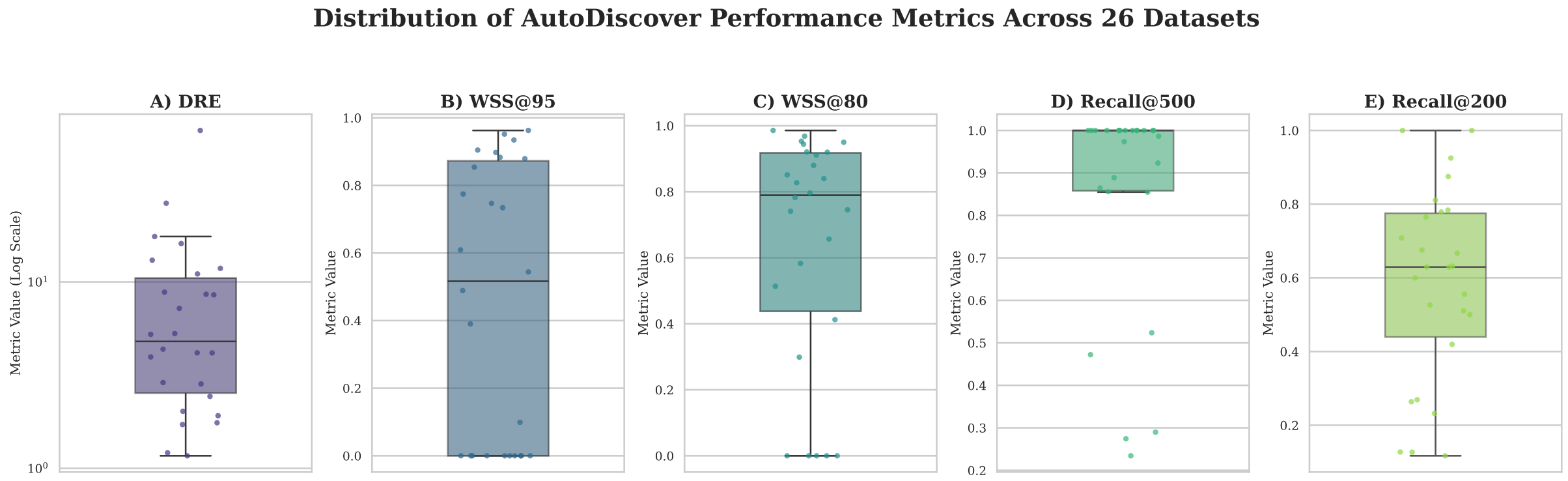

Figure 19: Distribution of AutoDiscover's peak performance across 26 datasets for five key metrics. Each point represents the result from one dataset. The boxplot shows the median (center line), interquartile range (box), and full range (whiskers). (A) Discovery Rate Efficiency (DRE) is plotted on a log scale due to its wide, skewed distribution. (B-E) Work Saved over Sampling (WSS) and Recall metrics are plotted on a linear scale from 0 to 1.

Table 12: Summary Statistics of AutoDiscover's Performance Metrics Across All 26 Datasets. This table presents the central tendency and range of performance for the best-performing trial on each dataset.

| Statistic | DRE | WSS@95 | WSS@80 | R@500 | R@200 |
|---|---|---|---|---|---|
| **Max** | 64.903 | 0.963 | 0.986 | 1.000 | 1.000 |
| **Min** | 1.168 | 0.000 | 0.000 | 0.234 | 0.117 |
| **Mean** | 9.196 | 0.448 | 0.634 | 0.852 | 0.582 |
| **Median** | 4.796 | 0.517 | 0.790 | 1.000 | 0.629 |

Table 12 quantifies the distributional performance of AutoDiscover. The median Discovery Rate Efficiency (DRE) of 4.80 indicates that our system is typically nearly five times more efficient than random screening. The performance on Work Saved over Sampling (WSS) highlights the trade-offs of our optimization strategy: the median WSS@80 is a robust 0.79, showing high efficiency in finding the majority of relevant documents. In contrast, the median WSS@95 is lower at 0.52, with a minimum of 0.00, reflecting that achieving the final few percentage points of recall was not the primary optimization target. Finally, the recall metrics show that within a fixed budget of 500 documents, a median of 100 % of relevant documents are found, demonstrating the system's high practical effectiveness.

**(A) Discovery Rate Efficiency (DRE):** The DRE distribution is plotted on a logarithmic scale due to its wide range and right-skew, with values spanning from approximately 1.07 to over 64.9. The median DRE is approximately 4.8, indicating that AutoDiscover is, for a typical dataset, nearly five times more efficient than random screening. The wide interquartile range demonstrates that while the system always outperforms random, its performance advantage varies significantly depending on dataset characteristics, reinforcing the need for per-dataset optimization.

**(B, C) Work Saved over Sampling (WSS@95 and WSS@80):** The WSS metrics show a different pattern. The median WSS@95 is approximately 0.52, but the distribution is heavily skewed towards the lower end, with a significant number of datasets scoring 0.00. This confirms the trade-off identified in the heatmap: our HPO strategy, optimized for DRE, often sacrifices performance on the high-recall endgame. In contrast, the median WSS@80 is much higher at approximately 0.79, with a tighter and more optimistic distribution. This indicates that AutoDiscover is highly robust at finding the first 80 % of relevant documents, the stage that often constitutes the bulk of the screening effort.

**(D, E) Early Recall (R@500 and R@200):** The recall metrics provide insight into the model's performance within a fixed screening budget. At 500 documents screened (Panel D), performance is exceptional, with a median recall near 1.00. This shows that for most datasets, virtually all relevant items are found within this initial budget. The distribution for Recall@200 (Panel E) is more varied, with a median of approximately 0.63. This remains a strong result, indicating that, on average, well over half the relevant documents are found after screening just 200 candidates. The variance in this metric suggests that the speed of early discovery depends heavily on dataset characteristics like prevalence and graph structure.

**Overall Conclusion from Distributional Analysis:** Collectively, these visualizations paint a clear picture of AutoDiscover as a powerful and highly efficient system, particularly when evaluated on overall efficiency (DRE) and majority-recall (WSS@80) metrics. The distributions highlight its robustness across a wide range of academic domains and data characteristics. The observed trade-offs, especially regarding WSS@95, are not flaws but direct consequences of our optimization objective, pointing towards promising avenues for future work in multi-objective or adaptive-objective active learning.
The heatmap reveals several critical insights. First, the wide range of DRE values, from over 64 on the `Brouwer_2019` dataset to lower, albeit still effective, values on others, underscores that a "one-size-fits-all" model is insufficient; per-dataset optimization allows AutoDiscover to adapt to these diverse environments.
Second, a notable trade-off emerges between the DRE and the Work Saved over Sampling at 95% recall (WSS@95). Several datasets with extremely high DRE scores (e.g., `Brouwer_2019`, `Walker_2018`) exhibit a WSS@95 of 0.00. This occurs because our HPO objective (maximizing DRE) prioritizes finding a large fraction of relevant documents as early as possible. This may lead to strategies that excel at rapid, early discovery but do not reach the very high 95% recall threshold within the 500-document screening limit of our experiments.

## 7.3 Comparative Analysis Against Static Baselines

The primary evaluation of AutoDiscover's effectiveness comes from a direct comparison against the static ASReview baselines across all 26 SYNERGY datasets. Table 13 presents these results, with the best-performing model for each metric on each dataset highlighted in bold. To provide a fair assessment of high-recall performance, we also calculate summary statistics for a subset of 16 datasets where AutoDiscover achieved a non-zero Work Saved over Sampling at 95% recall (WSS@95), indicated by rows without gray highlighting.

Table 13: Performance Comparison of AutoDiscover against Static Baselines across 26 SYNERGY Datasets. For AutoDiscover, rows where WSS@95 > 0 are highlighted in gray and used in the last summary statistics. All metrics are on a 0–1 scale.

| Dataset | ASReview (NB + TF-IDF) | | | | | ASReview (LR + TF-IDF) | | | | | **AutoDiscover (Ours, 80 HPO)*** | | | | |
|---|---|---|---|---|---|---|---|---|---|---|---|---|---|---|---|
| | DRE | WSS@95 | WSS@80 | R@500 | R@200 | DRE | WSS@95 | WSS@80 | R@500 | R@200 | DRE | WSS@95 | WSS@80 | R@500 | R@200 |
| Appenzeller-Herzog | 1.7284 | 0.7974 | 0.8972 | 0.9586 | 0.7737 | 1.3060 | 0.7819 | 0.8947 | 0.9556 | 0.7678 | 5.2829 | 0.0000 | 0.8512 | 0.9231 | 0.2692 |
| Bos_2018 | 29.5515 | 0.8151 | 0.9772 | 0.9000 | 0.9000 | 18.1383 | 0.8020 | 0.9608 | 0.9000 | 0.9000 | 11.0362 | 0.9046 | 0.9200 | 1.0000 | 0.5000 |
| Brouwer_2019 | 13.1300 | 0.8910 | 0.9854 | 0.8319 | 0.6363 | 18.1686 | 0.9067 | 0.9860 | 0.8465 | 0.6246 | 64.9030 | 0.0000 | 0.9858 | 0.8548 | 0.4194 |
| Chou_2003 | 1.1959 | 0.4455 | 0.8804 | 0.8178 | 0.7600 | 1.1974 | 0.3538 | 0.8652 | 0.8178 | 0.7600 | 3.9585 | 0.7341 | 0.7825 | 1.0000 | 0.6000 |
| Chou_2004 | 1.4727 | 0.2013 | 0.2388 | 0.4691 | 0.2963 | 1.3640 | 0.1550 | 0.1812 | 0.4074 | 0.2963 | 2.8862 | 0.0000 | 0.7454 | 0.8889 | 0.5556 |
| Donners_2021 | 1.5159 | 0.6640 | 0.7784 | 0.9333 | 0.9333 | 2.1105 | 0.6742 | 0.7936 | 0.9333 | 0.9333 | 4.1613 | 0.7470 | 0.7965 | 1.0000 | 1.0000 |
| Hall_2012 | 5.0450 | 0.9102 | 0.9750 | 0.9903 | 0.8845 | 2.8158 | 0.9072 | 0.9730 | 0.9840 | 0.8393 | 17.5159 | 0.9626 | 0.9684 | 1.0000 | 0.7788 |
| Jeyaraman_2020 | 1.0840 | 0.5440 | 0.6753 | 0.9737 | 0.6948 | 1.3860 | 0.5515 | 0.7109 | 0.9601 | 0.7479 | 2.0237 | 0.0000 | 0.5830 | 0.8646 | 0.5104 |
| Leenaars_2019 | 41.8898 | 0.8953 | 0.9879 | 0.9412 | 0.9412 | 59.6085 | 0.8953 | 0.9879 | 0.9412 | 0.9412 | 16.0552 | 0.9344 | 0.9501 | 1.0000 | 0.7647 |
| Leenaars_2020 | 1.2144 | 0.5727 | 0.7101 | 0.3964 | 0.2066 | 1.2079 | 0.6497 | 0.7905 | 0.4344 | 0.2167 | 4.1669 | 0.0000 | 0.0000 | 0.2899 | 0.1269 |
| Meijboom_2021 | 2.3279 | 0.6866 | 0.7742 | 0.9730 | 0.9167 | 1.8336 | 0.6594 | 0.7732 | 0.9730 | 0.9116 | 1.7570 | 0.5441 | 0.6570 | 1.0000 | 0.7838 |
| Menon_2022 | 1.0741 | 0.6342 | 0.7617 | 0.9733 | 0.8435 | 1.0191 | 0.6690 | 0.8097 | 0.9710 | 0.8996 | 1.9160 | 0.6092 | 0.7410 | 0.9865 | 0.8108 |
| Moran_2021 | 1.0780 | 0.1523 | 0.2795 | 0.1296 | 0.0553 | 1.0107 | 0.0833 | 0.2223 | 0.1180 | 0.0526 | 2.4329 | 0.0000 | 0.0000 | 0.2342 | 0.1171 |
| Muthu_2021 | 1.1027 | 0.3380 | 0.5236 | 0.5104 | 0.2608 | 1.2187 | 0.3551 | 0.5392 | 0.5529 | 0.2567 | 2.8371 | 0.0000 | 0.0000 | 0.5238 | 0.2321 |
| Nelson_2002 | 1.3384 | 0.3714 | 0.5303 | 0.9875 | 0.9370 | 1.3792 | 0.3678 | 0.4941 | 0.9875 | 0.9355 | 1.2119 | 0.3903 | 0.4126 | 1.0000 | 0.9250 |
| Oud_2018 | 2.4202 | 0.6914 | 0.8197 | 0.9500 | 0.8625 | 2.5186 | 0.7456 | 0.8139 | 0.9500 | 0.9475 | 7.2121 | 0.8540 | 0.8398 | 1.0000 | 1.0000 |
| Radjenovic_2013 | 4.5879 | 0.8668 | 0.9499 | 0.9614 | 0.7174 | 6.2979 | 0.8540 | 0.9545 | 0.9536 | 0.7765 | 11.8227 | 0.9518 | 0.9532 | 1.0000 | 0.7083 |
| Sep_2021 | 1.1666 | 0.1400 | 0.3532 | 0.9750 | 0.9194 | 1.1372 | 0.1967 | 0.4280 | 0.9750 | 0.9413 | 1.1681 | 0.0989 | 0.2989 | 1.0000 | 0.8750 |
| Smid_2020 | 3.6238 | 0.7839 | 0.9242 | 0.9630 | 0.8560 | 6.2829 | 0.7906 | 0.9055 | 0.9630 | 0.8272 | 5.2331 | 0.8830 | 0.8801 | 1.0000 | 0.6296 |
| Walker_2018 | 1.3414 | 0.5840 | 0.9442 | 0.9531 | 0.8380 | 1.2939 | 0.7200 | 0.9556 | 0.9761 | 0.8398 | 26.4307 | 0.0000 | 0.0000 | 0.2743 | 0.1273 |
| Wassenaar_2017 | 3.4757 | 0.7785 | 0.9299 | 0.8291 | 0.6544 | 3.7897 | 0.7970 | 0.9507 | 0.8662 | 0.7116 | 13.0731 | 0.0000 | 0.9442 | 0.8559 | 0.6757 |
| Wolters_2018 | 4.3902 | 0.8109 | 0.9009 | 0.9474 | 0.7839 | 4.8649 | 0.7690 | 0.8685 | 0.8837 | 0.7341 | 8.5259 | 0.8790 | 0.9206 | 1.0000 | 0.6316 |
| van_Dis_2020 | 1.9020 | 0.6636 | 0.8545 | 0.6283 | 0.3409 | 2.1259 | 0.7123 | 0.8392 | 0.6096 | 0.2710 | 8.5865 | 0.0000 | 0.0000 | 0.4722 | 0.2639 |
| van_de_Schoot_2018 | 7.0330 | 0.8958 | 0.9695 | 0.9737 | 0.9536 | 4.6855 | 0.8936 | 0.9657 | 0.9737 | 0.9501 | 8.8136 | 0.8976 | 0.9114 | 0.9737 | 0.5263 |
| van_der_Valk_2021 | 1.1359 | 0.3858 | 0.5502 | 0.9888 | 0.6926 | 1.4301 | 0.3963 | 0.5423 | 0.9886 | 0.7216 | 1.7180 | 0.4889 | 0.5138 | 1.0000 | 0.6292 |
| van_der_Waal_2022 | 2.0625 | 0.7605 | 0.8539 | 0.9697 | 0.8274 | 3.1110 | 0.7517 | 0.8852 | 0.9697 | 0.8476 | 4.3584 | 0.7745 | 0.8274 | 1.0000 | 0.6667 |
| *Summary Statistics (All 26 Datasets)* | | | | | | | | | | | | | | | |
| Max | 41.8898 | 0.9102 | 0.9879 | 0.9903 | 0.9536 | 59.6085 | 0.9072 | 0.9879 | 0.9886 | 0.9501 | 64.9030 | 0.9626 | 0.9858 | 1.0000 | 1.0000 |
| Min | 1.0741 | 0.1400 | 0.2388 | 0.1296 | 0.0553 | 1.0107 | 0.0833 | 0.1812 | 0.1180 | 0.0526 | 1.1681 | 0.0000 | 0.0000 | 0.2342 | 0.1171 |
| Average | 5.3034 | 0.6262 | 0.7702 | 0.8433 | 0.7110 | 5.8193 | 0.6323 | 0.7727 | 0.8420 | 0.7174 | 9.1957 | 0.4482 | 0.6340 | 0.8516 | 0.5818 |
| Median | 1.8152 | 0.6753 | 0.8542 | 0.9516 | 0.8057 | 1.9721 | 0.7162 | 0.8522 | 0.9518 | 0.8019 | 4.7958 | 0.5165 | 0.7895 | 1.0000 | 0.6294 |
| *Summary Statistics (AutoDiscover, for 16 datasets with WSS@95 > 0 only)* | | | | | | | | | | | | | | | |
| Max | 41.8898 | 0.9102 | 0.9879 | 0.9903 | 0.9536 | 59.6085 | 0.9072 | 0.9879 | 0.9886 | 0.9501 | 17.5159 | 0.9626 | 0.9684 | 1.0000 | 1.0000 |
| Min | 1.0741 | 0.1400 | 0.3532 | 0.8178 | 0.6926 | 1.0191 | 0.1967 | 0.4280 | 0.8178 | 0.7216 | 1.1681 | 0.0989 | 0.2989 | 0.9737 | 0.5000 |
| Average | 6.8974 | 0.6723 | 0.8117 | 0.9528 | 0.8581 | 7.4019 | 0.6704 | 0.8138 | 0.9478 | 0.8667 | 6.6540 | 0.7284 | 0.7733 | 0.9975 | 0.7394 |
| Median | 2.3741 | 0.7260 | 0.8672 | 0.9664 | 0.8735 | 2.6672 | 0.7487 | 0.8669 | 0.9664 | 0.8998 | 4.7958 | 0.8143 | 0.8336 | 1.0000 | 0.7365 |

*Screening for AutoDiscover was limited to 500 papers.

**Analysis of Overall Performance.** A comprehensive analysis of the results presented in Table 13 reveals the distinct performance profile of AutoDiscover compared to the established static baselines. The summary statistics, aggregated across all 26 SYNERGY datasets, highlight the substantial gains in overall screening efficiency achieved by our adaptive, graph-aware framework, alongside a nuanced trade-off regarding high-recall performance.

**Superior Overall Screening Efficiency:** The most striking result is AutoDiscover's superior performance on the Discovery Rate Efficiency (DRE) metric. With a mean DRE of 9.20 and a median of 4.80, AutoDiscover significantly outperforms both the Naive Bayes (NB) baseline (mean 5.30, median 1.82) and the Logistic Regression (LR) baseline (mean 5.82, median 1.97). The median DRE, which is less sensitive to outliers, indicates that our system is typically more than twice as efficient as these strong static methods. This demonstrates that by dynamically adapting its Query Strategy and leveraging the rich relational information in the literature graph, AutoDiscover finds a larger fraction of relevant documents for a given proportion of screening effort. This advantage is particularly pronounced in several datasets, such as Brouwer_2019 (DRE 64.90) and Walker_2018 (DRE 26.43), where the adaptive approach unlocked massive efficiency gains.

**Trade-offs in High-Recall Scenarios:** This focus on overall efficiency, driven by our hyperparameter optimization (HPO) objective, leads to a notable trade-off in reaching the highest recall thresholds. The mean Work Saved over Sampling at 95% recall (WSS@95) for AutoDiscover is 0.45, which is lower than the baselines (NB: 0.63, LR: 0.63). This is primarily due to ten datasets where AutoDiscover's best HPO run did not achieve 95% recall within the 500-paper screening limit, resulting in a WSS@95 of 0.00 for those cases.
However, this metric requires careful interpretation. For the 16 datasets where AutoDiscover did reach the 95% recall target, its performance is highly competitive. The median WSS@95 for this subset is 0.81, and the mean is 0.73, indicating that when the system is configured to pursue high recall, it does so effectively. Furthermore, the system's robustness is more clearly demonstrated by the WSS@80 metric.

Here, AutoDiscover achieves a median of 0.79, on par with the baselines, showcasing its strong capability to find the vast majority of relevant documents efficiently.

**Effectiveness Within a Fixed Screening Budget:** From a practical standpoint, a researcher often operates with a fixed budget of time or the number of papers they can screen. The Recall@k metrics demonstrate AutoDiscover's exceptional utility in this context. With a median Recall@500 of 1.00, our system finds virtually all relevant documents for a typical dataset within this initial screening budget. Even with a much smaller budget of 200 papers, the median Recall@200 is 0.63, meaning over half of the essential literature is typically identified very early in the process.
In summary, the comparative analysis validates the core hypothesis of this thesis. AutoDiscover's adaptive, graph-aware agent consistently provides a more efficient discovery process than static methods, as measured by DRE. While its prioritization of overall efficiency can sometimes deprioritize finding the final, most elusive relevant documents, its strong performance on WSS@80 and Recall@k metrics confirms its practical value as a robust tool for accelerating systematic literature reviews.

## 7.4 Analysis of Hyperparameter Optimization Results

To understand the relationship between the optimization process and the final performance metrics, we visualize the hyperparameter optimization (HPO) history for all 26 datasets. The following figures group these histories based on whether the best-found configuration achieved a Work Saved over Sampling at 95% recall (WSS@95) greater than zero.
Figure 20 displays the HPO histories for the 10 datasets where the best trial resulted in a WSS@95 of 0.00. In these plots, the objective value (DRE) often reaches high values, but the optimization process may not have discovered a configuration that could also efficiently find the last few relevant documents required to meet the 95% recall threshold within the screening limit. This visually confirms the trade-off inherent in prioritizing DRE, which favors rapid early discovery over exhaustive late-stage recall.
In contrast, Figure 21 shows the HPO histories for the 16 datasets where AutoDiscover successfully achieved a positive WSS@95 score. For many of these datasets, the optimization process shows a clearer convergence towards high-performing configurations. This suggests that for these particular data distributions, the TPE sampler was able to identify hyperparameter settings that effectively balanced the goal of high overall efficiency (DRE) with the ability to achieve high recall, leading to strong performance across multiple metrics.

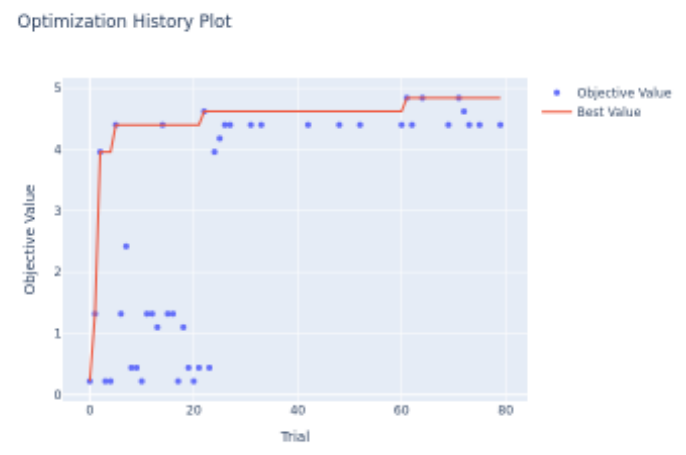

(a) Appenzeller-Herzog

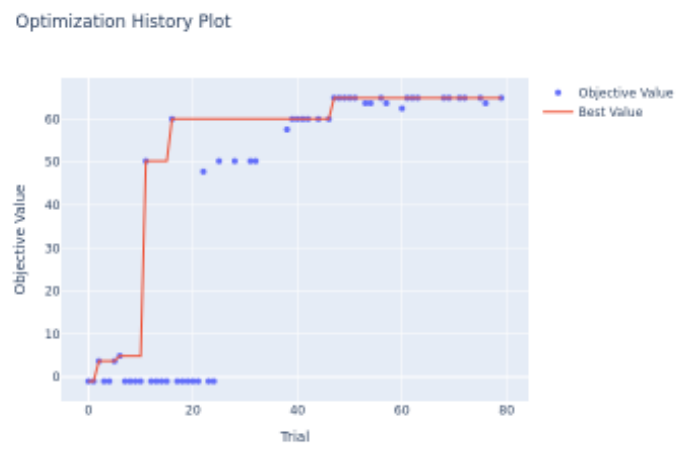

(b) Brouwer

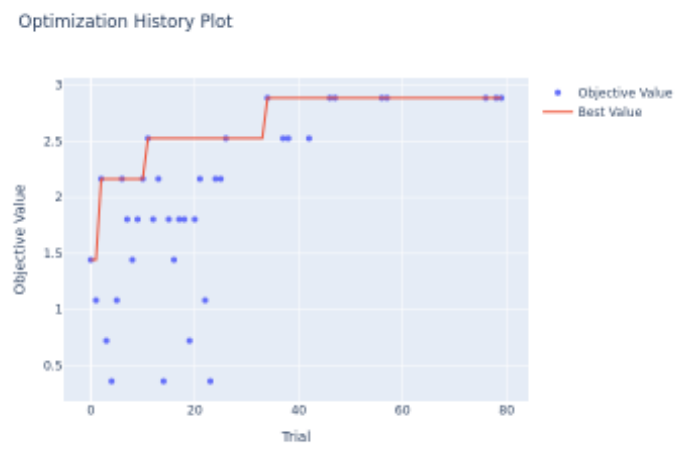

(c) Chou (2004)

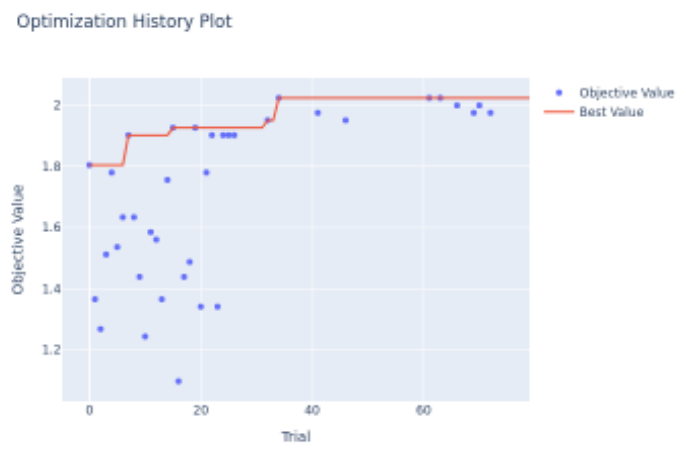

(d) Jeyaraman

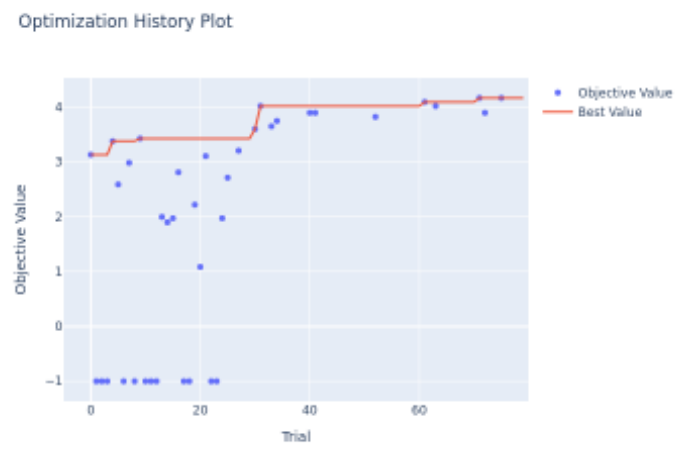

(e) Leenaars (2020)

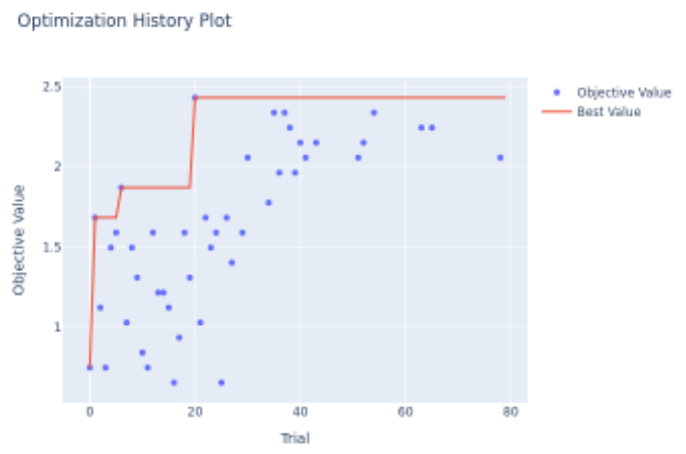

(f) Moran

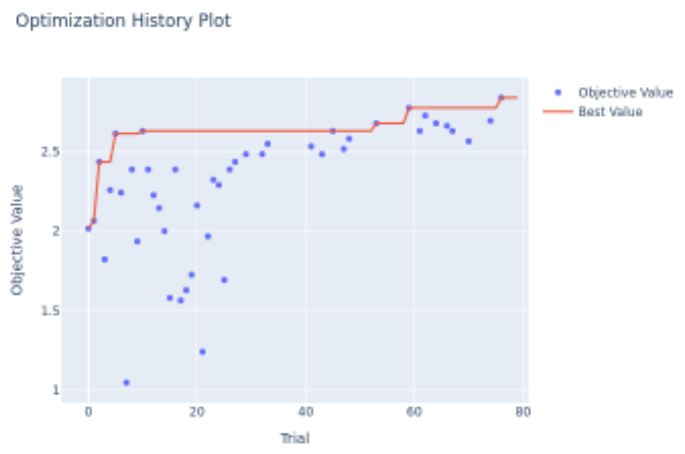

(g) Muthu

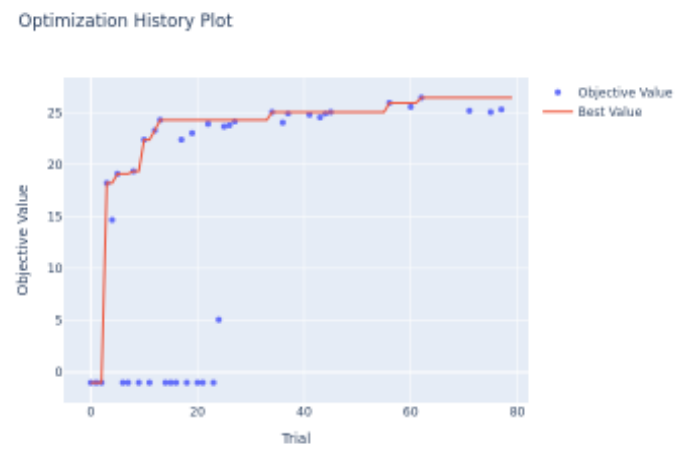

(h) Walker

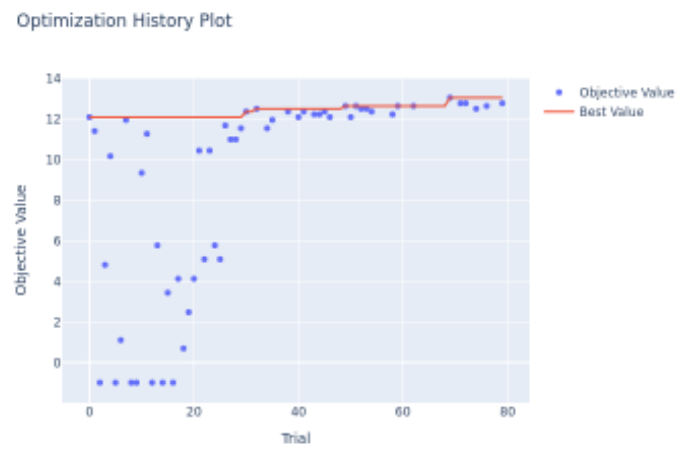

(i) Wassenaar

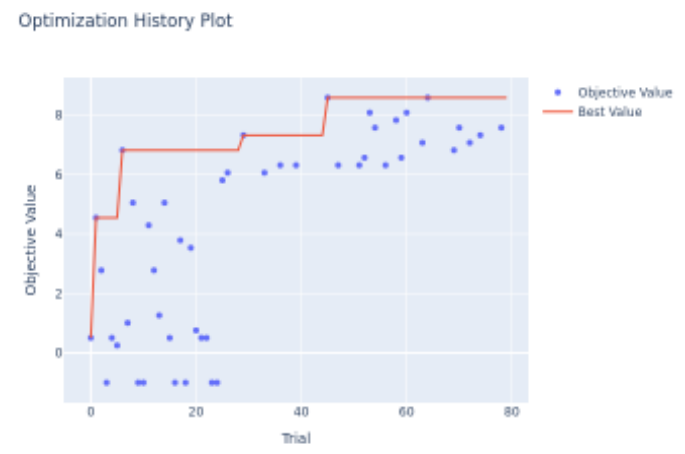

(j) van Dis

Figure 20: Hyperparameter Optimization history for the 10 datasets where AutoDiscover did not achieve a WSS@95 score greater than zero within the 500-paper screening limit. Each plot shows the objective value (DRE) over 80 trials.

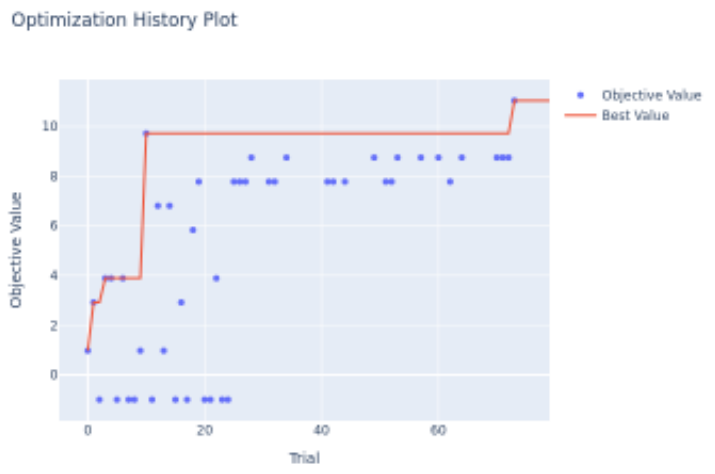

(a) Bos

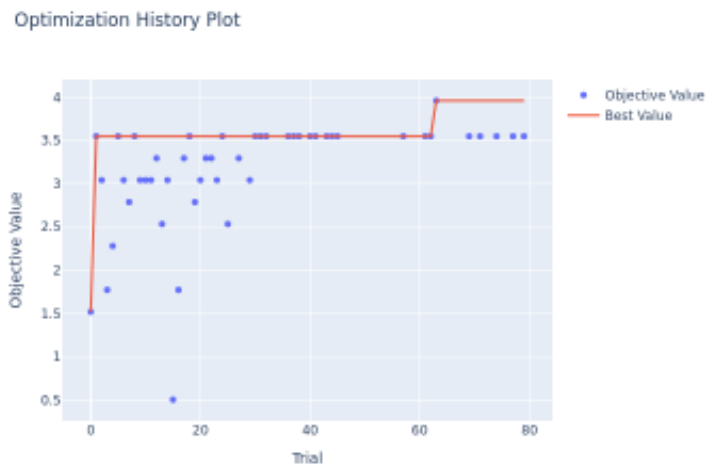

(b) Chou (2003)

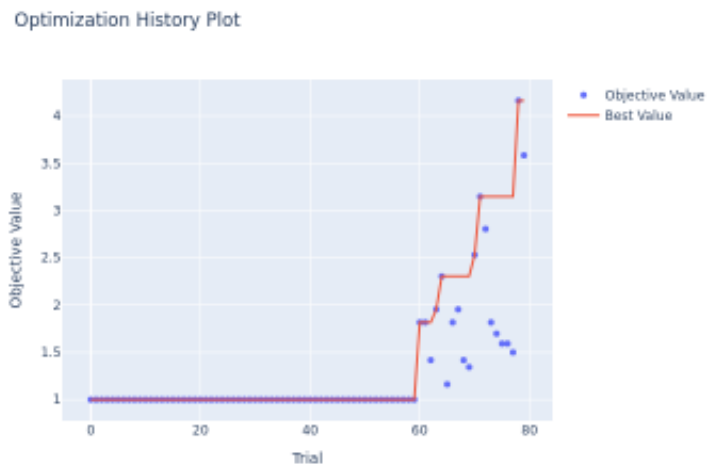

(c) Donners

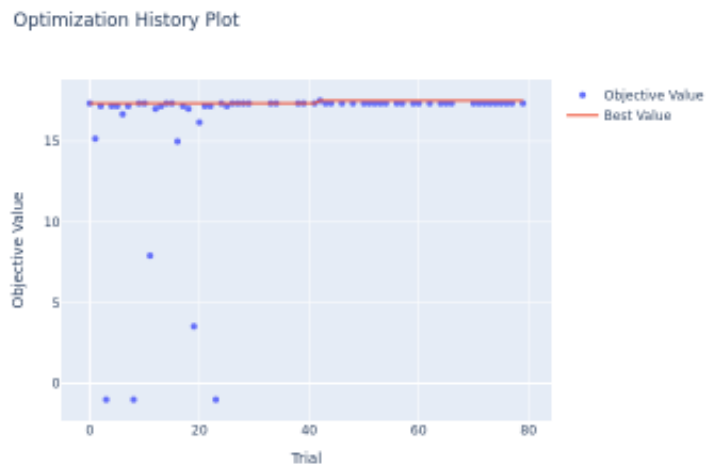

(d) Hall

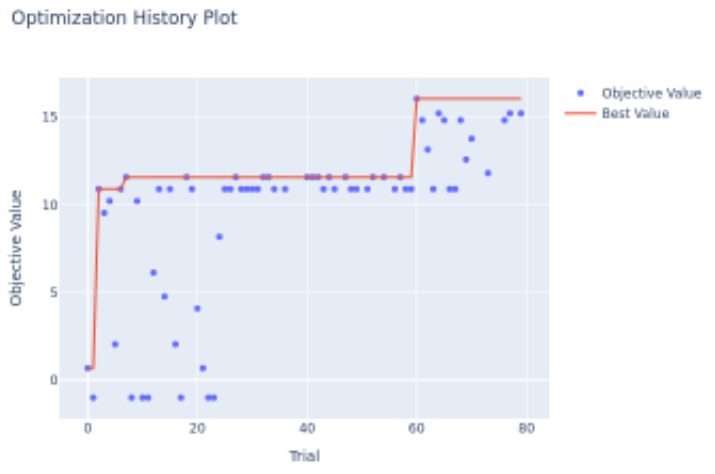

(e) Leenaars (2019)

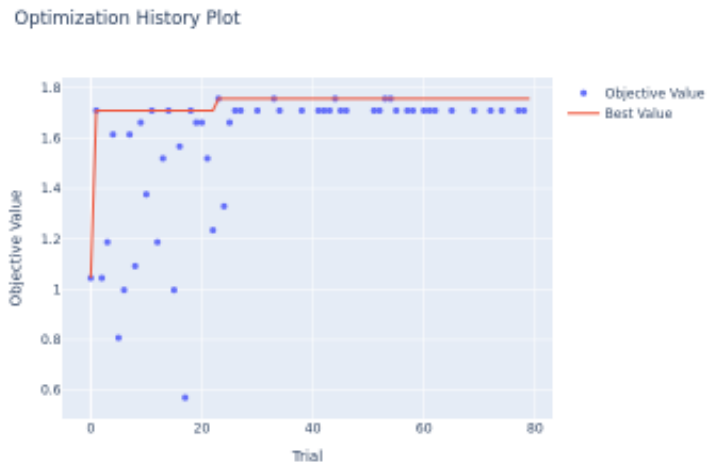

(f) Meijboom

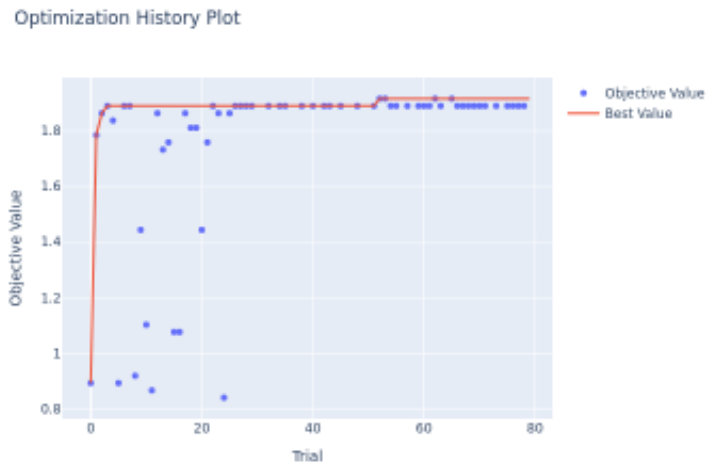

(g) Menon

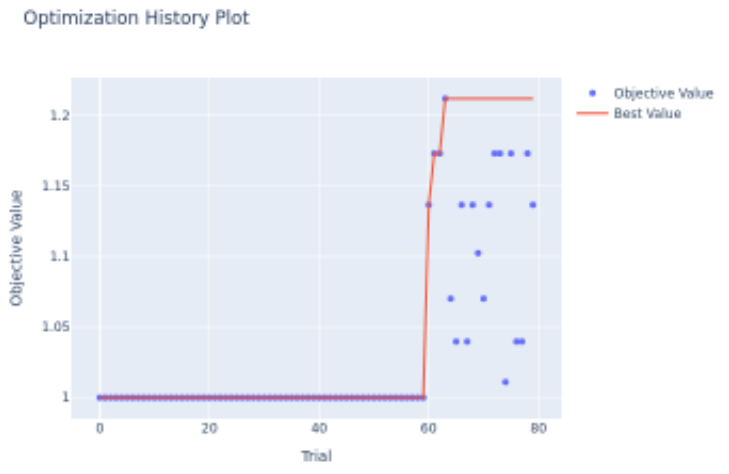

(h) Nelson

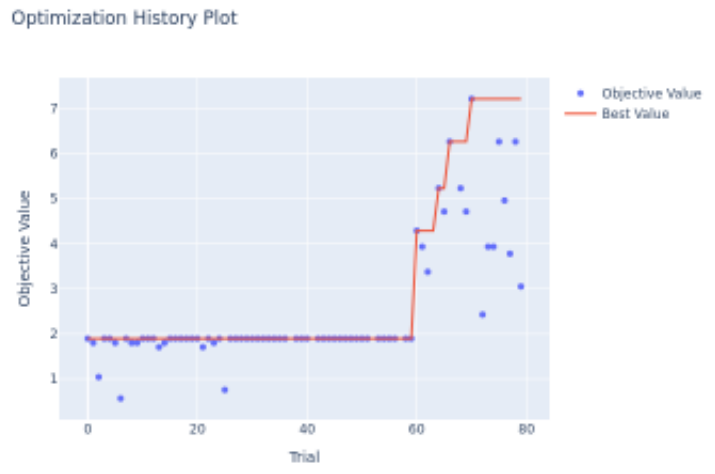

(i) Oud

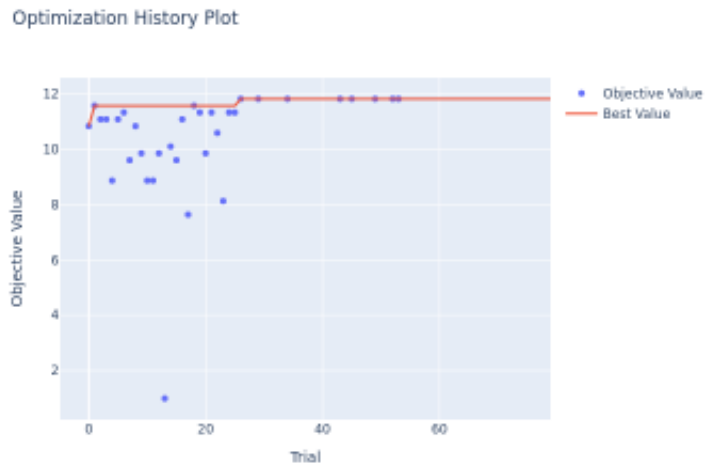

(j) Radjenovic

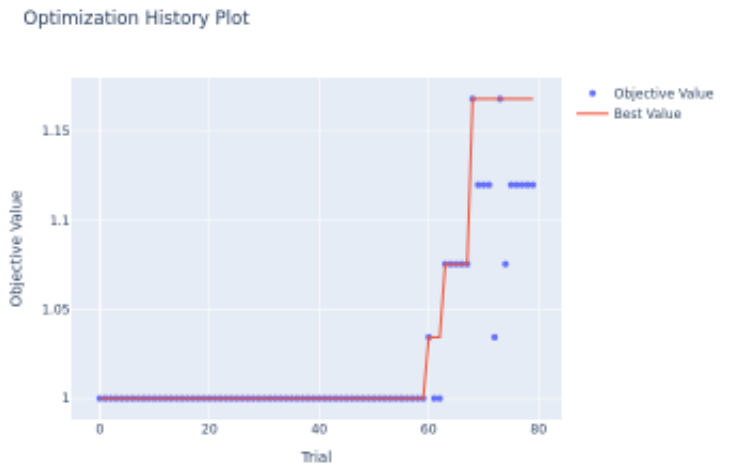

(k) Sep

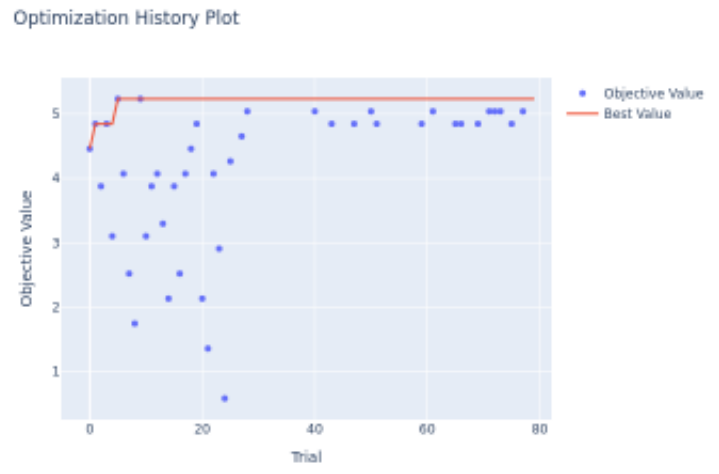

(l) Smid

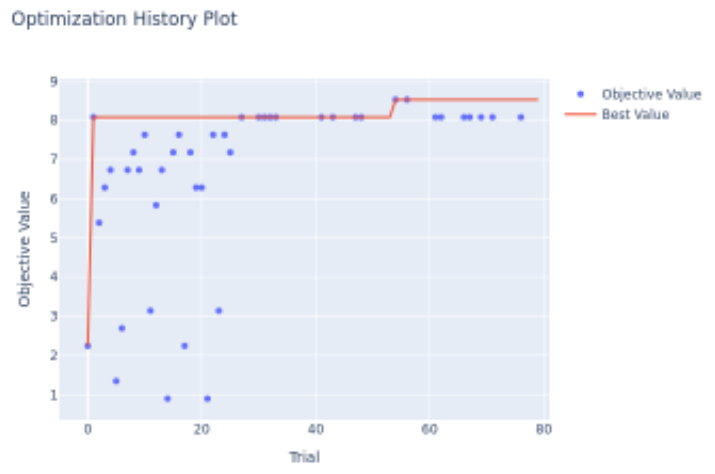

(m) Wolters

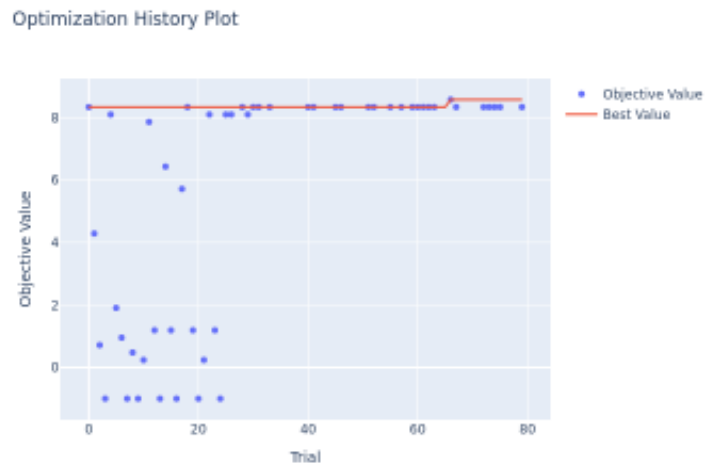

(n) van de Schoot

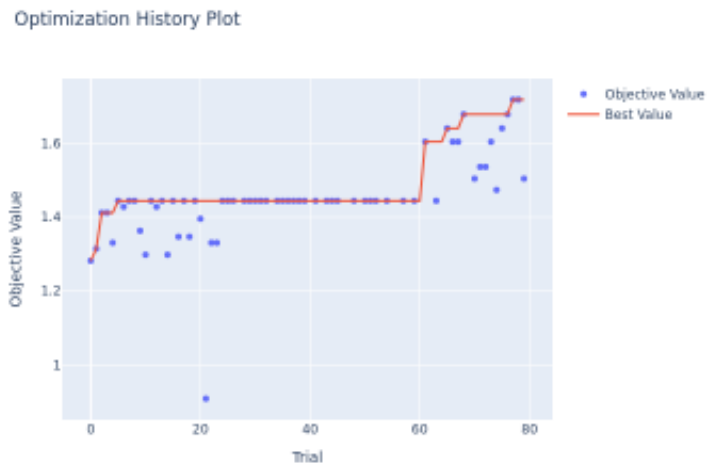

(o) van der Valk

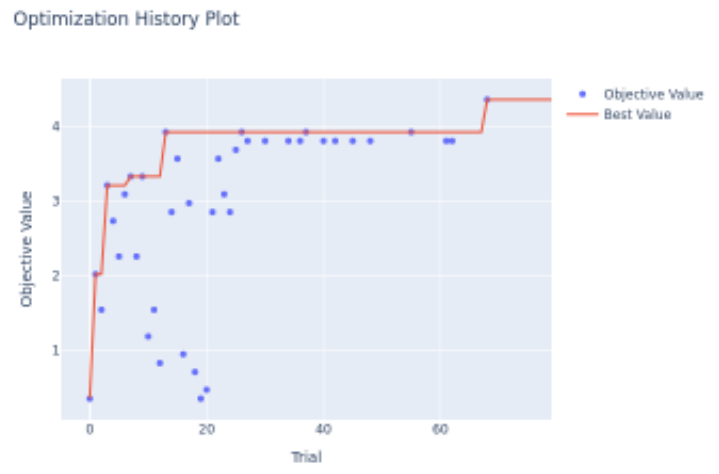

(p) van der Waal

Figure 21: Hyperparameter Optimization history for the 16 datasets where AutoDiscover achieved a positive WSS@95 score. These plots often show clearer convergence towards high-performing configurations.

**Conclusion on HPO Robustness.** A collective analysis of the HPO histories in Figure 20 and Figure 21 provides strong evidence for the robustness of our optimization process. Despite a relatively modest budget of 80 trials per dataset, the TPE sampler consistently identifies high-performing regions in the hyperparameter space. In nearly all cases, the best objective value (blue dot) is not an isolated outlier but is surrounded by other strong trials, indicating that the final selected configuration is not the result of a single lucky guess. Rather, the process reliably converges toward a stable region of good performance, lending confidence that the reported results are a representative showcase of AutoDiscover's capabilities.

## 7.5 Interpreting Agent Behavior: A Case Study with TS-Insight

While quantitative metrics demonstrate AutoDiscover's effectiveness, they do not reveal *how* the adaptive agent achieves its results. To open this "black box," we use TS-Insight, our visual analytics dashboard, to perform a case study on a single experimental run. By examining the agent's behavior on the `Appenzeller-Herzog_2019` dataset, we can demonstrate how TS-Insight fulfills its three primary XAI (Explainable AI) objectives: verifying the agent's mechanics, explaining its decisions, and diagnosing its behavior under uncertainty.

### 7.5.1 Verification: Confirming the Learning and Forgetting Mechanisms

A fundamental requirement for trusting any learning system is to verify that its feedback loop is functioning as designed. Figure 22 provides a detailed view from the TS-Insight dashboard that allows for direct confirmation of the DTS agent's learning process.

We can trace the agent's actions and their consequences through the coordinated plots. For instance, in the "Barcode" view (Panel C), the blue stroke for Arm 8 at sampling step $t = 235$ indicates a successful query where a relevant document was found. This single event can be directly correlated with the upward tick in the blue *alpha-count* line in Panel B at the same time step, confirming that the success reward was correctly registered. Furthermore, the dashboard makes the "forgetting" mechanism of Discounted Thompson Sampling visible. The data for these plots is logged directly from the agent at each query step, providing a direct, unmediated view of its internal state. This is clearly demonstrated by observing the evidence counters for both success and failure:

- **Decay of Success Evidence (Alpha):** For Arm 7, after its last successful query around step 260, the blue alpha line in Panel B begins a gradual, exponential decay. This shows that as time passes without further successes, the agent correctly "forgets" this past reward, even when the arm is subsequently chosen but fails to find a relevant document.
- **Decay of Failure Evidence (Beta):** For Arm 8, after its first failure (orange stroke in Panel C), the orange beta (failure) counter line in Panel B ticks up by one. In the subsequent period where this arm is not chosen, we can see the beta (failure) counter also start to decay.

This visual evidence of both alpha and beta decay provides definitive proof that the agent is correctly down-weighting old evidence, both rewards and punishments, and prioritizing recent feedback. This is the critical feature that allows the agent to adapt in a non-stationary environment. When an arm is not selected for a prolonged period, its evidence counters converge towards the lower bound we set to prevent zero-division errors, effectively resetting its influence and allowing other, more recently successful arms to dominate.

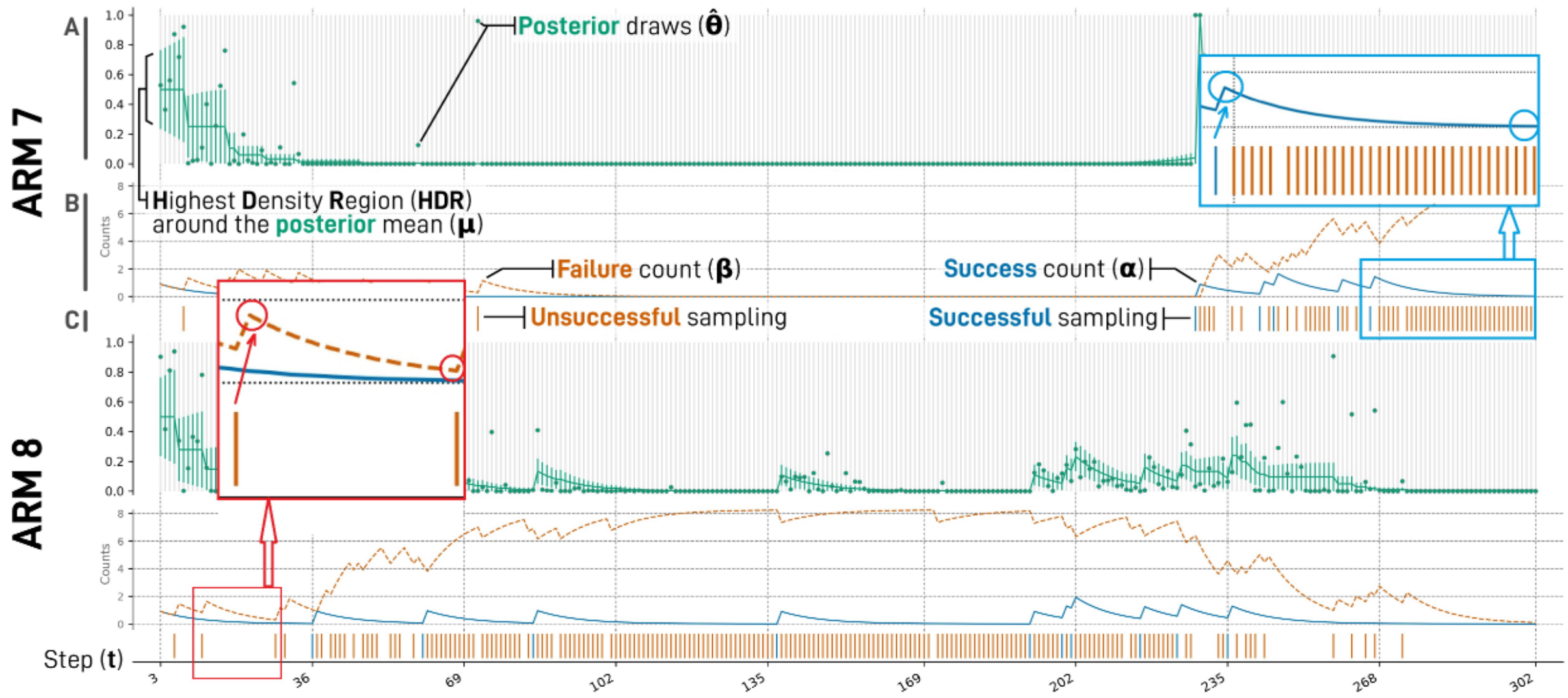

Figure 22: An excerpt from the TS-Insight dashboard for the Appenzeller-Herzog_2019 run, demonstrating algorithm verification. **(A)** The HDR Evolution Plot shows the agent's belief. **(B)** The Alpha/Beta plot confirms reward updates (visible upward ticks) and the exponential decay of evidence from discounting. **(C)** The Barcode provides a history of arm pulls and their outcomes (blue for success, orange for failure).

### 7.5.2 Explanation and Diagnosis: Deconstructing an Exploratory Decision

Beyond verification, TS-Insight is designed to explain why a particular decision was made (Explore or Exploit) and to diagnose the agent's level of certainty about the decision. Figure 24 provides a snapshot of the agent's state at sampling step $t = 228$, a pivotal moment in this particular run.

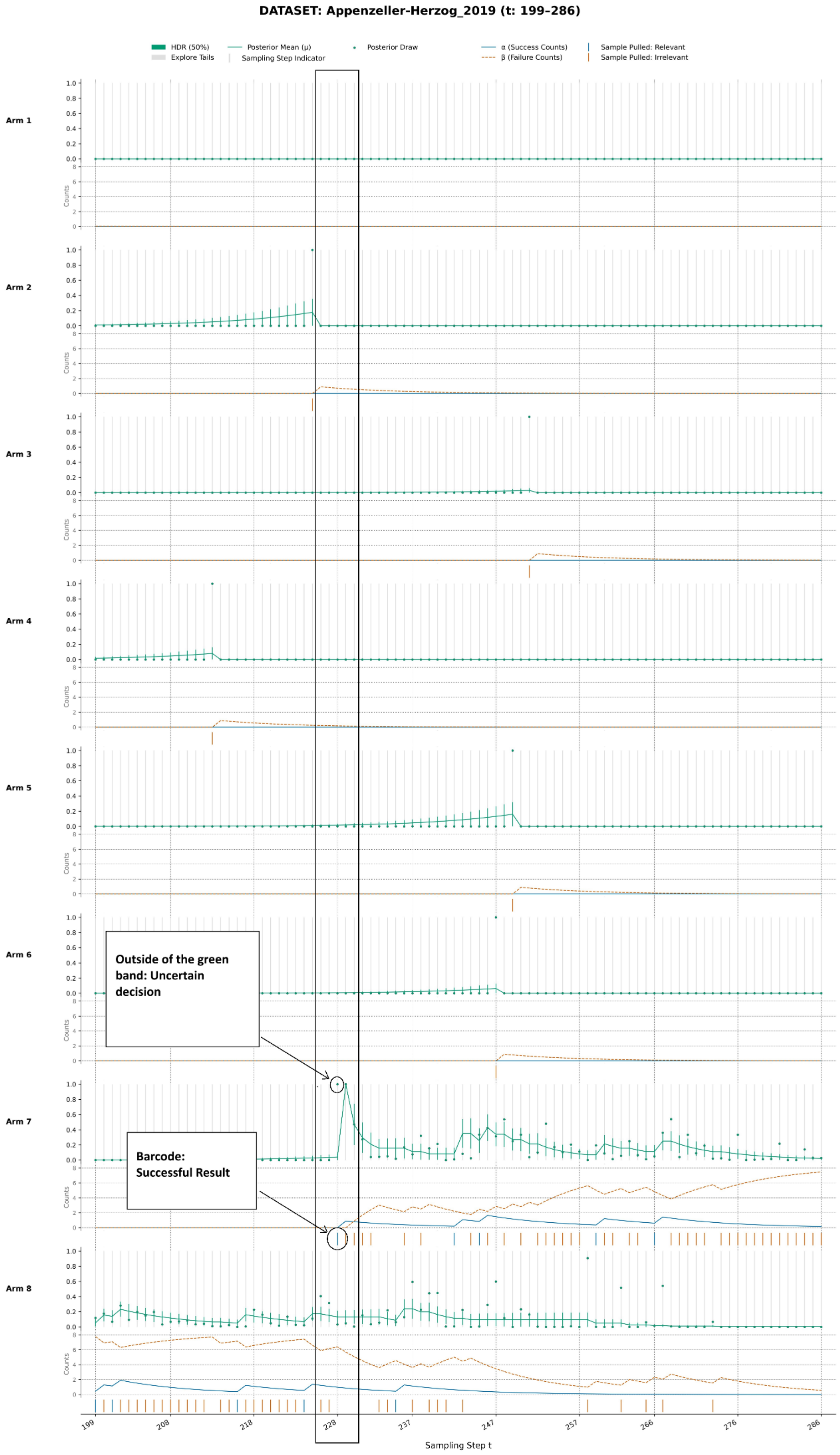

Figure 23: An excerpt from the TS-Insight dashboard for the Appenzeller-Herzog_2019 run, demonstrating a critical exploratory decision and its successful outcome. The annotations highlight the two key events: the selection of Arm 7 at sampling step $t = 228$ was an uncertain, exploratory choice, as its posterior draw was outside the main confidence band (the 50% HDR). This decision was rewarded with a successful discovery, as indicated by the blue stroke in the barcode, which is used to update the agent's beliefs for subsequent steps.

**Explanation of an Exploratory Choice.** The following analysis deconstructs a specific decision made by the agent. It is important to note that the visualizations presented are rendered directly from the log files of the experimental run, offering a faithful reconstruction of the algorithmic choices made at each step. At step $t = 228$, the agent faced a classic exploration-exploitation dilemma. The XAI Snapshot shows that Arm 8 had the highest posterior mean ($\mu$), making it the optimal choice for pure exploitation. However, the agent instead selected Arm 7. The snapshot clearly explains this seemingly suboptimal choice: due to the high variance (uncertainty) in its belief about Arm 7, the random value it drew from the posterior ($\hat{\theta}$) was the highest of all arms. This was a deliberate act of **exploration**. As shown in the barcode (Figure 23), this exploratory gamble paid off with a successful discovery, causing the agent to update its belief and favor Arm 7 more heavily in subsequent steps.

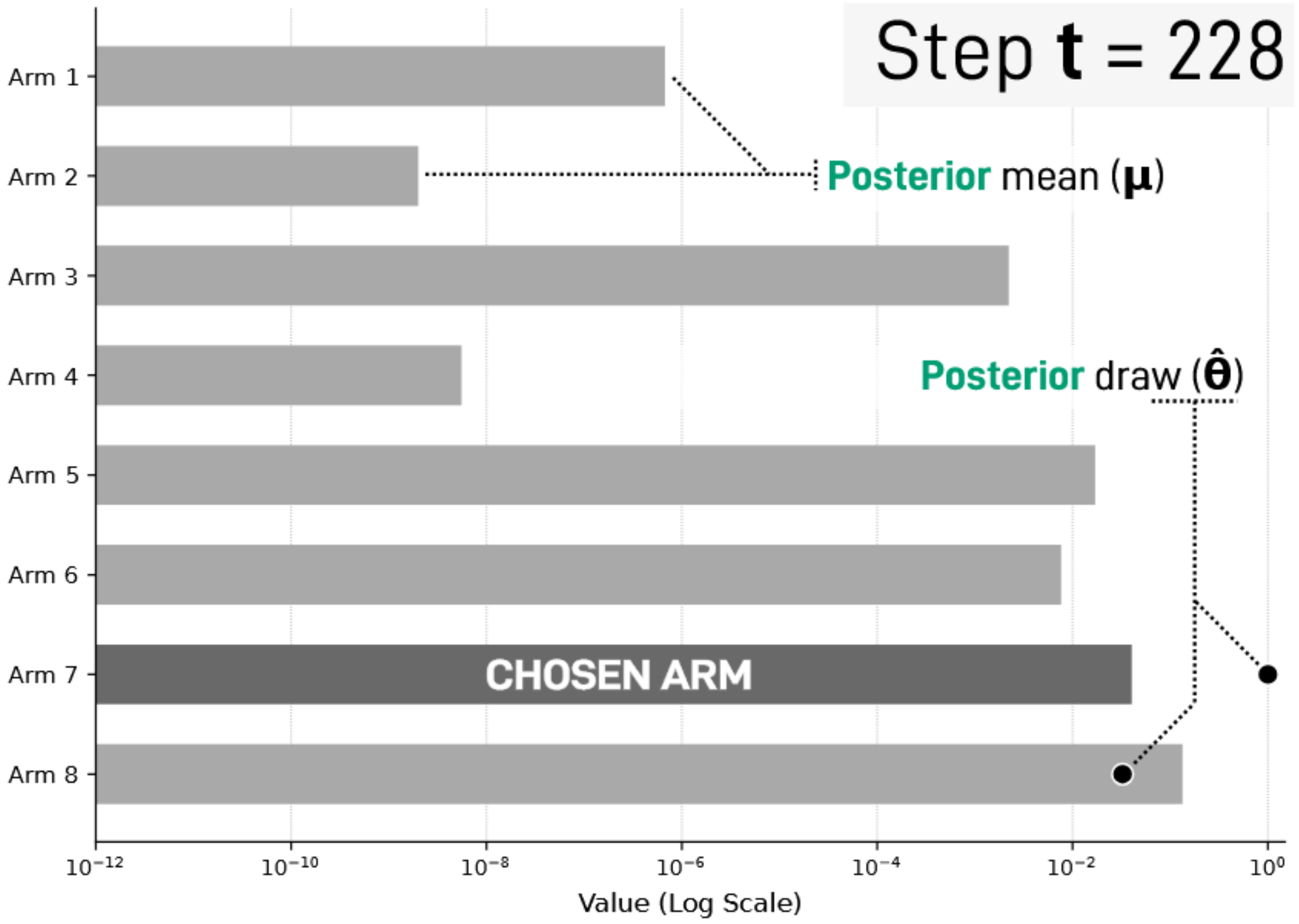

Figure 24: The XAI Snapshot view explains the agent's choice at step $t = 228$. Although Arm 8 has the highest posterior mean (the greedy choice), Arm 7 was chosen because its random posterior draw was the highest, clearly illustrating an exploration step.

**Diagnosing Uncertainty and Risk Tolerance.** The reliability of this exploratory decision can be diagnosed by referring back to the HDR Evolution Plot (Figure 23). The central shaded band represents the 50% Highest Density Region (HDR), a parameter that is user-configurable within TS-Insight to define a desired risk tolerance. For this experiment, it was set at 50%, but a human supervisor could set it to 90% for a risk-averse strategy or 25% to encourage more exploration.

At step $t = 228$, the posterior draw for Arm 7 was a "long-shot" that fell far outside this central 50% band. This immediately signals to the user that the agent has made a decision that exceeds the currently configured confidence level. Academically, this visualization contrasts the agent's exploratory action with the greedy, exploitation-focused choice, which would have been to select the arm with the highest posterior mean ($\mu$). A draw from the uncertain tails of a wide belief distribution, as seen here, represents a calculated risk. Therefore, by allowing a supervisor to set this HDR threshold, TS-Insight provides a direct mechanism to not only gauge the risk of the agent's strategy at any point but also to assess whether its exploratory behavior aligns with their specified risk profile.

By combining these views, TS-Insight provides a complete narrative of the agent's behavior, transforming the complex DTS mechanism from a black box into an interpretable and verifiable system.

# 8 Limitations and Future Work

The AutoDiscover framework represents a significant step towards adaptive, graph-aware literature screening. Its performance on the SYNERGY benchmark demonstrates the value of combining heterogeneous graph representations with a reinforcement learning agent. However, this research is not without its limitations. The design choices made and the scope of the evaluation highlight several important areas for future investigation. These can be grouped into three primary categories: architectural and algorithmic enhancements, methodological rigor in evaluation, and practical deployment considerations.

## 8.1 Architectural and Algorithmic Enhancements

**Static Graph Topology.** A primary limitation of the current AutoDiscover framework lies in its static graph topology. While the embedded Graph Attention (GAT) mechanism provides a degree of dynamism by re-calculating edge influence via attention scores during each retraining phase, the underlying graph structure, the set of edges connecting the nodes, is constructed in Section 5.2 once and remains fixed. This initial construction relies on pre-trained SPECTER2 embeddings, which are task-agnostic.
This design was a pragmatic choice to ensure computational tractability and to isolate the effects of the DTS agent. However, it prevents the system from leveraging the task-specific representations learned by the GNN to refine the graph topology itself. A significant avenue for future research is to develop a fully dynamic graph construction process. This would involve periodically re-evaluating potential semantic links based on the GNN's updated embeddings, allowing the system to add new edges between nodes that have become semantically closer and prune those that have drifted apart. We hypothesize that such a dynamic topology would enable the system to uncover discovery pathways mid-review, better adapting its structural understanding and more effectively finding 'islands' of relevance that were disconnected in the initial, generic graph.

**Expanding the Query Arm Portfolio.** The nine query strategies used in this thesis were selected to provide a diverse and computationally tractable portfolio covering the fundamental axes of active learning. We hypothesize that while this portfolio is robust, its performance could be further enhanced. This presents a promising avenue for future research, including the incorporation of more sophisticated or computationally intensive query strategies, such as those based on expected error reduction, or developing hybrid arms that combine these existing primitives in new ways.

**Fixed Retraining Schedule.** In our experiments, the GNN was retrained after a fixed batch of 10 new labels was acquired. This was a pragmatic choice to standardize the experimental protocol across 80 HPO trials per dataset 6.3 and to manage computational resources. A more intelligent, adaptive retraining policy could yield significant benefits. Future work could investigate policies where retraining is triggered by a drop in model confidence, or where the user has the ability to pause retraining when the GNN is performing well to avoid overfitting and reduce computational load.

**Enriching Node Feature Representations.** The current model initializes non-paper nodes (e.g., author, institution) with learnable random embeddings. This places a heavy burden on the GNN to infer their meaning from limited labeled data. Future versions could significantly improve performance by initializing these nodes with **rich a priori features**. For instance, author nodes could be featurized with their publication count or h-index, and institution nodes with geographic or ranking data, injecting valuable prior knowledge into the system from the outset.

**Refining the Agent's Learning Mechanism.** The DTS agent, as described in Section 5.3.2, while effective, operates on two simplifications. First, it uses a **binary reward signal** (1 for relevant, 0 for irrelevant). A more sophisticated reward function, perhaps weighted by the information gain or the structural importance of a newly discovered document, could provide a more nuanced learning signal. Second, the agent uses a **fixed discount factor** $\gamma$ determined via HPO (Table 10). An advanced implementation could feature a dynamic discount factor that adapts to the perceived rate of change in the environment, aligning more closely with the theory presented in Section C.

## 8.2 Methodological and Evaluation Rigor

**Benchmarking Against Adaptive Baselines.** Our evaluation in Section 7 compared AutoDiscover against strong static baselines. However, to more precisely isolate the contribution of our specific architecture (DTS + GNN), future work must include comparisons against other **adaptive baselines**. This would involve benchmarking against, for example, a simpler bandit algorithm like UCB1 controlling the same arm portfolio, or a bandit controlling a portfolio of different classifiers (e.g., Naive Bayes vs. Logistic Regression), to provide a more comprehensive performance context.

**Systematic Ablation and Correlation Studies.** The case studies presented in Section 5.1.6 provided compelling but anecdotal evidence of the DTS agent's importance. A more rigorous approach would involve conducting a **systematic ablation study** across all 26 datasets, comparing the full AutoDiscover system to a version that uses only the GNN exploit arm. Furthermore, a detailed **correlation analysis** between dataset characteristics (Section 4) and performance metrics would move the analysis from reporting results to explaining **why** AutoDiscover excels or struggles on different types of problems.

## 8.3 Towards Practical Deployment and Generalization

**The Hyperparameter Optimization Bottleneck.** The strongest results reported in Section 7 were achieved via an extensive, per-dataset hyperparameter optimization of 80 trials. While academically necessary to establish the model's potential, this is impractical for real-world use as discussed in Section 6.3. A critical next step is to address this bottleneck by investigating the **generalization of a single, robust hyperparameter configuration**. This would involve identifying a set of "default" parameters (e.g., derived from the median of all HPO runs) and evaluating its out-of-the-box performance across the entire benchmark, a true test of the system's practicality.

**Multi-Objective Optimization.** Our HPO process (Section 6.3) prioritized Discovery Rate Efficiency (DRE), which, as our results show, sometimes came at the cost of achieving very high recall (WSS@95). This highlights a limitation of using a single optimization objective. Future work should explore **multi-objective optimization** techniques that can find configurations on the Pareto front, allowing a user to choose a model that aligns with their specific goal, whether it is maximum early-stage discovery or guaranteed high recall.

**Human-in-the-Loop Integration.** Finally, while the framework includes an interactive arm (Section B), its dynamics were not evaluated. A crucial future direction is to conduct a **formal user study**. This would not only assess the usability of the system but also investigate the synergistic effects of combining the agent's recommendations with expert intuition. Such a study would provide invaluable insights into building trust and optimizing the collaborative intelligence between human reviewers and AI-powered tools like AutoDiscover.

# 9 Conclusion

This thesis addressed the significant challenges of label scarcity, extreme class imbalance, and the non-stationary nature of the screening process in systematic literature reviews. Traditional active learning systems, which rely on static query strategies and overlook the rich relational structure of scholarly metadata, often fall short in these demanding environments. The central research question of this work was to determine if a dynamically adaptive, graph-aware reinforcement learning agent could accelerate high-precision discovery under these conditions.

To answer this question, this work presented the design, implementation, and validation of AutoDiscover, a novel framework that integrates a Heterogeneous graph Attention Network (HAN) with a Discounted Thompson Sampling (DTS) agent. A secondary contribution was the development of TS-Insight, an open-source visual analytics dashboard designed to enhance the transparency and explainability of the adaptive agent's decision-making process. Through a comprehensive empirical evaluation on the 26-dataset SYNERGY benchmark, this research has yielded several definitive conclusions.

The empirical evaluation demonstrated that AutoDiscover achieves superior overall screening efficiency compared to established static baselines. The framework exhibited a median Discovery Rate Efficiency (DRE) of 4.80, indicating a typical performance nearly five times more efficient than random screening and more than double that of the static models. Furthermore, the adaptive DTS agent was proven to be indispensable for overcoming the "cold-start" problem; an ablation study showed that without the agent's ability to balance exploration and exploitation, the underlying GNN model failed to bootstrap the learning process from a single positive seed. The system proved highly robust in early-to-mid-stage discovery, achieving a median Work Saved over Sampling of 0.79 at 80% recall and identifying, on average, all relevant documents within the first 500 screened.

The results also illuminated a deliberate trade-off inherent in the optimization process. By prioritizing overall efficiency (DRE), the system's performance on achieving the final few percentage points of recall (WSS@95) was sometimes subordinate to that of the baselines. This finding, along with the performance variance across datasets, validates the core hypothesis that a "one-size-fits-all" model is insufficient and that an adaptive approach is necessary to navigate the diverse and evolving landscape of a systematic literature review.

The architectural choices of AutoDiscover were also validated. The use of heterogeneous graphs proved essential for creating a rich and resilient representation of scholarly metadata, while pre-trained semantic embeddings were confirmed to be superior to keyword-based models for capturing conceptual similarity. The success of the framework underscores the efficacy of using a Discounted Thompson Sampling agent to manage a portfolio of query strategies, allowing the system to dynamically adapt as the utility of each strategy changes over the course of the review. Finally, the TS-Insight dashboard was shown to be an effective tool for verifying the agent's internal mechanics, explaining its choices, and diagnosing its level of certainty, providing a critical layer of trust and interpretability.

In conclusion, this thesis successfully demonstrates that the synergy between graph-based deep learning and adaptive reinforcement learning provides a robust and efficient solution to the cold-start imbalance challenge in systematic literature reviews. AutoDiscover stands as a validated framework that not only outperforms static methods but also provides a principled, adaptive approach to accelerating AL for literature review.

# A Terminology

This section provides definitions for all of the key terms used throughout this thesis, organized by domain. The definitions are tailored for a reader with a general computer science background but who may not be a specialist in all sub-fields discussed.

## A.1 Systematic literature Reviews and Information Science

**Systematic Literature Review (SLR)** A rigorous, protocol-driven method of identifying, appraising, and synthesizing all relevant studies on a specific topic to answer a research question. Its structured approach is designed to minimize bias and to be reproducible.

**Label Scarcity** A common condition in machine learning for SLRs where the number of known relevant documents (positive labels) is extremely small at the beginning of the process, often just one or two.

**Cold-Start Problem** The challenge of making effective recommendations or classifications when the system has little to no initial labeled data to learn from. This is a severe form of label scarcity.

**Class Imbalance** A dataset property where the classes are not represented equally. In SLRs, this is often extreme, with relevant documents (the positive class) forming a very small minority (e.g., <5%) of the total dataset.

**Annotation** The process of assigning a label (e.g., "relevant" or "irrelevant") to a data point. In this context, it refers to an expert manually reading a paper's abstract and title to determine its relevance.

**MeSH (Medical Subject Headings)** A comprehensive, controlled vocabulary created by the National Library of Medicine (NLM) used for indexing, cataloging, and searching for biomedical and health-related information.

## A.2 Core Machine Learning and Active Learning

**Active Learning (AL)** A machine learning paradigm where the learning algorithm can interactively query a user (an "oracle") to label new data points. The goal is to achieve high accuracy with minimal labeling effort by strategically selecting which data points to label next.

**Oracle** In active learning, the source of ground-truth labels. This is typically a human expert who provides the correct classification for a queried data point.

**Query Strategy** The algorithm, strategy, or heuristic used in an active learning loop to decide which unlabeled instance(s) should be presented to the oracle for labeling next.

**Static Query Strategy** A query strategy that remains fixed throughout the entire active learning process (e.g., always using uncertainty sampling). This contrasts with an adaptive strategy.

**Query Arm Portfolio** Our term for the diverse set of query strategies available to the AutoDiscover agent. Each strategy is an "arm" that the agent can "pull."

**Uncertainty Sampling** A common query strategy that involves selecting instances for which the model is least confident in its prediction (e.g., where the predicted probability is closest to 0.5 in a binary problem).

**Diversity Sampling** A query strategy that aims to select unlabeled instances that are structurally or featurally dissimilar to the instances already in the training set, promoting broad coverage of the data space.

**BALD (Bayesian Active Learning by Disagreement)** An advanced uncertainty sampling strategy that queries the data point which is expected to cause the largest reduction in the model's posterior uncertainty. It measures the disagreement among different possible models given the current data.

[Online Learning] A machine learning paradigm in which data points arrive in a sequential manner, and the model updates its parameters incrementally with each new example. This is essential for dynamic environments, such as active learning in systematic literature reviews, where labeled data accumulates over time and retraining the model from scratch at every step is impractical.

**Pre-trained Model** A model (often a large neural network) that has been trained on a massive, general-purpose dataset (e.g., all of Wikipedia). These models can then be used as a starting point for more specialized tasks.

**Fine-tuning** The process of taking a pre-trained model and continuing its training on a smaller, task-specific dataset to adapt it to that particular task.

## A.3 Graph Theory and Graph Neural Networks

**Heterogeneous Graph** A graph that contains different types of nodes and/or different types of edges. This contrasts with a homogeneous graph, which has only one type of node and one type of edge.

**Graph Neural Network (GNN)** A class of neural networks specifically designed to perform inference on data structured as a graph, leveraging the relationships (edges) between data points (nodes).

**Message Passing** The core operational principle of most GNNs, where nodes iteratively aggregate feature information (messages) from their neighbors to update their own feature representations (embeddings).

**Graph Attention Network (GAT)** A specific type of GNN that uses a self-attention mechanism to allow nodes to assign different levels of importance (attention weights) to their neighbors during the message passing process.

**HAN (Heterogeneous graph Attention Network)** An extension of GAT designed for heterogeneous graphs. It learns distinct attention mechanisms for each type of edge, allowing it to weigh the importance of different kinds of relationships differently.

**Node Embedding** A low-dimensional vector representation of a node in a graph, learned by a GNN. This vector aims to capture the node's features and its position within the graph's topology.

**Label Propagation / Spreading** A family of semi-supervised learning algorithms that work on graphs. They operate by propagating labels from a small set of labeled nodes to the rest of the graph, assuming that connected nodes are likely to share the same label.

## A.4 Reinforcement Learning and Bandit Theory

**Reinforcement Learning (RL)** A broad area of machine learning where an "agent" learns to make a sequence of decisions in an "environment" to maximize a cumulative reward signal.

**Agent** An autonomous entity in an RL problem that perceives its environment and learns to take actions to achieve its goals. In AutoDiscover, the DTS controller is the agent.

**Multi-Armed Bandit (MAB)** A classic reinforcement learning problem that exemplifies the exploration-exploitation trade-off. An agent must choose between multiple "arms" (e.g., slot machines), each with an unknown reward probability, to maximize its cumulative reward.

**Exploration-Exploitation Dilemma** The fundamental trade-off in reinforcement learning between exploiting actions that are known to yield high rewards and exploring new actions to discover potentially better rewards for the future.

**Thompson Sampling (TS)** A Bayesian algorithm for solving the MAB problem. It maintains a probability distribution (a belief) over the expected reward of each arm and makes decisions by sampling from these beliefs.

**Posterior Distribution** In Bayesian statistics, the updated probability distribution for a parameter after taking into account observed data. In TS, this is the agent's belief about an arm's reward rate after observing its past successes and failures.

**Beta Distribution** A family of continuous probability distributions defined on the interval [0, 1], parameterized by two positive shape parameters, $\alpha$ and $\beta$. It is the standard way to model the posterior belief about a binomial probability, making it perfect for TS with binary rewards.

**Stationary Environment** An environment in which the underlying rules or reward distributions remain constant over time.

**Non-stationary Environment** An environment where the underlying reward probabilities or dynamics change over time. The active learning process is non-stationary because the utility of query strategies changes as the model learns.

**Discounted Thompson Sampling (DTS)** An extension of Thompson Sampling designed for non-stationary environments. It uses a discount factor, $\gamma$, to give more weight to recent observations, allowing the agent to "forget" outdated information and adapt to changes.

**Discount Factor ($\gamma$)** A parameter in the range (0, 1] used in Discounted Thompson Sampling to control the agent's memory. It systematically reduces the weight of past observations, forcing the agent to prioritize recent feedback and adapt to a non-stationary environment.

**Regret** In bandit theory, a measure of how much worse an agent performed compared to an optimal agent that knew the best arm from the beginning. Minimizing regret is the theoretical goal of bandit algorithms.

## A.5 Evaluation Metrics

**Discovery Rate Efficiency (DRE)** A metric measuring the overall efficiency of a screening process. It is the ratio of the recall (fraction of relevant items found) to the proportion of the dataset screened. A DRE of 5.0 means the system was five times more efficient than random screening.

**Work Saved over Sampling (WSS@p)** A metric that quantifies the proportion of work saved by a model compared to random sampling to achieve a certain percentage of recall, $p$. For example, WSS@95 measures the saved effort to find 95% of all relevant documents. A value of 0.90 means 90

**Recall@k** A metric that measures the proportion of all relevant documents that have been found after screening a fixed number ($k$) of documents.

**Time to Discovery (TD)** The number of documents that needed to be screened to find a specific relevant document.

**Average Time to Discovery (ATD)** The average TD across all discovered relevant documents, typically expressed as a percentage of the total dataset size for comparability. A lower ATD indicates higher efficiency.

**Relevant References Found (RRF@p)** A metric equivalent to Recall@k, but where k is a percentage of the total dataset size. For example, RRF@10 is the recall after screening 10% of the entire dataset.

## A.6 Models and Technologies

**SPECTER2** A pre-trained transformer model specifically designed to produce high-quality embeddings for scientific documents, capturing semantic meaning and awareness of citation context.

**TF-IDF (Term Frequency-Inverse Document Frequency)** A classical statistical measure used to evaluate how important a word is to a document in a collection or corpus. It is a common baseline for text representation.

**Optuna** A hyperparameter optimization framework used in this thesis to automatically search for the best-performing model configurations.

**TPE (Tree-structured Parzen Estimator)** An efficient Bayesian optimization algorithm used by Optuna to guide the hyperparameter search process more intelligently than random search.

**UMAP (Uniform Manifold Approximation and Projection)** A dimensionality reduction technique used for visualizing high-dimensional data, such as document embeddings, in a 2D or 3D space while preserving both local and global structure.

**TS-Insight** The open-source visual analytics dashboard developed as part of this thesis to provide explainability for Thompson Sampling-based agents.

# B Detailed Query Arm Portfolio: Principles, Heuristics, and Implementation

The Discounted Thompson Sampling (DTS) agent in AutoDiscover, as detailed in Section 5.3.2, adaptively selects from a portfolio of distinct query strategies (arms). Each arm employs a specific heuristic to identify the most informative unlabeled node(s) for annotation at a given stage of the active learning process. The following subsections detail the operational principle, formal selection heuristic, algorithmic implementation sketch (derived from AutoDiscover's internal functions), and specific adaptation notes for each arm, drawing upon established active learning literature.

## B.1 GNN Exploit Arm

**Principle:**

Selects the unlabeled paper that the current GNN model $\theta$ deems most likely to be relevant. This strategy focuses on immediate positive class discovery based on the model's strongest current beliefs, akin to a greedy approach.

**Formal Selection Heuristic:**

Given the set of unlabeled nodes $\mathcal{U}$, select $u^* \in \mathcal{U}$ such that:

$$u^* = \arg\max_{u \in \mathcal{U}} P_\theta(y_u = 1)$$

where $P_\theta(y_u = 1)$ is the probability output by the GNN model for node $u$ belonging to the positive (relevant) class.

**Algorithmic Implementation (Algorithm 1, derived from `query_exploit`):**

**Algorithm 1:** GNN Exploit Arm

**Input:** GNN output probabilities $\mathbf{P}$ (shape: $N_{nodes} \times N_{classes}$), unlabeled node mask $\mathcal{M}_{unlabeled}$

**Output:** Index of selected node $u^*$, or -1 if no valid selection

**if** $\neg\exists u$ *s.t.* $\mathcal{M}_{unlabeled}[u]$ **then return**-1 ;

$\mathcal{I}_{unlabeled} \leftarrow \{u \mid \mathcal{M}_{unlabeled}[u] \text{ is true}\}$;

$\mathbf{P}_{unlabeled,class1} \leftarrow \mathbf{P}[\mathcal{I}_{unlabeled}, 1]$ ; `// Probabilities for positive class`

$idx_{local} \leftarrow$ torch.argmax($\mathbf{P}_{unlabeled,class1}$);

$u^* \leftarrow \mathcal{I}_{unlabeled}[idx_{local}]$;

**return** $u^*$

**Adaptation Notes:**

This is a standard exploitation strategy in active learning [38], directly applied to the relevance predictions of AutoDiscover's HAN model.

## B.2 Entropy Uncertainty Arm

**Principle:**

Selects the unlabeled paper about which the GNN model has the highest predictive uncertainty, quantified by the Shannon entropy of its output probability distribution. This strategy aims to reduce overall model uncertainty by querying the most ambiguous instances.

**Formal Selection Heuristic:**

Select $u^* \in \mathcal{U}$ such that:

$$u^* = \arg\max_{u \in \mathcal{U}} H(P_\theta(y_u)), \quad \text{where } H(p) = -\sum_{c=0}^{N_{classes}-1} p_c \log_2 p_c$$

**Algorithmic Implementation (Algorithm 2, derived from `query_uncertainty_entropy`):**

**Algorithm 2:** Entropy Uncertainty Arm

**Input:** GNN output probabilities $\mathbf{P}$, unlabeled node mask $\mathcal{M}_{unlabeled}$, small constant $\epsilon$
**Output:** Index of selected node $u^*$, or -1 if no valid selection

**if** $\neg\exists u$ *s.t.* $\mathcal{M}_{unlabeled}[u]$ **then return** -1 ;
$\mathcal{I}_{unlabeled} \leftarrow \{u \mid \mathcal{M}_{unlabeled}[u] \text{ is true}\}$;
$\mathbf{P}_{unlabeled} \leftarrow \mathbf{P}[\mathcal{I}_{unlabeled}, :]$ ;
$\mathbf{E}_{unlabeled} \leftarrow -\sum_c \mathbf{P}_{unlabeled}[:, c] \cdot \text{torch.log}(\mathbf{P}_{unlabeled}[:, c] + \epsilon)$ ; // $\epsilon \approx 10^{-12}$
$idx_{local} \leftarrow \text{torch.argmax}(\mathbf{E}_{unlabeled})$;
$u^* \leftarrow \mathcal{I}_{unlabeled}[idx_{local}]$;
**return** $u^*$

**Adaptation Notes:**

A classical uncertainty sampling method [27], widely used in active learning for text classification.

## B.3 Margin Uncertainty Arm

**Principle:**

Selects the unlabeled paper where the GNN model is least confident in distinguishing between its top two predicted classes, i.e., the instance with the smallest margin between the highest and second-highest class probabilities. This targets samples near the decision boundary.

**Formal Selection Heuristic:**

Select $u^* \in \mathcal{U}$ such that:

$$u^* = \arg\min_{u \in \mathcal{U}} (P_\theta(y_u = c_1) - P_\theta(y_u = c_2))$$

where $c_1$ and $c_2$ are the classes with the highest and second-highest predicted probabilities for node $u$, respectively.

**Algorithmic Implementation (Algorithm 3, derived from `query_margin`):**

**Algorithm 3:** Margin Uncertainty Arm

**Input:** GNN output probabilities $\mathbf{P}$, unlabeled node mask $\mathcal{M}_{unlabeled}$
**Output:** Index of selected node $u^*$, or -1 if no valid selection

**if** $\neg\exists u$ *s.t.* $\mathcal{M}_{unlabeled}[u]$ **then return** -1 ;
$\mathcal{I}_{unlabeled} \leftarrow \{u \mid \mathcal{M}_{unlabeled}[u] \text{ is true}\}$;
$\mathbf{P}_{unlabeled} \leftarrow \mathbf{P}[\mathcal{I}_{unlabeled}, :]$ ;
$\mathbf{P}_{top2}, _ \leftarrow \text{torch.topk}(\mathbf{P}_{unlabeled}, 2, \text{dim} = 1)$;
$\mathbf{M}_{unlabeled} \leftarrow \mathbf{P}_{top2}[:, 0] - \mathbf{P}_{top2}[:, 1]$;
$idx_{local} \leftarrow \text{torch.argmin}(\mathbf{M}_{unlabeled})$;
$u^* \leftarrow \mathcal{I}_{unlabeled}[idx_{local}]$;
**return** $u^*$

**Adaptation Notes:**

A standard uncertainty sampling strategy [40, 47], adapted for probabilistic outputs from the GNN.

## B.4 BALD (Bayesian Active Learning by Disagreement) Arm

**Principle:**

Selects the unlabeled paper whose label is expected to yield the largest reduction in the model's posterior uncertainty over its parameters. This is approximated by maximizing the mutual information between model predictions and model parameters, often estimated using Monte Carlo (MC) dropout to simulate draws from the model's posterior.

**Formal Selection Heuristic:**

Select $u^* \in \mathcal{U}$ that maximizes the BALD score, approximated as the difference between the entropy of the average prediction and the average entropy of predictions over MC samples:

$$u^* = \arg\max_{u \in \mathcal{U}} \left( H\left[\mathbb{E}_{p(\omega|\mathcal{D}_{train})} P(y_u|\omega)\right] - \mathbb{E}_{p(\omega|\mathcal{D}_{train})}\left[H[P(y_u|\omega)]\right]\right)$$

where $\omega$ represents model parameters, $\mathcal{D}_{train}$ is the current labeled data, and expectations $\mathbb{E}_{p(\omega|\mathcal{D}_{train})}$ are approximated by averaging over $N_{MC}$ stochastic forward passes (e.g., with dropout enabled).

**Algorithmic Implementation (Algorithm 4, derived from `query_bald`):**

**Algorithm 4:** BALD Arm

**Input:** GNN model $\mathcal{M}_{gnn}$, graph data $\mathcal{D}_{graph}$, unlabeled mask $\mathcal{M}_{unlabeled}$, number of MC samples $N_{MC}$, device, small constant $\epsilon$

**Output:** Index of selected node $u^*$, or -1 if no valid selection

**if** $\neg \exists u$ *s.t.* $\mathcal{M}_{unlabeled}[u]$ **then return** -1 ;

$\mathcal{M}_{gnn}$.train() ; `// Enable dropout for MC sampling`

**All_MC_Probs** $\leftarrow$ empty list;

**for** $i \leftarrow 1$ **to** $N_{MC}$ **do**

  Logits, _ $\leftarrow \mathcal{M}_{gnn}(\mathcal{D}_{graph}.x_dict, \mathcal{D}_{graph}.edge_index_dict)$;

  Probs_unlabeled $\leftarrow$ F.softmax(Logits$[\mathcal{M}_{unlabeled}]$, dim = 1);

  Append Probs_unlabeled to **All_MC_Probs**;

**end**

$\mathcal{M}_{gnn}$.eval() ; `// Disable dropout`

$\mathbf{P}_{MC_stack} \leftarrow$ torch.stack(**All_MC_Probs**, dim = 0) ; `// Shape:` $[N_{MC}, N_{unlabeled}, N_{classes}]$

$\mathbf{P}_{avg_over_samples} \leftarrow$ torch.mean($\mathbf{P}_{MC_stack}$, dim = 0) ; `// Shape:` $[N_{unlabeled}, N_{classes}]$

$H_{avg_P} \leftarrow -\sum_c \mathbf{P}_{avg_over_samples}[:, c] \cdot$ torch.log($\mathbf{P}_{avg_over_samples}[:, c] + \epsilon$) ; `// Entropy of average probabilities`

$H_{each_P_sample} \leftarrow -\sum_c \mathbf{P}_{MC_stack}[:, :, c] \cdot$ torch.log($\mathbf{P}_{MC_stack}[:, :, c] + \epsilon$) ; `// Entropy for each MC prediction sample`

$Avg_{H_P} \leftarrow$ torch.mean($H_{each_P_sample}$, dim = 0) ; `// Average of entropies`

BALD_scores $\leftarrow H_{avg_P} - Avg_{H_P}$;

$\mathcal{I}_{unlabeled} \leftarrow \{u \mid \mathcal{M}_{unlabeled}[u] \text{ is true}\}$;

$idx_{local} \leftarrow$ torch.argmax(BALD_scores);

$u^* \leftarrow \mathcal{I}_{unlabeled}[idx_{local}]$;

**return** $u^*$

**Adaptation Notes:**

AutoDiscover implements the BALD strategy [22] using MC dropout with its HAN model to approximate Bayesian model uncertainty. The number of MC samples ($N_{MC}$) is a tunable hyperparameter (Table 10).

## B.5 Embedding Diversity Arm

**Principle:**

Selects the unlabeled paper whose learned GNN embedding is most dissimilar to the embeddings of all papers already labeled. This promotes exploration of diverse regions in the learned feature space, aiming to acquire labels for varied types of instances.

**Formal Selection Heuristic:**

Given the set of unlabeled nodes $\mathcal{U}$, the set of labeled nodes $\mathcal{L}$, and their GNN-derived embeddings $\mathbf{z}$, select $u^* \in \mathcal{U}$ such that:

$$u^* = \arg\max_{u \in \mathcal{U}} \left( \min_{l \in \mathcal{L}} \text{distance}(\mathbf{z}_u, \mathbf{z}_l) \right)$$

AutoDiscover defaults to using cosine distance (1 - cosine similarity) for ‘distance‘.

**Algorithmic Implementation (Algorithm 5, derived from `query_diversity_embedding`):**

**Algorithm 5:** Embedding Diversity Arm

**Input:** Paper node embeddings $\mathbf{Z}$ (shape $N_{nodes} \times D_{emb}$), labeled mask $\mathcal{M}_{train}$, unlabeled mask $\mathcal{M}_{unlabeled}$, distance metric type (`cosine` or `euclidean`)
**Output:** Index of selected node $u^*$, or -1 if no valid selection

**if** $\neg\exists u$ *s.t.* $\mathcal{M}_{unlabeled}[u]$ **then return**-1 ;
**if** $\neg\exists l$ *s.t.* $\mathcal{M}_{train}[l]$ **then**
  Select random $u \in \mathcal{U}$; **return** $u$ ; `// Fallback if no labeled yet`
**end**
$\mathcal{I}_{unlabeled} \leftarrow \{u \mid \mathcal{M}_{unlabeled}[u] \text{ is true}\}$;
$\mathcal{I}_{labeled} \leftarrow \{l \mid \mathcal{M}_{train}[l] \text{ is true}\}$;
$\mathbf{Z}_{unlabeled} \leftarrow \mathbf{Z}[\mathcal{I}_{unlabeled}]$.detach();
$\mathbf{Z}_{labeled} \leftarrow \mathbf{Z}[\mathcal{I}_{labeled}]$.detach();
**if** *metric is 'cosine'* **then**
  $\mathbf{Z}_{U_norm} \leftarrow \text{F.normalize}(\mathbf{Z}_{unlabeled}, p = 2, \text{dim} = 1)$;
  $\mathbf{Z}_{L_norm} \leftarrow \text{F.normalize}(\mathbf{Z}_{labeled}, p = 2, \text{dim} = 1)$;
  $\text{DistMatrix} \leftarrow 1.0 - \text{torch.matmul}(\mathbf{Z}_{U_norm}, \mathbf{Z}_{L_norm}.\text{t}())$;
**else if** *metric is 'euclidean'* **then**
  $\text{DistMatrix} \leftarrow \text{torch.cdist}(\mathbf{Z}_{unlabeled}, \mathbf{Z}_{labeled}, p = 2.0)$;
**end**
$\mathbf{d}_{min_to_labeled}, _ \leftarrow \text{torch.min}(\text{DistMatrix}, \text{dim} = 1)$;
$idx_{local} \leftarrow \text{torch.argmax}(\mathbf{d}_{min_to_labeled})$;
$u^* \leftarrow \mathcal{I}_{unlabeled}[idx_{local}]$;
**return** $u^*$

**Adaptation Notes:**

Implements embedding-based diversity sampling [6], utilizing the paper embeddings learned by the HAN model.

## B.6 Bias-Aware Uncertainty Arm

**Principle:**

Prioritizes querying unlabeled papers that are both highly uncertain (high entropy) and have a non-negligible GNN-predicted probability of belonging to the rare positive (relevant) class. This aims to focus uncertainty sampling on potentially fruitful candidates in imbalanced scenarios like systematic literature review screening.

**Formal Selection Heuristic:**

Selects the unlabeled node $u \in \mathcal{U}$ that maximizes the product of its GNN prediction entropy and its GNN-predicted probability of being positive:

$$u^* = \arg\max_{u \in \mathcal{U}} \left(H(P_\theta(y_u)) \times P_\theta(y_u = 1)\right)$$

**Algorithmic Implementation (Algorithm 6, derived from `query_biased_uncertainty`):**

**Algorithm 6:** Bias-Aware Uncertainty Arm

**Input:** GNN output probabilities $\mathbf{P}$, unlabeled node mask $\mathcal{M}_{unlabeled}$, small constant $\epsilon$

**Output:** Index of selected node $u^*$, or -1 if no valid selection

**if** $\neg\exists u$ *s.t.* $\mathcal{M}_{unlabeled}[u]$ **then return**-1 ;

$\mathcal{I}_{unlabeled} \leftarrow \{u \mid \mathcal{M}_{unlabeled}[u] \text{ is true}\}$;

$\mathbf{P}_{unlabeled} \leftarrow \mathbf{P}[\mathcal{I}_{unlabeled}, :]$ ;

$\mathbf{E}_{unlabeled} \leftarrow -\sum_c \mathbf{P}_{unlabeled}[:, c] \cdot \text{torch.log}(\mathbf{P}_{unlabeled}[:, c] + \epsilon)$;

$\mathbf{P}_{pos_class} \leftarrow \mathbf{P}_{unlabeled}[:, 1]$;

BiasedScore $\leftarrow \mathbf{E}_{unlabeled} \cdot \mathbf{P}_{pos_class}$;

$idx_{local} \leftarrow$ torch.argmax(BiasedScore);

$u^* \leftarrow \mathcal{I}_{unlabeled}[idx_{local}]$;

**return** $u^*$

**Adaptation Notes:**

This strategy adapts the core idea of Least Confidence with Bias (LCB) from Chen and Lin (2011) [12]. Instead of using "least confidence," AutoDiscover employs "entropy" as the uncertainty measure and multiplies it by the positive class probability to create a bias towards uncertain yet potentially relevant instances.

## B.7 Centrality-Aware Uncertainty Arm

**Principle:**

Combines model uncertainty with graph structural information by selecting unlabeled papers that are both uncertain (high entropy) and central within the literature graph (high normalized degree). This targets ambiguous nodes that are also potentially influential or representative due to their connectivity.

**Formal Selection Heuristic:**

Selects the unlabeled node $u \in \mathcal{U}$ that maximizes a weighted sum of its normalized GNN prediction entropy and its normalized (log-transformed) degree centrality:

$$u^* = \arg\max_{u \in \mathcal{U}} (w_H \cdot \text{norm}(H(P_\theta(y_u))) + w_D \cdot \text{norm}(\log(1 + \deg_{\mathcal{G}}(u))))$$

where $w_H$ and $w_D$ are predefined weights (defaulting to 0.5 each in AutoDiscover), and $\deg_{\mathcal{G}}(u)$ is the degree of node $u$ in the relevant subgraph.

**Algorithmic Implementation (Algorithm 7, derived from `query_centrality_uncertainty`):**

**Algorithm 7:** Centrality-Aware Uncertainty Arm

**Input:** GNN probabilities $\mathbf{P}$, unlabeled mask $\mathcal{M}_{unlabeled}$, pre-calculated node degrees $\mathbf{D}_{degrees}$, weights $w_H, w_D$, small constant $\epsilon$

**Output:** Index of selected node $u^*$, or -1 if no valid selection

**if** $\neg\exists u$ *s.t.* $\mathcal{M}_{unlabeled}[u]$ **then return** -1 ;
$\mathcal{I}_{unlabeled} \leftarrow \{u \mid \mathcal{M}_{unlabeled}[u]$ is true$\}$;
$\mathbf{P}_{unlabeled} \leftarrow \mathbf{P}[\mathcal{I}_{unlabeled}, :]$ ;
$\mathbf{E}_{unlabeled} \leftarrow -\sum_c \mathbf{P}_{unlabeled}[:, c] \cdot \text{torch.log}(\mathbf{P}_{unlabeled}[:, c] + \epsilon)$;
NormEnt $\leftarrow \mathbf{E}_{unlabeled}$/torch.log(torch.tensor($\mathbf{P}$.*shape*[1]));
$\mathbf{D}_{unlabeled_subset} \leftarrow \mathbf{D}_{degrees}[\mathcal{I}_{unlabeled}]$.float();
LogDeg $\leftarrow$ torch.log1p($\mathbf{D}_{unlabeled_subset}$);
MinLogDeg $\leftarrow$ torch.min(LogDeg); MaxLogDeg $\leftarrow$ torch.max(LogDeg);
**if** *MaxLogDeg* $>$ *MinLogDeg* **then**
    NormDeg $\leftarrow$ (LogDeg − MinLogDeg)/(MaxLogDeg − MinLogDeg + $\epsilon$);
**end**
**else**
    NormDeg $\leftarrow$ torch.zeros_like(LogDeg)
**end**
CombinedScore $\leftarrow (w_H \cdot \text{NormEnt}) + (w_D \cdot \text{NormDeg})$;
$idx_{local} \leftarrow$ torch.argmax(CombinedScore);
$u^* \leftarrow \mathcal{I}_{unlabeled}[idx_{local}]$;
**return** $u^*$

**Adaptation Notes:**

This arm is inspired by strategies like QUIRE [24] that combine informativeness and representativeness. AutoDiscover uses GNN-derived entropy for informativeness and normalized log-degree centrality from the heterogeneous graph for representativeness/influence.

## B.8 LP Graph Exploit (LGE) Arm

This arm directly leverages the underlying graph structure for semi-supervised label inference, operating independently of the main HAN model's predictions. It is founded on the principle of label propagation, where known labels diffuse through the graph to inform predictions for unlabeled nodes.

**Principle:**

The LGE arm embodies the idea that connected nodes in a graph are likely to share the same label [60]. It applies a Label Propagation (LP) algorithm [59] to a specific subgraph defined by a chosen edge type (e.g., semantic similarity or citation links between papers). Starting with the currently known relevant and irrelevant documents, their labels are iteratively propagated to their neighbors. The arm then exploits these propagated pseudo-labels by selecting the unlabeled paper that, according to the LP process, has received the strongest "relevant" signal. This strategy is particularly potent in early AL stages when the GNN is under-trained, as LP can effectively utilize sparse label information if the chosen graph structure is informative.

**Formal Selection Heuristic:**

Given the set of unlabeled nodes $\mathcal{U}$ and the output probabilities $\mathbf{P}_{LP}$ from the Label Propagation model run on the target subgraph, the LGE arm selects the node $u^*$ that maximizes the predicted probability of belonging to the positive class (class 1):

$$u^* = \arg\max_{u \in \mathcal{U}} P_{LP}(y_u = 1)$$

**Algorithmic Implementation (Algorithm 8):**

The implementation, detailed in Algorithm 8, utilizes the efficient Label Propagation model from PyTorch Geometric [15]. It first constructs the initial label matrix $\mathbf{Y}_{initial}$ where known labels are one-hot encoded and unlabeled nodes are initialized to zero. The LP model then iteratively updates these initial labels based on graph connectivity (defined by $E_{LP}$) and the propagation parameter $\alpha_{LP}$ over $N_{layers}$ iterations. The resulting matrix $\mathbf{P}_{LP}$ contains the propagated class probabilities for all nodes. Finally, the arm queries the unlabeled node with the highest predicted probability for the positive class.

**Algorithm 8:** LP Graph Exploit (LGE) Arm

**Input:** Graph data $\mathcal{D}_{graph}$, train mask $\mathcal{M}_{train}$, unlabeled mask $\mathcal{M}_{unlabeled}$, true labels $\mathbf{Y}_{true}$ (for $\mathcal{M}_{train}$), target edge type $E_{LP}$ (e.g., `('paper','semantic','paper')`), LP params $(N_{layers}, \alpha_{LP})$, $N_{nodes}$, $N_{classes}$, device, `assume_undirected`

**Output:** Index of selected node $u^*$, or -1 if no valid selection

**if** $\neg\exists u$ *s.t.* $\mathcal{M}_{unlabeled}[u]$ ***or*** $\neg\exists l$ *s.t.* $\mathcal{M}_{train}[l]$ **then return**-1 ;
EdgeIndex $\leftarrow \mathcal{D}_{graph}[E_{LP}]$.edge_index.to(device) ;
**if** *EdgeIndex is None* ***or*** *EdgeIndex.numel() == 0* **then return**-1 ;
**if** *assume_undirected* **then**
    EdgeIndex $\leftarrow$ PyGUtils.to_undirected(EdgeIndex, $N_{nodes}$)
**end**
EdgeIndex, _ $\leftarrow$ PyGUtils.remove_self_loops(EdgeIndex) ;
EdgeIndex, _ $\leftarrow$ PyGUtils.add_self_loops(EdgeIndex, num_nodes = $N_{nodes}$) ;
$\mathbf{Y}_{initial} \leftarrow$ torch.zeros($N_{nodes}, N_{classes}$, device = device);
$\mathbf{Y}_{initial}[\mathcal{M}_{train}] \leftarrow$ F.one_hot($\mathbf{Y}_{true}[\mathcal{M}_{train}], N_{classes}$).float();
LP_model $\leftarrow$ PyGLabelPropagation($N_{layers}, \alpha_{LP}$).to(device);
`// Propagate labels without tracking gradients`
**with** `torch.no_grad()`:;
    $\mathbf{P}_{LP} \leftarrow$ `LP_model`($\mathbf{Y}_{initial}$, `EdgeIndex`);
$\mathcal{I}_{unlabeled} \leftarrow \{u \mid \mathcal{M}_{unlabeled}[u]$ is true$\}$;
$\mathbf{P}_{LP_unlabeled} \leftarrow \mathbf{P}_{LP}[\mathcal{I}_{unlabeled}, :]$ ;
$idx_{local} \leftarrow$ torch.argmax($\mathbf{P}_{LP_unlabeled}[:, 1]$);
$u^* \leftarrow \mathcal{I}_{\text{unlabeled}}[idx_{\text{local}}]$;
**return** $u^*$

**Configuration in AutoDiscover:**

The Label Propagation model itself is based on the work of Zhou et al. (2003) [59]. In AutoDiscover, this arm defaults to using the `('paper', 'semantic', 'paper')` edge type. This choice leverages semantic similarity links derived from SPECTER2 embeddings [43], which are pre-trained with an awareness of citation networks, thereby capturing nuanced content-based relationships often correlated with scholarly impact and topical relevance. The number of propagation layers ($N_{layers}$) and the teleport probability $\alpha_{LP}$ are hyperparameters tuned via Optuna (see Table 10). By default, the chosen edge subgraph is treated as undirected. The arm exploits the LP output by selecting the unlabeled node with the highest predicted probability for the positive (relevant) class.

### B.9 Random Interactive Arm

**Principle:**

Allows for either purely random selection from the unlabeled pool or direct user intervention to select a specific document. This serves as a baseline exploration strategy and a mechanism for incorporating expert domain knowledge ad-hoc.

**Formal Selection Heuristic:**

If user-input is "random", select $u^* \sim \text{Uniform}(\mathcal{U})$. Otherwise, select user-specified $u_{user} \in \mathcal{U}$.

**Algorithmic Implementation (Algorithm 9, derived from `query_random_interactive`):**

**Algorithm 9:** Random Interactive Arm

**Input:** Unlabeled node mask $\mathcal{M}_{unlabeled}$, Total node count $N_{nodes}$
**Output:** Index of selected node $u^*$, or -1 if no valid selection

**if** $\neg\exists u$ *s.t.* $\mathcal{M}_{unlabeled}[u]$ **then return** -1 ;
$\mathcal{I}_{unlabeled} \leftarrow \{u \mid \mathcal{M}_{unlabeled}[u] \text{ is true}\}$;
**repeat**
  Input $s_{user}$ (e.g., an index or "r" for random);
  **if** $s_{user}$ == *"r"* **then**
    $idx_{local} \leftarrow$ torch.randint(0, $|\mathcal{I}_{unlabeled}|$, (1, )).item();
    $u^* \leftarrow \mathcal{I}_{unlabeled}[idx_{local}]$;
    valid ← true;
  **else if** $s_{user}$ *is a valid integer string* **then**
    $u_{attempt} \leftarrow$ Integer.parseInt($s_{user}$);
    **if** $0 \leq u_{attempt} < N_{nodes}$ ***and*** $\mathcal{M}_{unlabeled}[u_{attempt}]$ **then**
      $u^* \leftarrow u_{attempt}$;
      valid ← true;
    **end**
    **else**
      print "Invalid or already labeled index";
    **end**
  **end**
  **else**
    print "Invalid input format";
  **end**
**until** *valid selection*;
**return** $u^*$

**Adaptation Notes:**

A standard baseline and interactive querying method. While available in AutoDiscover's arm portfolio, its selection by the DTS agent in fully automated experimental runs is typically superseded by the heuristic-driven arms unless all others yield exceptionally low utility. It primarily serves as a utility for user-guided exploration or as a simple stochastic baseline.

This diverse portfolio provides the DTS agent with a rich set of options, enabling it to adapt its querying behavior dynamically to the specific challenges posed by different datasets and the evolving state of model knowledge during the active learning process.

# C Theoretical Grounding of Discounted Thompson Sampling

**Algorithm 10:** DISCOUNTED THOMPSON SAMPLING (DTS) for query-arm selection

**Input:** arms $\mathcal{K} = \{1, \ldots, K\}$,
discount $\gamma \in (0, 1]$
**foreach** $k \in \mathcal{K}$ **do**
  set prior $\theta_k \sim \text{Beta}(1, 1)$ (parameters $\alpha_k = 1, \beta_k = 1$)
**for** $t = 1, 2, \ldots$ **do**
  `// --- Thompson draw`
  **foreach** $k \in \mathcal{K}$ **do**
    sample $\tilde{\theta}_k \sim \text{Beta}(\alpha_k, \beta_k)$
  select arm $k_t \leftarrow \arg\max_k \tilde{\theta}_k$
  apply strategy $k_t$ and receive reward $r_t \in \{0, 1\}$
  `// --- discounted posterior update for arm` $k_t$
  $\alpha_{k_t} \leftarrow \gamma \alpha_{k_t} + r_t$
  $\beta_{k_t} \leftarrow \gamma \beta_{k_t} + (1 - r_t)$

**Notation.** At round $t$ the (unknown) relevance rate of arm $k$ is $\mu_{k,t} \in [0, 1]$, with best arm value $\mu_t^\star = \max_k \mu_{k,t}$. Instantaneous regret is $\Delta_t = \mu_t^\star - \mu_{k_t,t}$ and cumulative regret is $R_T = \sum_{t=1}^{T} \Delta_t$. Non-stationarity is measured by the $\ell_1$ variation budget $V_T = \sum_{t=2}^{T} \|\boldsymbol{\mu}_t - \boldsymbol{\mu}_{t-1}\|_1$.

## C.1 Regret-theoretic motivation.

The DTS meta-agent in AUTODISCOVER is grounded in regret- minimisation results for *non-stationary* multi-armed bandits [18,37]. Let $V_T = \sum_{t=2}^{T} \|\boldsymbol{\mu}_t - \boldsymbol{\mu}_{t-1}\|_1$ denote the $\ell_1$–variation of the arm means and choose a discount factor $\gamma \in (0, 1)$ that matches the time-scale of drift. Under the standard bounded-drift assumptions of [18], Discounted Thompson Sampling satisfies

$$\mathbb{E}[\text{Reg}_T] \; = \; \widetilde{\mathcal{O}}\big(\sqrt{K\,T\,V_T} + K\big),$$

where $\text{Reg}_T$ is cumulative regret, $K$ the number of arms, and $\widetilde{\mathcal{O}}$ hides logarithmic factors. Hence the agent adapts to changing arm utilities with only $\widetilde{\mathcal{O}}\big(\sqrt{T\,V_T}\big)$ excess irrelevant screens over an oracle that always picks the best strategy. Although $V_T$ is hard to measure exactly in active-learning loops, the bound formalises why AUTODISCOVER can keep reallocating queries once early "cold-start" heuristics lose their edge.

**Theorem C.1** (Adaptive regret of DTS under drift)**.** *Run Algorithm 10 with* $\gamma = 1 - \frac{1}{H}$, $H = \left\lceil \sqrt{KT/\max\{1, V_T\}} \right\rceil$ *and flat* $\text{Beta}(1, 1)$ *priors. Assume a minimal positive gap:*
$\Delta_{\min} = \min_{t,k:\mu_{k,t} < \mu_t^\star} (\mu_t^\star - \mu_{k,t}) > 0$. *Then*

$$\boxed{\mathbb{E}[R_T] \; = \; \widetilde{\mathcal{O}}\big(\sqrt{KTV_T} + K\big)}$$

*where* $\widetilde{\mathcal{O}}(\cdot)$ *hides logarithmic factors in* $K$*,* $T$ *and* $1/\Delta_{\min}$*.*

*Sketch.* (i) Partition $[1, T]$ into blocks of length $H$; because $\gamma = 1 - 1/H$, each block's sufficient statistics depend on at most the previous $H$ rounds. (ii) Inside one block the total drift is $\le V_T/H$, so the stationary Bernoulli-TS bound [39] gives $\widetilde{\mathcal{O}}(\sqrt{KH})$ regret plus an $\mathcal{O}(V_T/H)$ penalty. (iii) Summing over $T/H$ blocks and optimising $H$ yields the claim. □

**Corollary C.2** (Screening overhead)**.** *Each unit of regret is one avoidable irrelevant screen; hence DTS screens at most* $\widetilde{\mathcal{O}}\big(\sqrt{KT\,V_T}\big)$ *additional irrelevant papers beyond an oracle, and collapses to the classical* $\widetilde{\mathcal{O}}(\sqrt{KT})$ *bound when* $V_T = 0$*.*

*Remark* C.1. The first term in Theorem C.1 reacts to non-stationarity ($V_T$), while the additive $K$ is the one-shot cost of learning the arm priors.

# D Metric Calculation Algorithms

This appendix provides the detailed pseudocode for the algorithms used to calculate the primary performance metrics reported in this thesis. These algorithms operate on the discovery history logs generated by the active learning runs.

**Algorithm 11:** Compute Discovery-Rate Efficiency (DRE)

**Input:** Discovery percentages $R[1 \dots T]$ (in %), queried counts $Q[1 \dots T]$, total publications $N$
**Output:** Discovery-Rate Efficiency DRE

$R_{\text{final}} \leftarrow R[T]$;
; `// Final discovery percentage (0{100%)`
$S_{\text{final}} \leftarrow \big(Q[T]/N\big) \times 100$;
; `// Final screened percentage (0{100%)`
**if** $S_{\text{final}} > 0$ **then**
  DRE $\leftarrow \dfrac{R_{\text{final}}}{S_{\text{final}}}$;
**end**
**else**
  DRE $\leftarrow 0$;
**end**
**return** DRE

**Algorithm 12:** Compute Recall@ $k$

**Input:** discovery_curve_percent$[0 \dots T]$ ; `// percent discovered at each query count`
queried_count_history$[0 \dots T]$ ; `// cumulative items screened at each step`
$k \in \mathbb{N}$ ; `// target number of items`
**Output:** Recall@$k$

idx $\leftarrow$ searchsorted(queried_count_history, $k$);
Recall@$k \leftarrow \frac{\text{discovery_curve_percent[idx]}}{100.0}$;
**return** Recall@$k$;

**Algorithm 13:** Compute Work Saved over Sampling at $\boldsymbol{p}$**%** Recall (WSS@ $p$)

**Input:** discovery_count_history$[0 \dots T]$; `// cumulative relevants found`
queried_count_history$[0 \dots T]$; `// cumulative items screened`
$P$; `// total relevant items`
$N$; `// total items in dataset`
$p \in (0, 100]$; `// target recall percentage`
**Output:** WSS@$p$

target_discoveries $\leftarrow \lceil \frac{p}{100} \cdot P \rceil$;
idx_list $\leftarrow$ where(discovery_count_history $\geq$ target_discoveries);
**if** *idx_list is empty* **then**
  **return** 0.0 ; `// Target recall was not reached`
idx $\leftarrow$ first(idx_list);
$q_p \leftarrow$ queried_count_history[idx];

$$\text{WSS@}p \;\leftarrow\; 1.0 \;-\; \frac{q_p}{\left(\frac{p}{100}\right) N}$$

**if** WSS@$p < 0$ **then**
  **return** 0.0 ; `// Performance was worse than random`
**return** WSS@$p$;

# E Comprehensive Performance Visualizations

This appendix provides detailed performance plots for every dataset in the SYNERGY benchmark, complementing the summary results presented in Section 7. The figures are organized into two sections: first, the discovery efficiency curves, and second, an analysis of the DTS agent's query strategy outcomes per iteration. Within each section, the datasets are grouped according to whether the best-performing HPO run achieved a Work Saved over Sampling at 95% recall (WSS@95) greater than zero.

## E.1 Discovery Curves: Percentage of Relevant Documents Found

The following figures display the discovery curves (% Recall vs. % Screened) for the best HPO run of AutoDiscover on each dataset. These plots visualize the overall efficiency of the screening process.

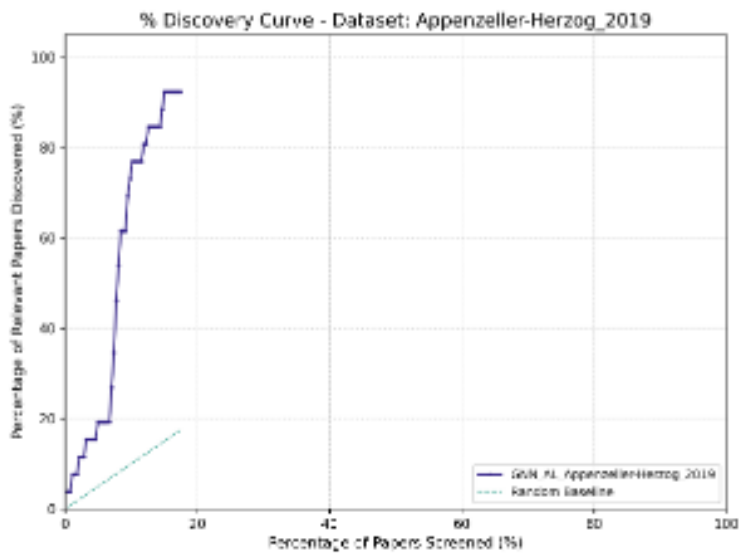

(a) Appenzeller-Herzog

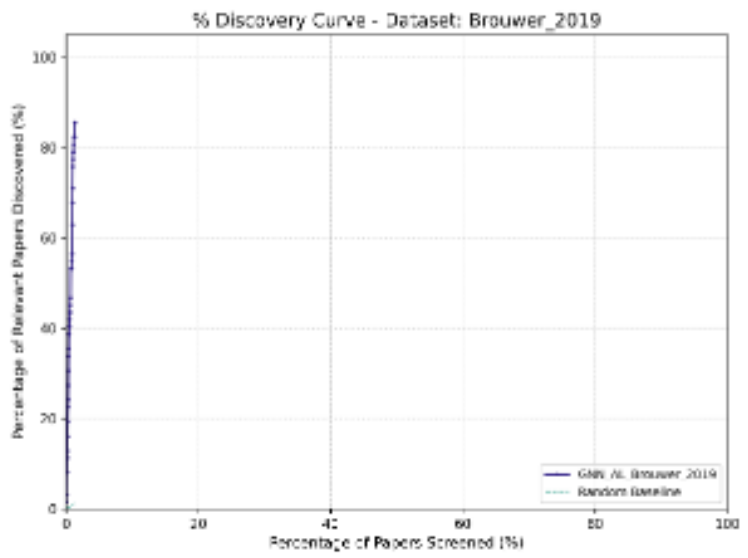

(b) Brouwer

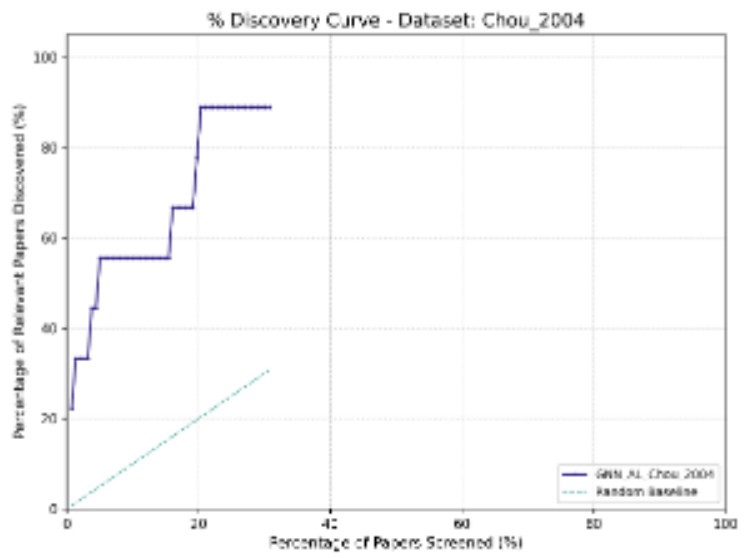

(c) Chou (2004)

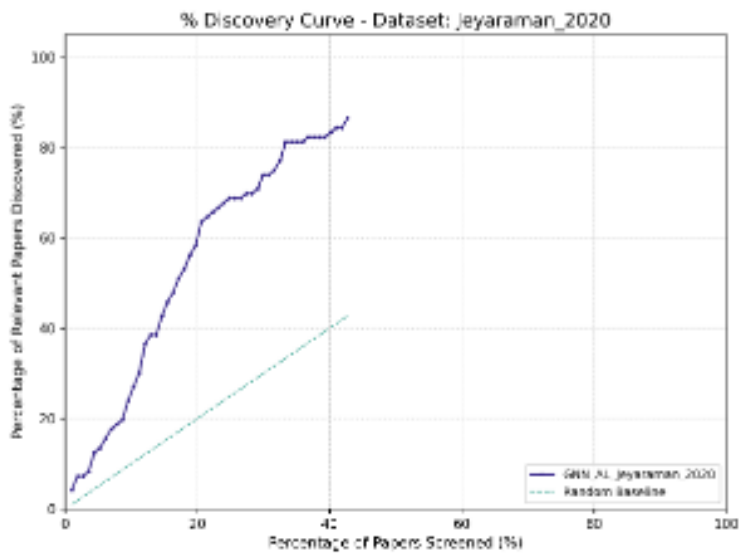

(d) Jeyaraman

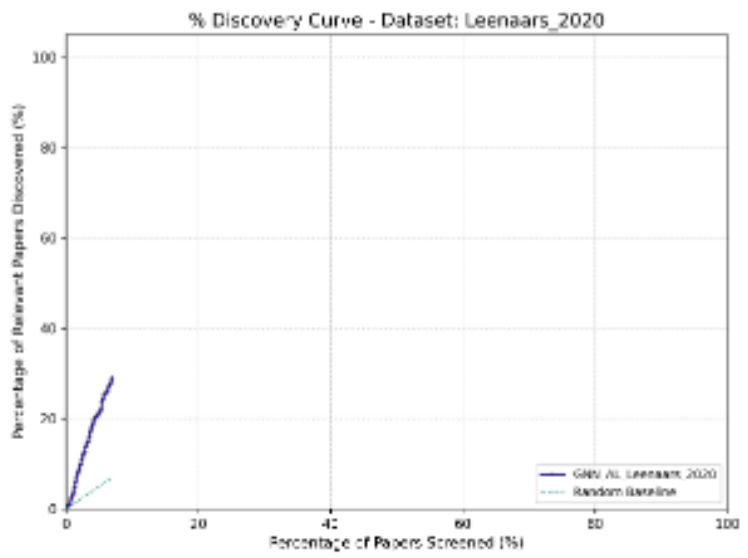

(e) Leenaars (2020)

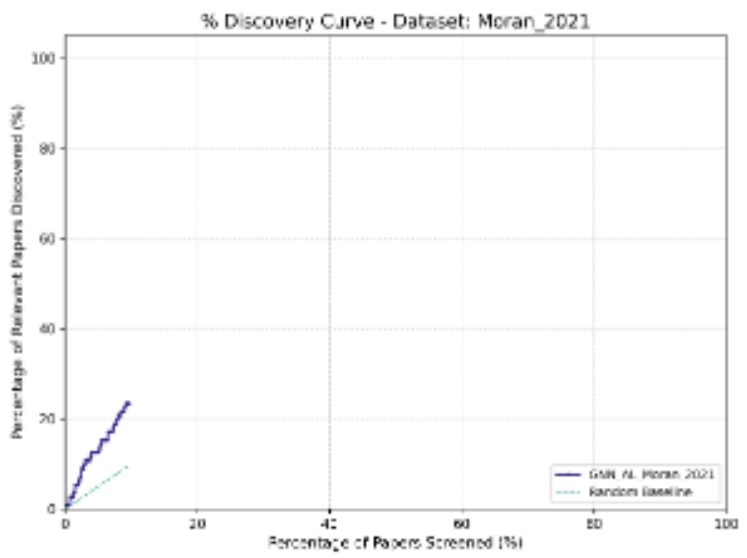

(f) Moran

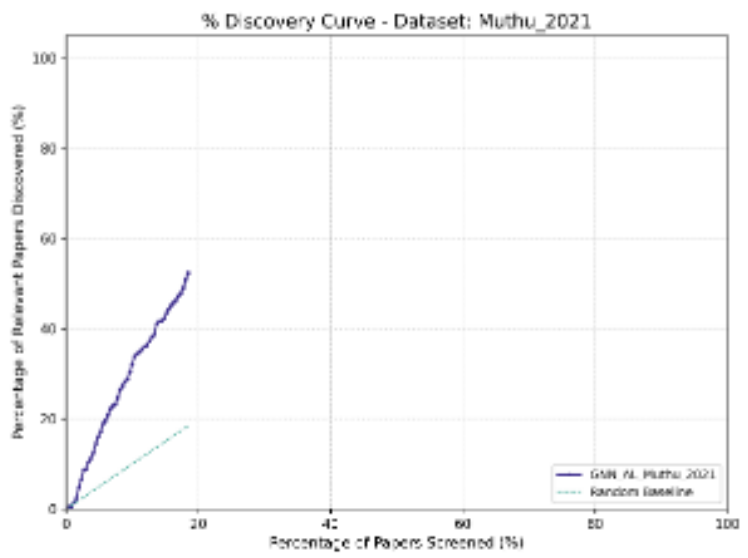

(g) Muthu

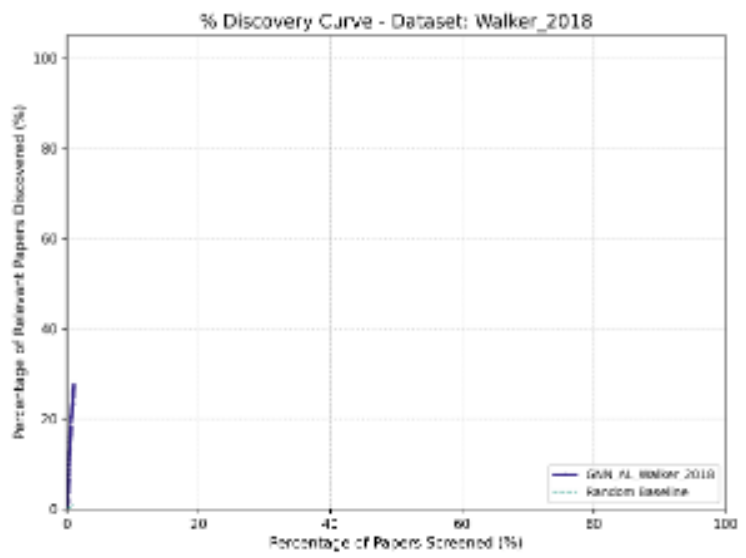

(h) Walker

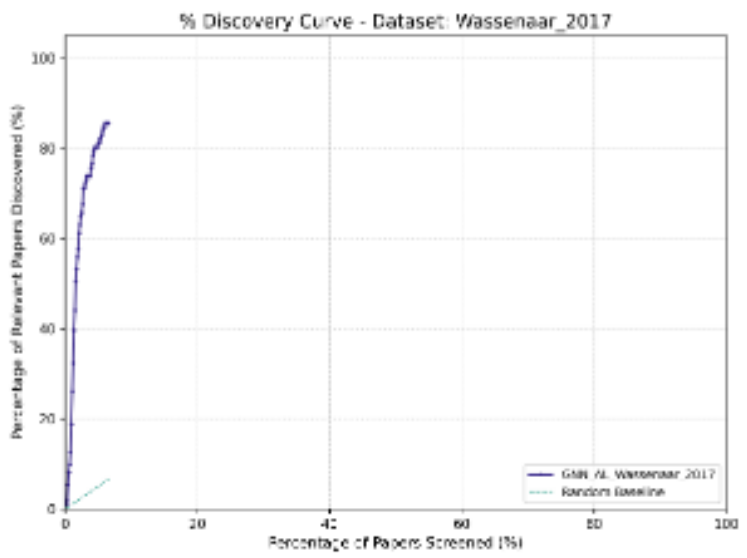

(i) Wassenaar

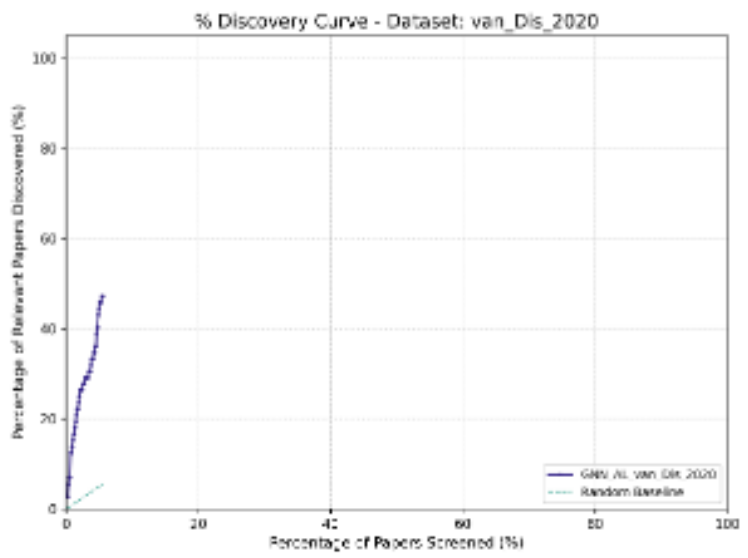

(j) van Dis

Figure 25: Discovery curves for the 10 datasets where AutoDiscover's best run did not achieve a WSS@95 greater than zero, often due to the HPO objective prioritizing overall DRE over high-recall targets within the screening limit.

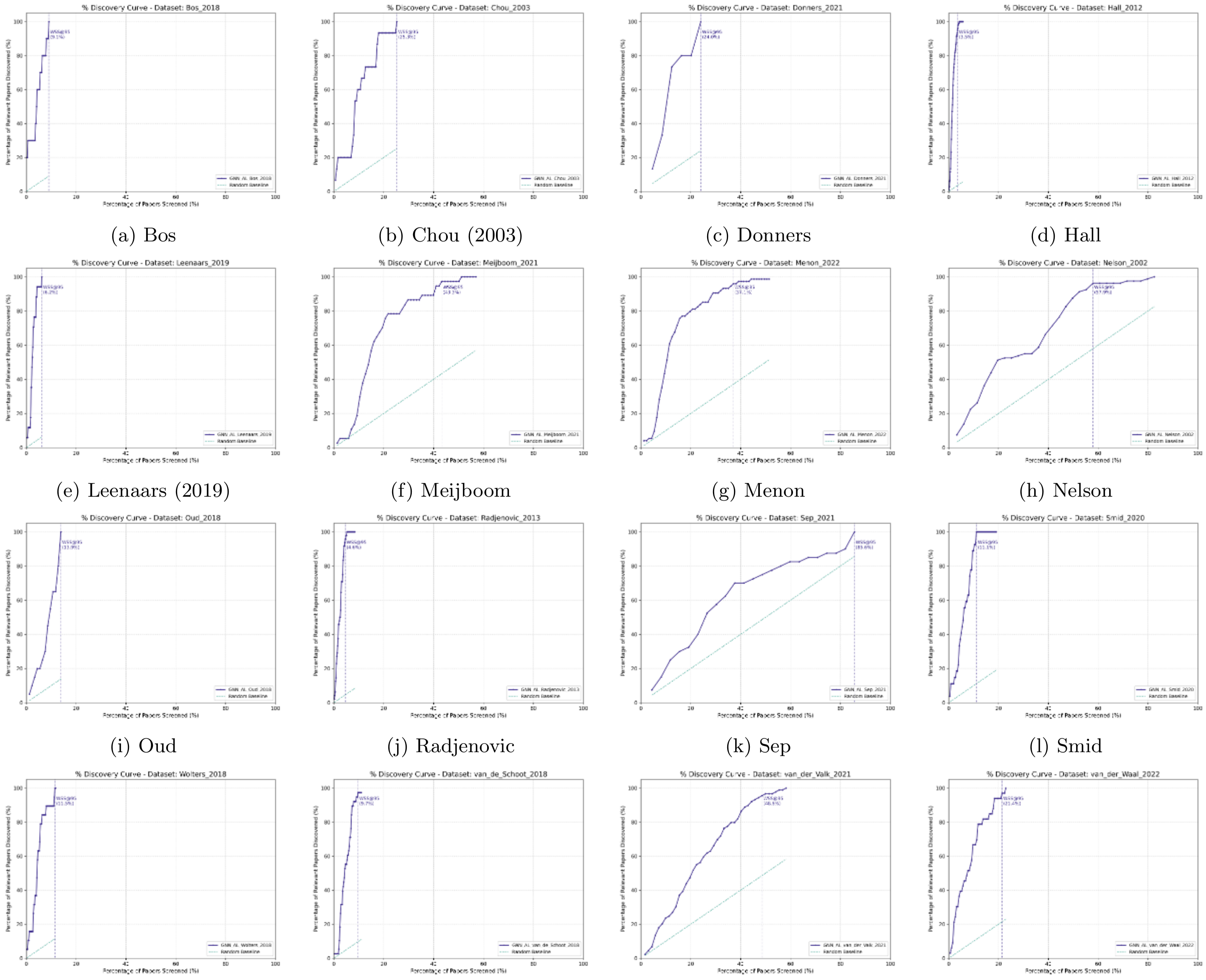

(a) Bos (b) Chou (2003) (c) Donners (d) Hall

(e) Leenaars (2019) (f) Meijboom (g) Menon (h) Nelson

(i) Oud (j) Radjenovic (k) Sep (l) Smid

(m) Wolters (n) van de Schoot (o) van der Valk (p) van der Waal

Figure 26: Discovery curves for the 16 datasets where AutoDiscover's best run achieved a positive WSS@95 score, indicating strong performance on both overall efficiency and high-recall targets.

## E.2 Arm Query Outcomes per Iteration

This appendix provides a detailed visualization of the Discounted Thompson Sampling (DTS) agent's behavior for each of the 26 SYNERGY datasets. The diverging stacked bar charts below illustrate the outcome of every query made by each arm during the 50 screening iterations (500 total queries). Upward blue bars represent the number of relevant documents found by an arm in a given iteration, while downward orange bars represent the number of irrelevant documents. These plots offer insight into which strategies the agent favored and how their effectiveness evolved over time.

## Arm Outcomes for Datasets with WSS@95 = 0

The following figures visualize the agent's behavior for the 10 datasets where the best-performing HPO run did not achieve a WSS@95 score greater than zero.

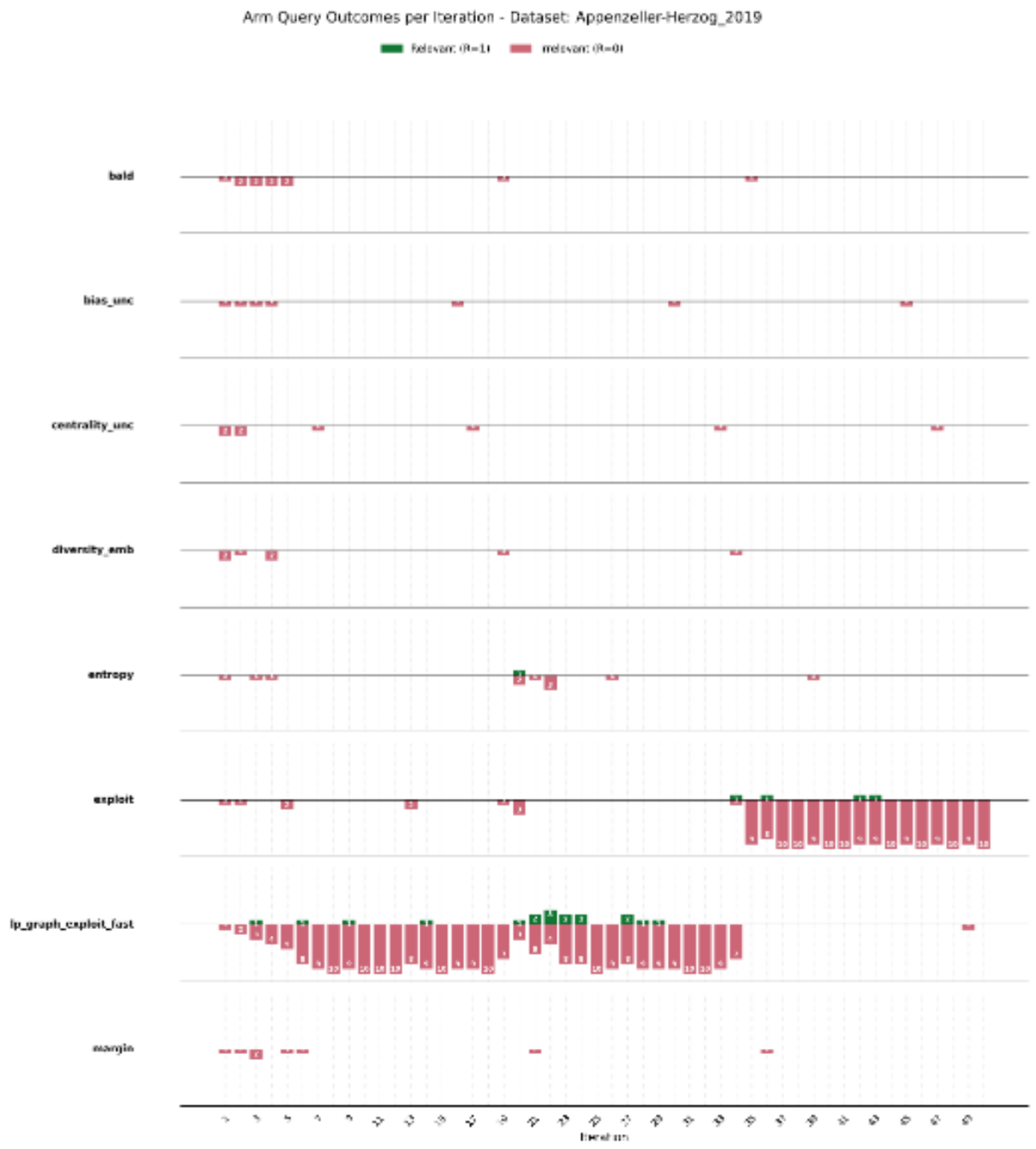

(a) Appenzeller-Herzog

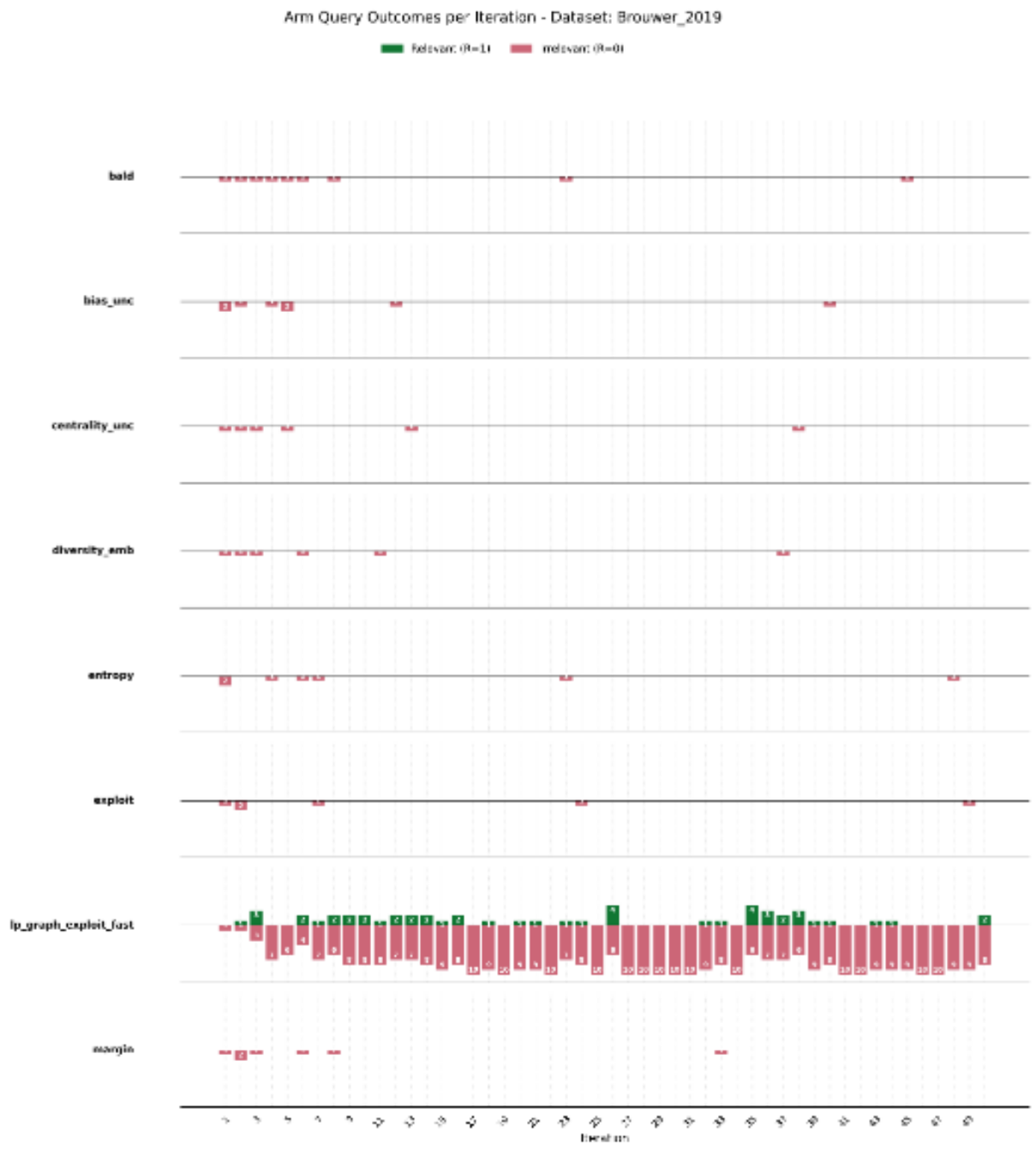

(b) Brouwer

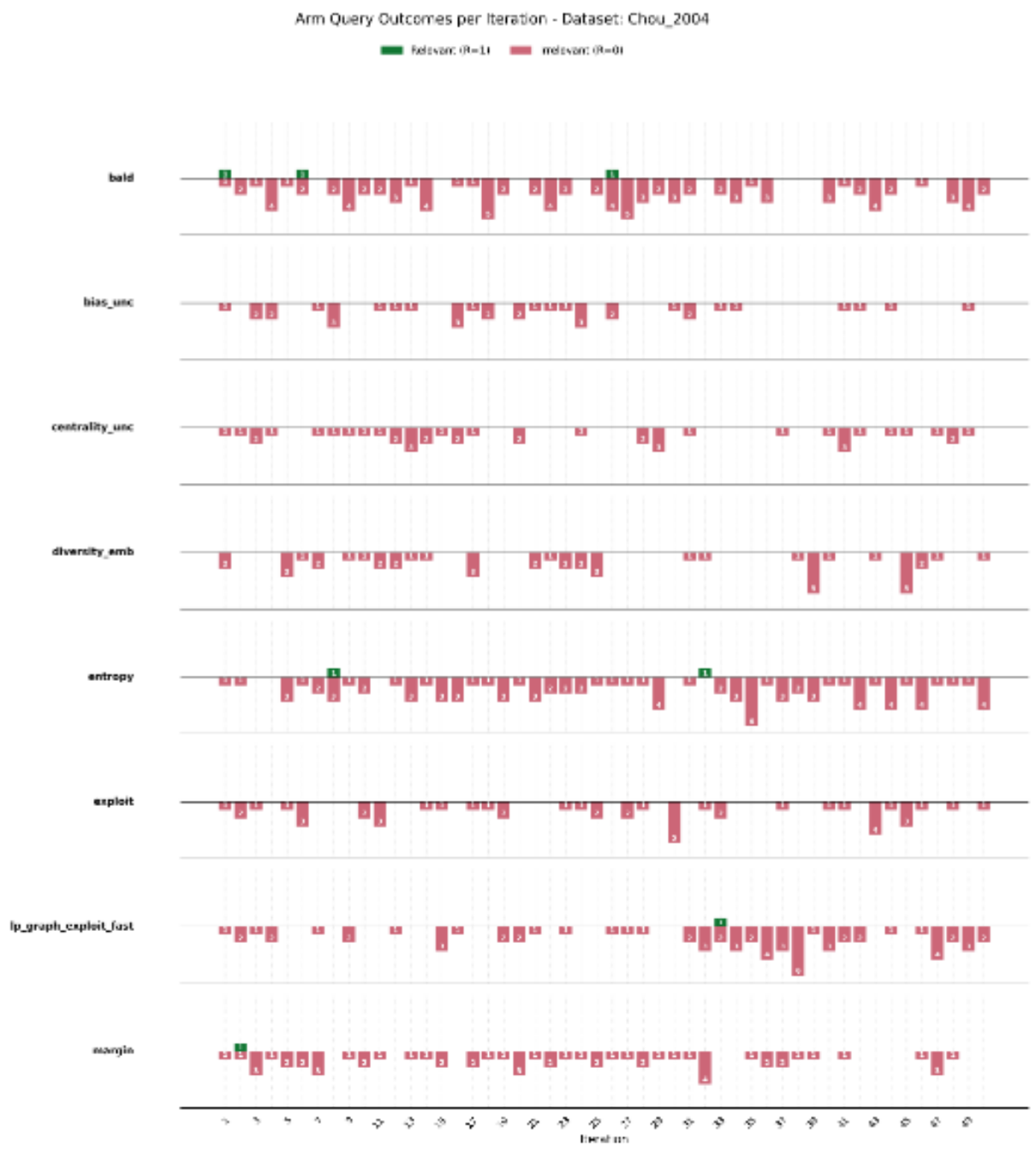

(c) Chou (2004)

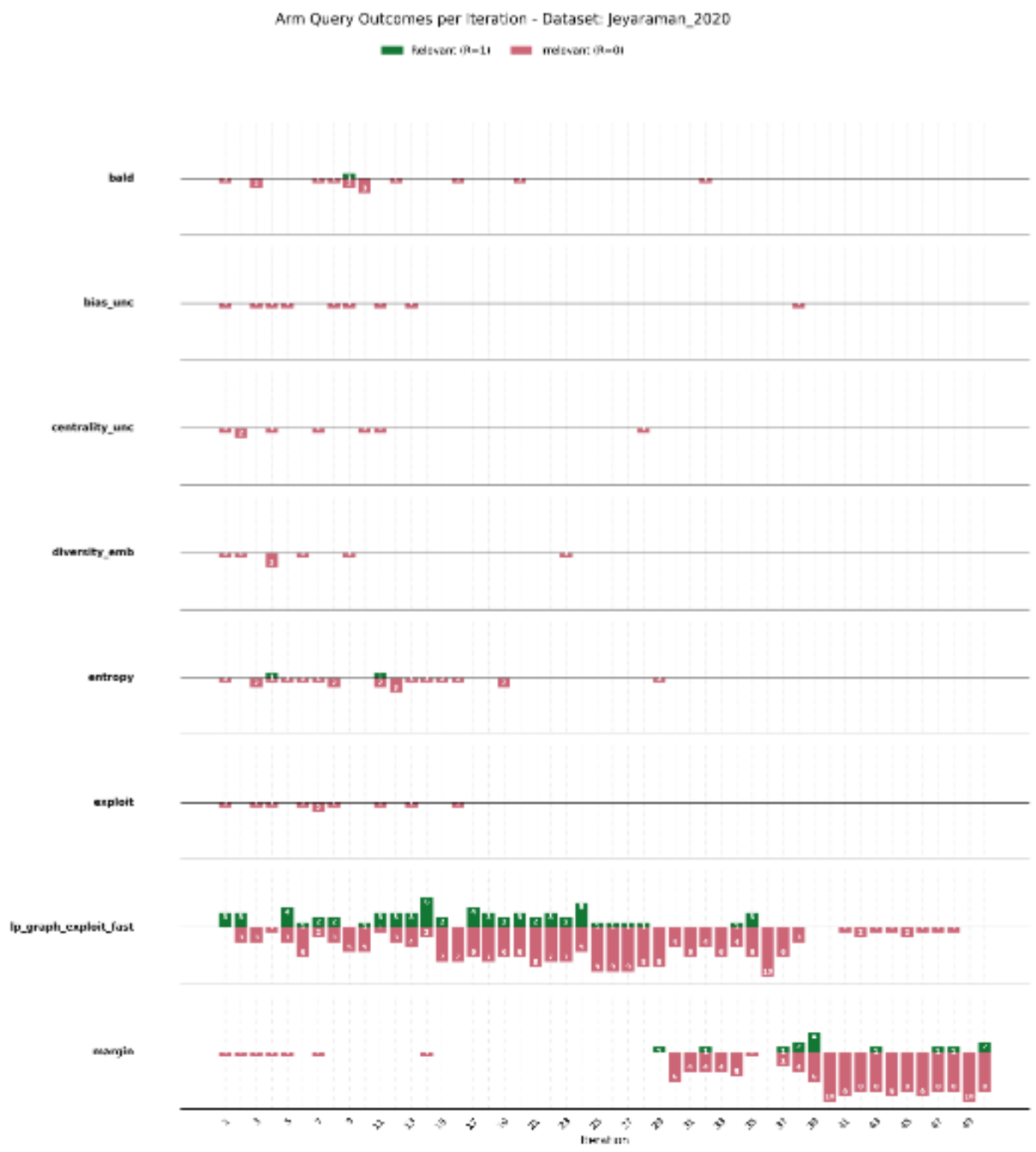

(d) Jeyaraman

Figure 27: Arm query outcomes for the first set of datasets where WSS@95 was not achieved.

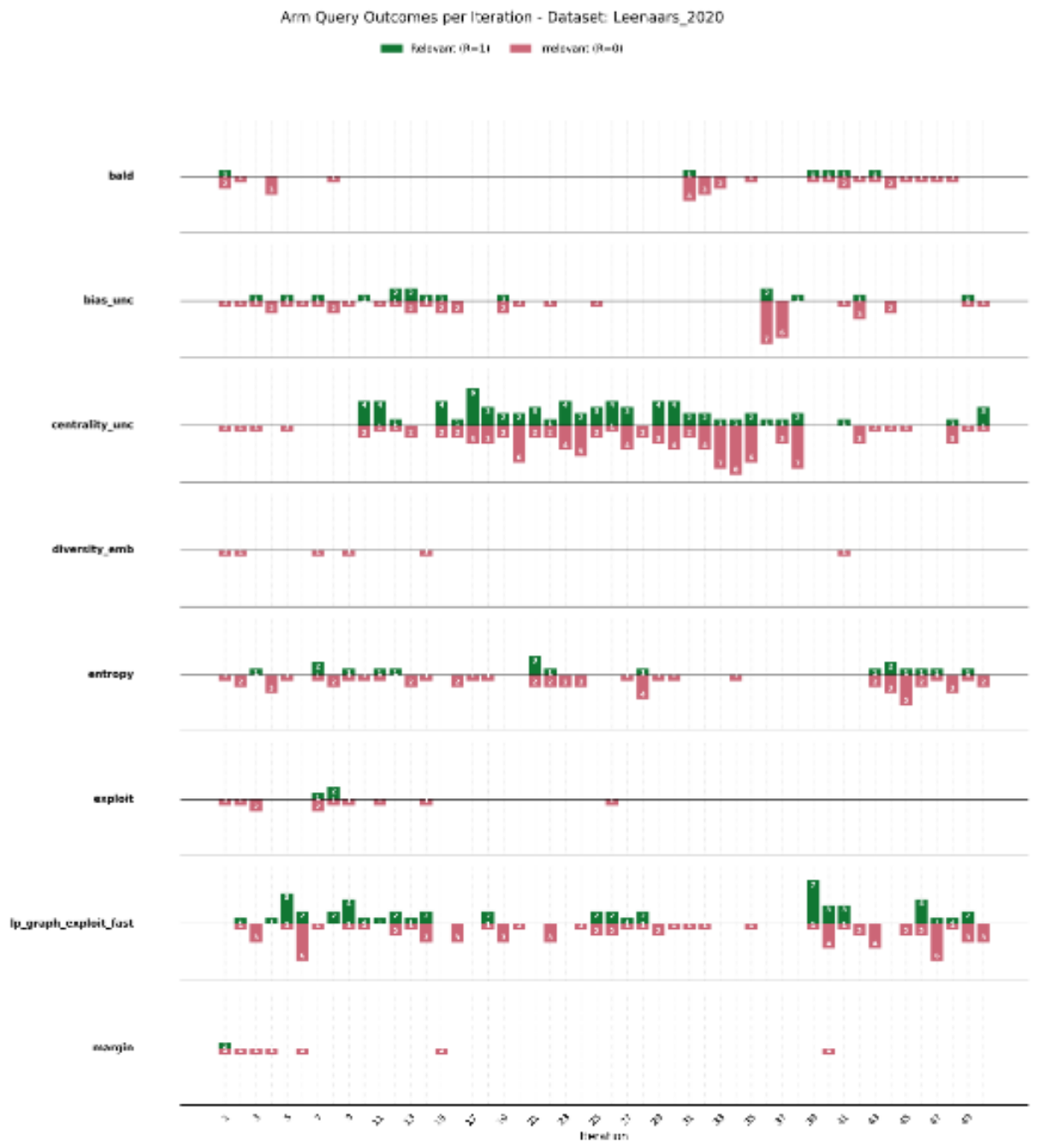

(a) Leenaars (2020)

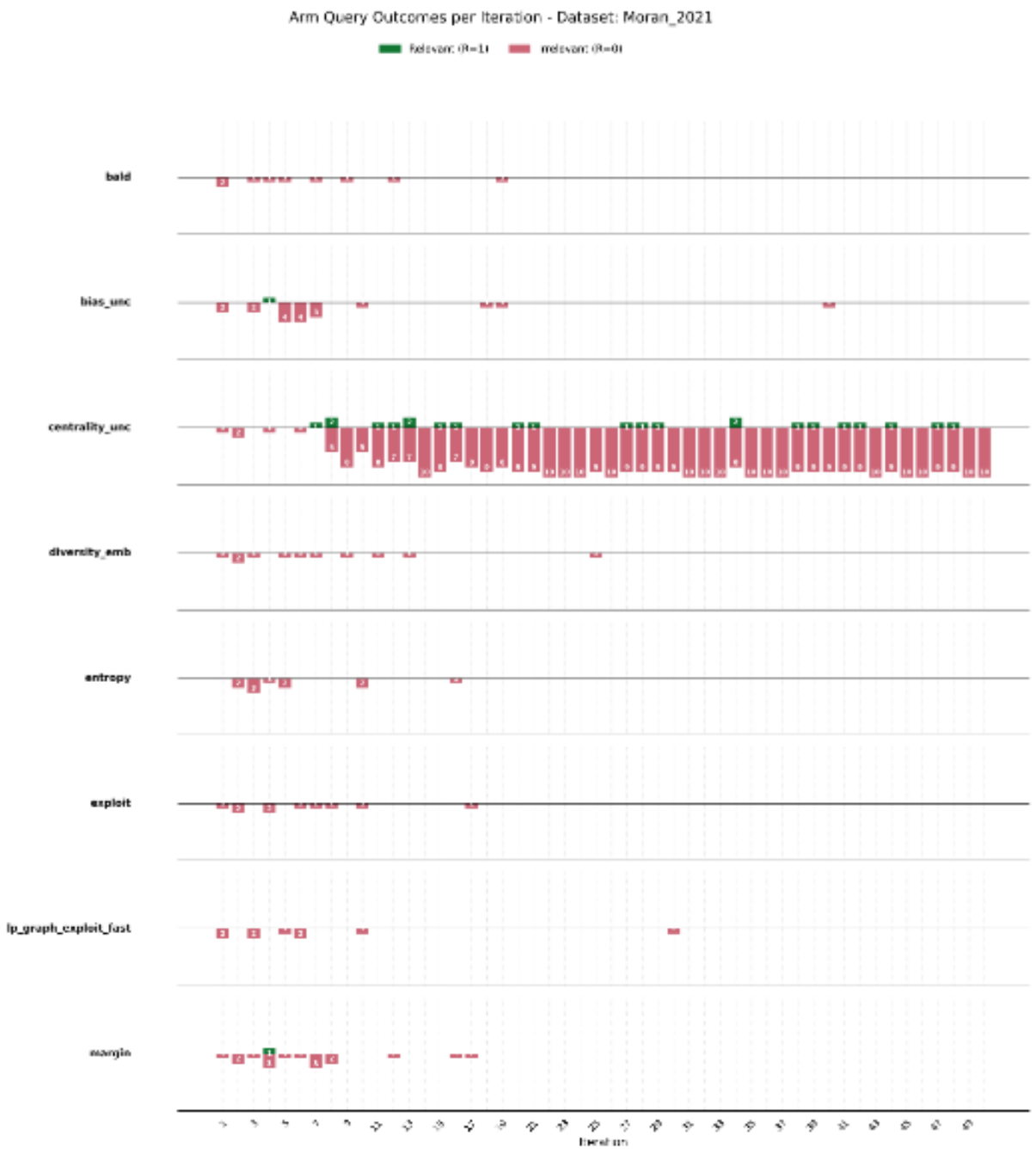

(b) Moran

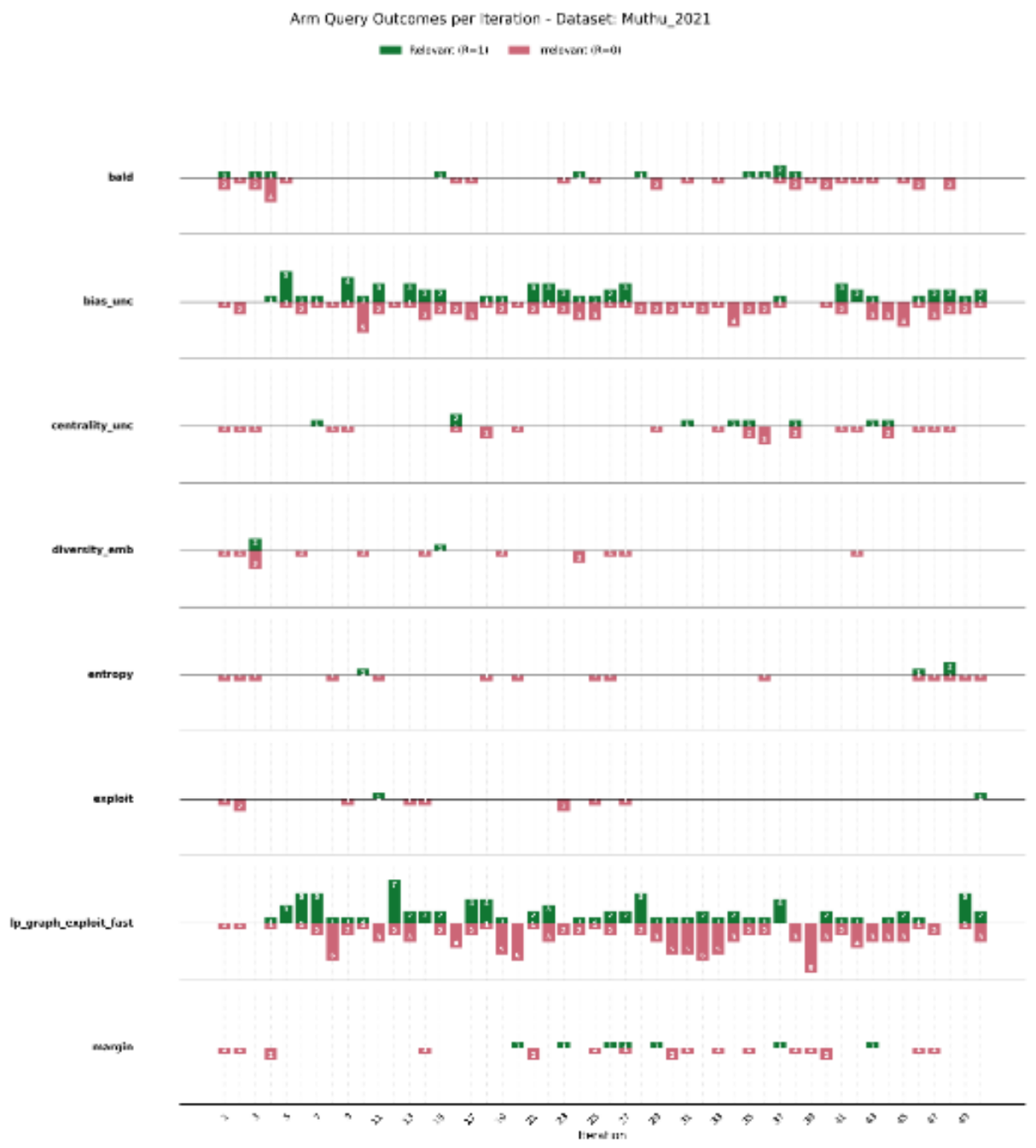

(c) Muthu

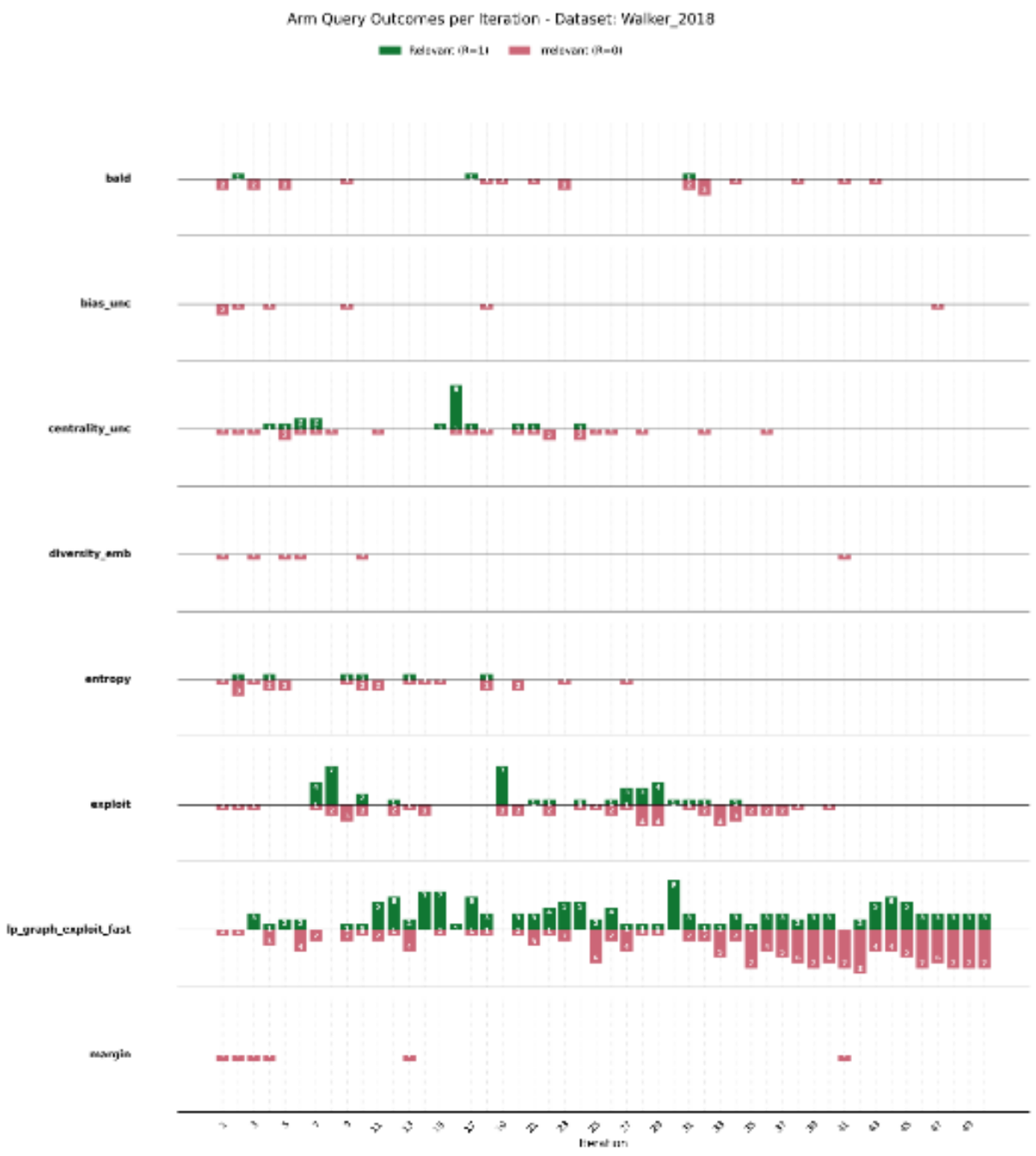

(d) Walker

Figure 28: Arm query outcomes for the second set of datasets where WSS@95 was not achieved.

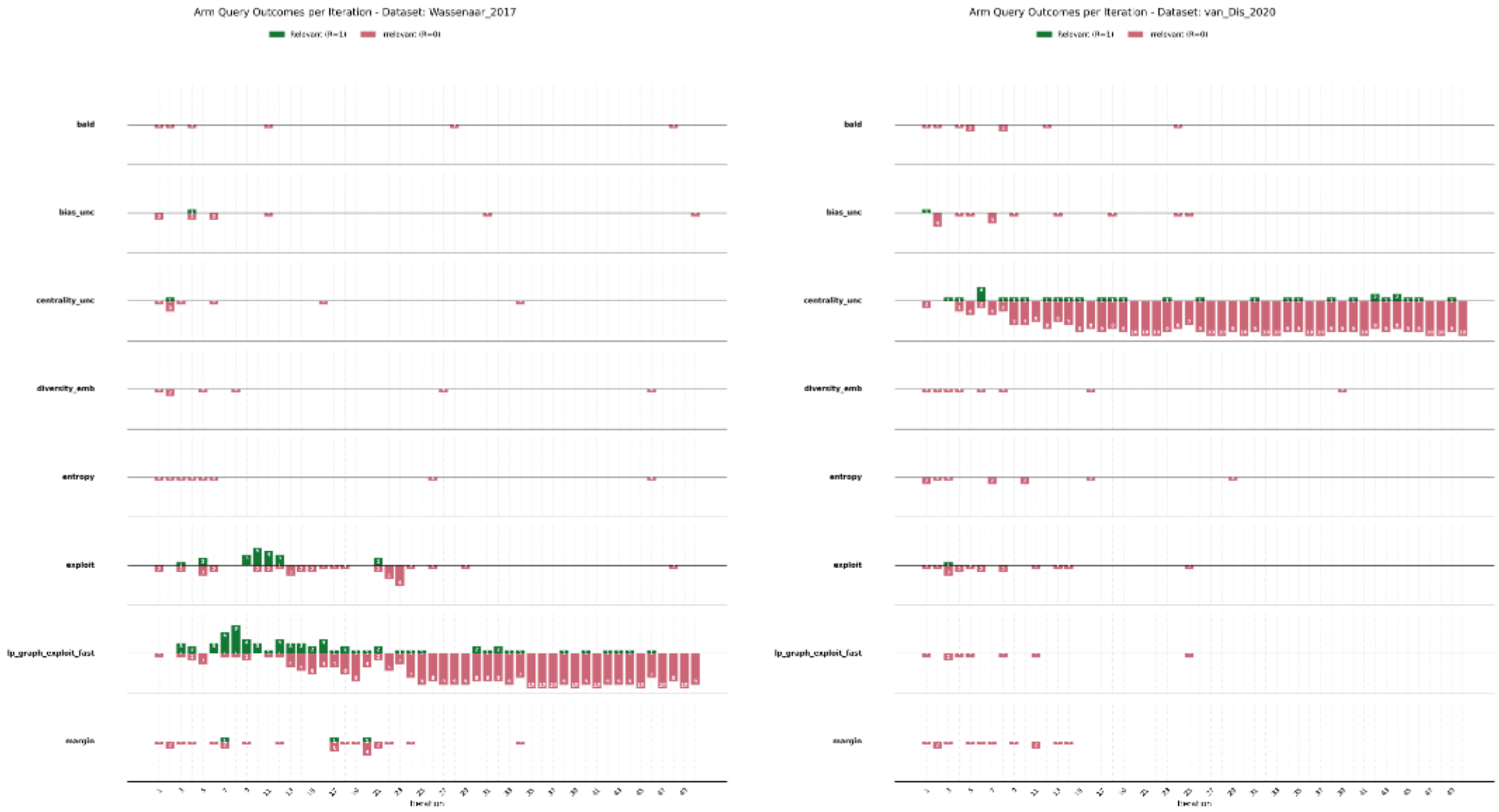

(a) Wassenaar

(b) van Dis

Figure 29: Arm query outcomes for the final set of datasets where WSS@95 was not achieved.

## Arm Outcomes for Datasets with WSS@95 > 0

The following figures visualize the agent's behavior for the 16 datasets where AutoDiscover achieved a positive WSS@95 score, indicating strong end-to-end screening performance.

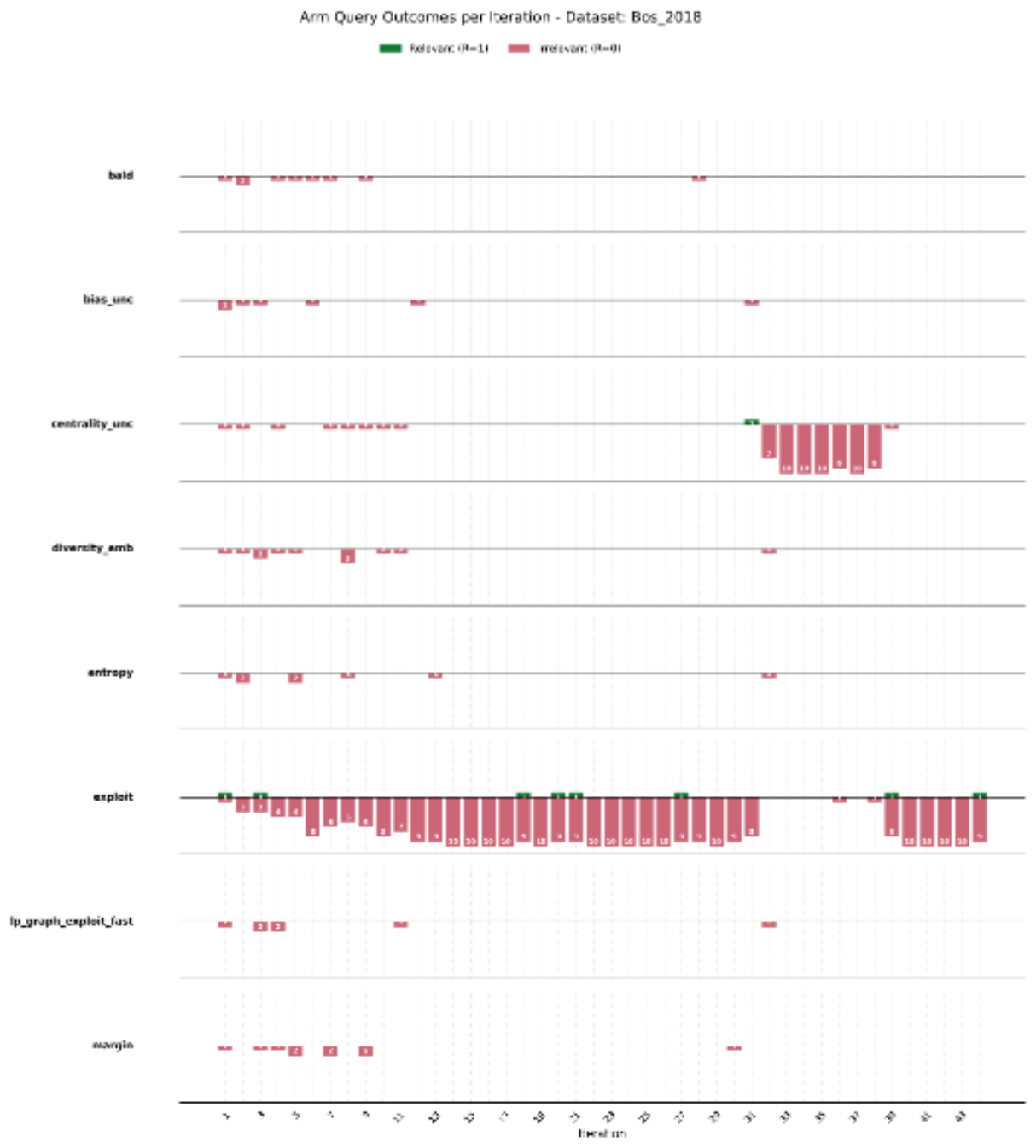

(a) Bos

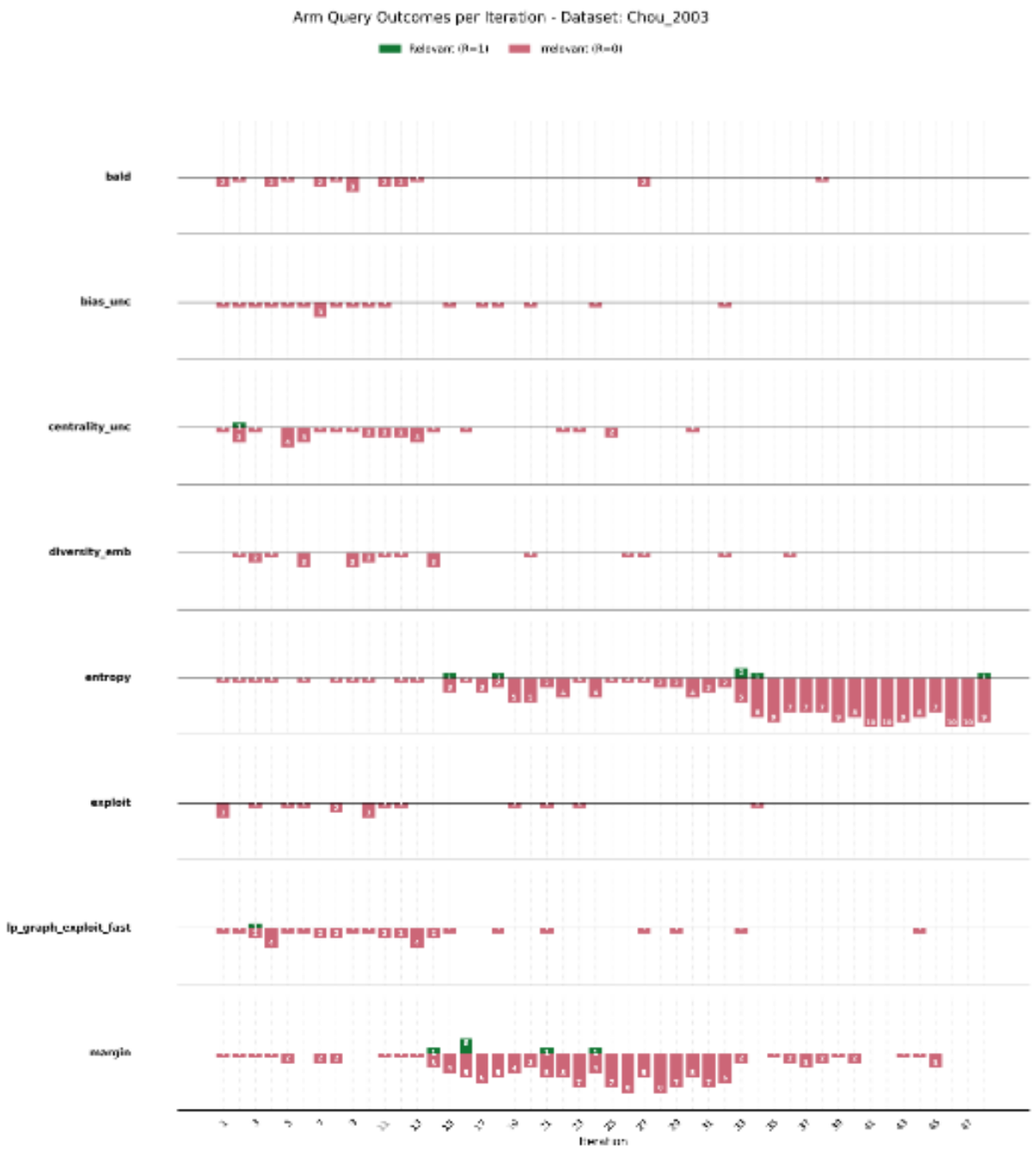

(b) Chou (2003)

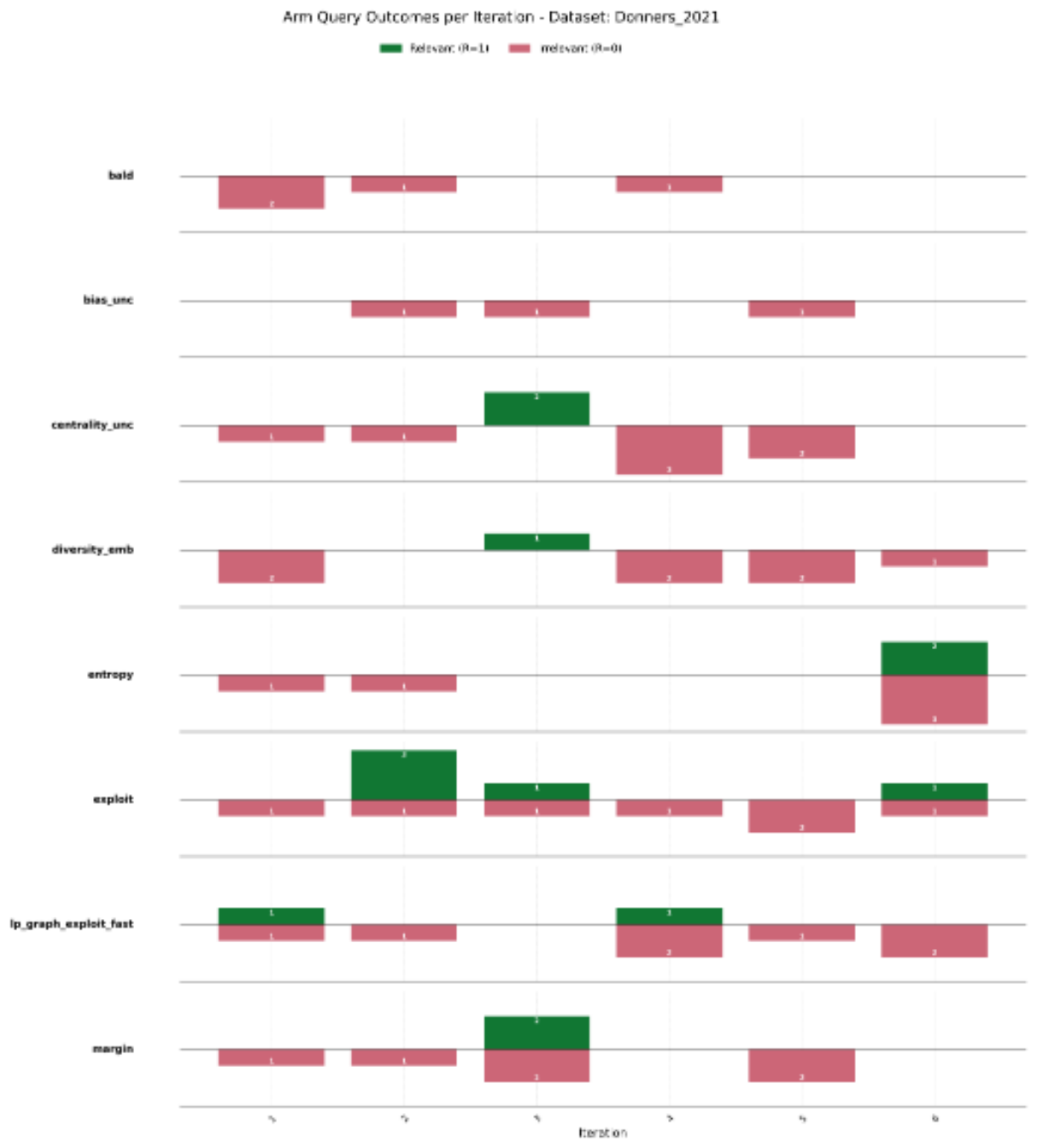

(c) Donners

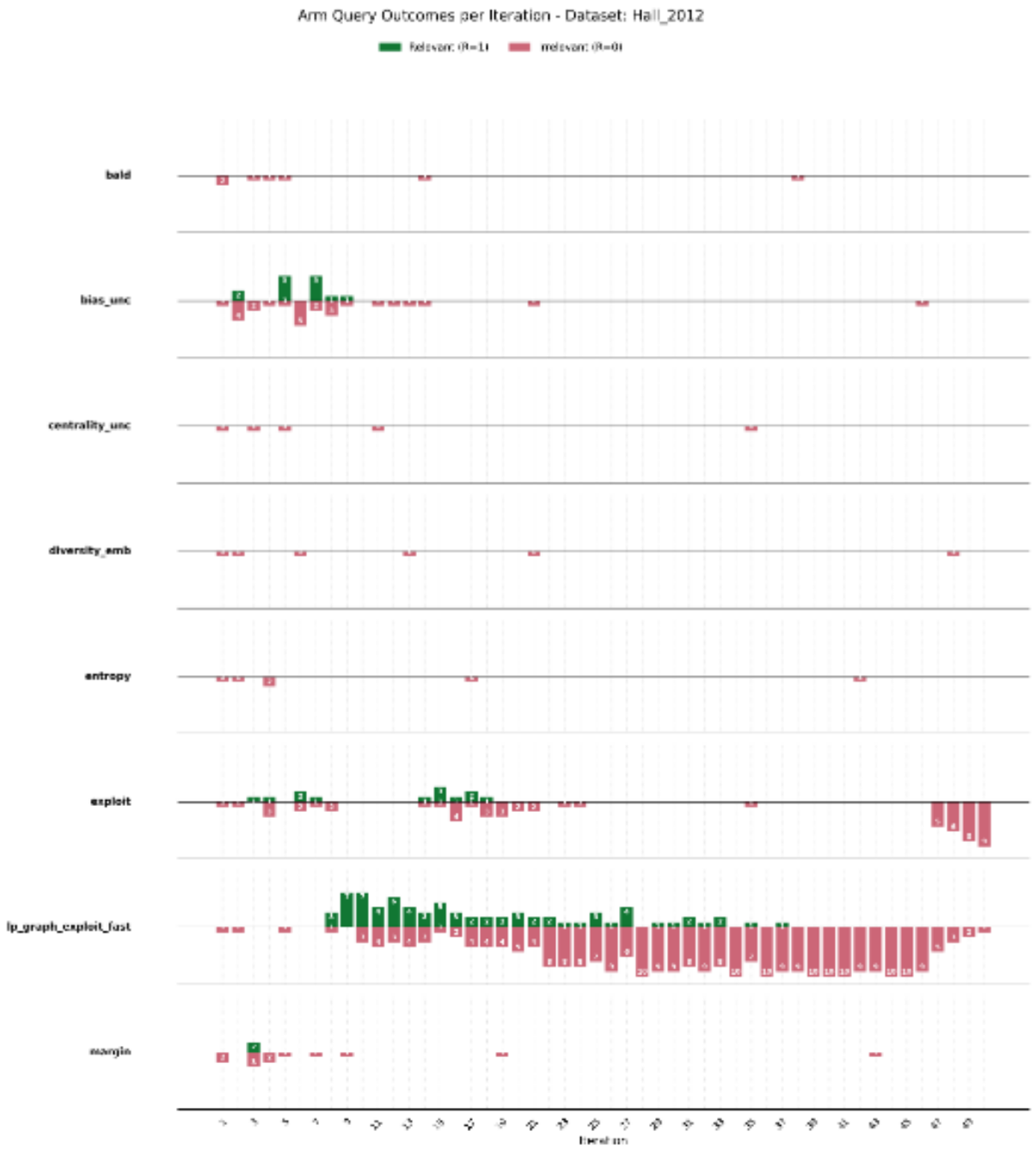

(d) Hall

Figure 30: Arm query outcomes for the first set of datasets with positive WSS@95 scores.

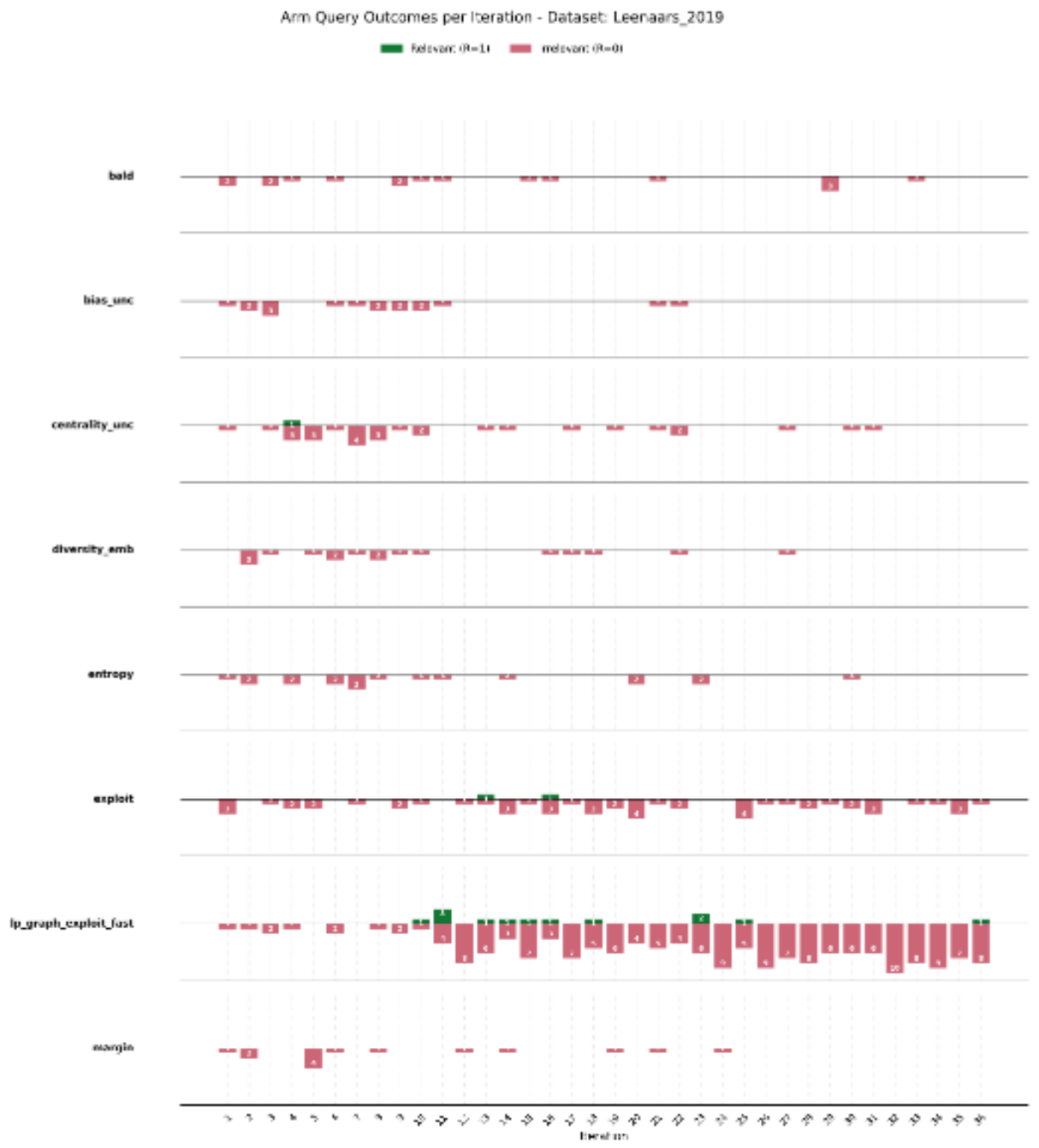

(a) Leenaars (2019)

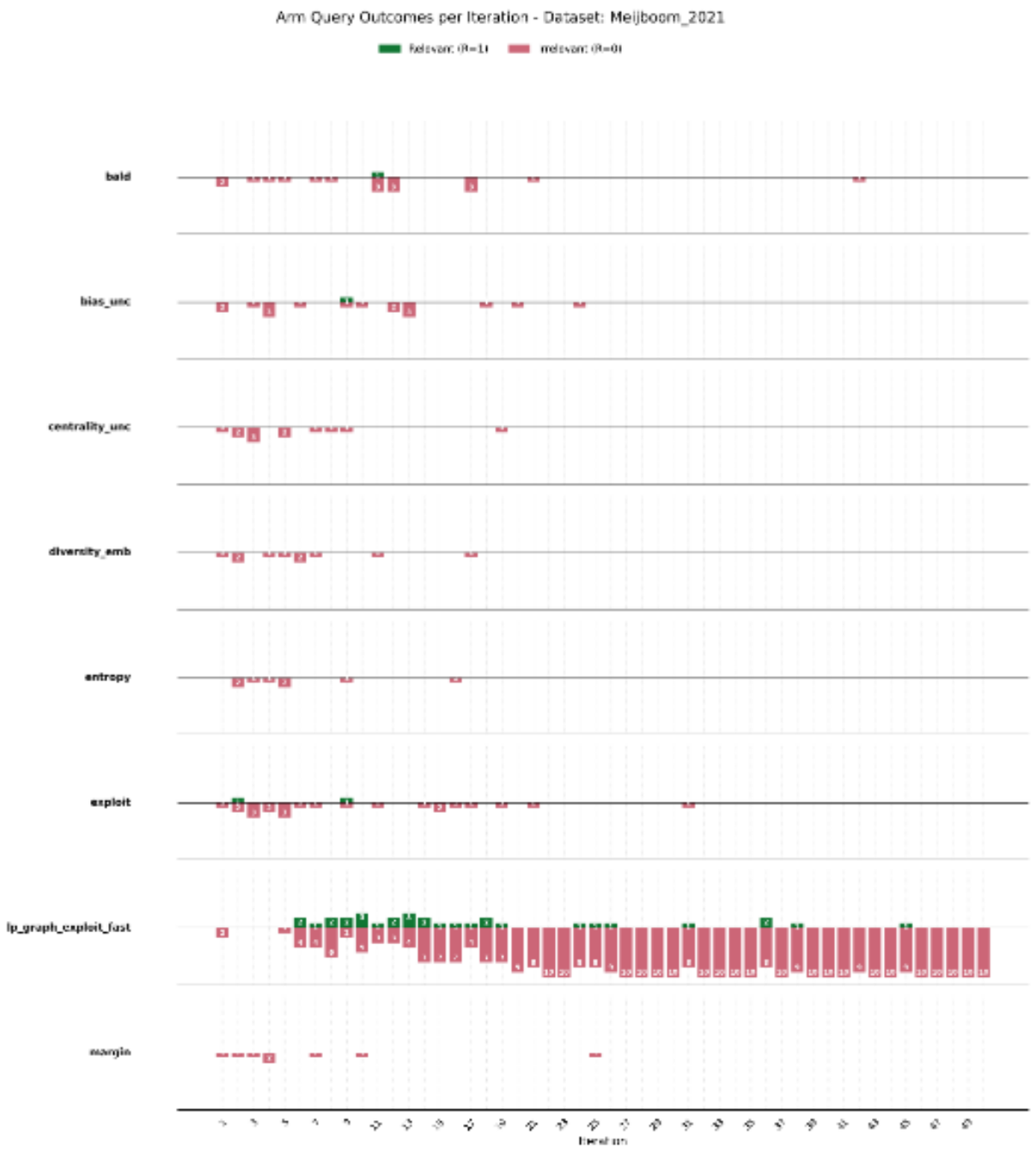

(b) Meijboom

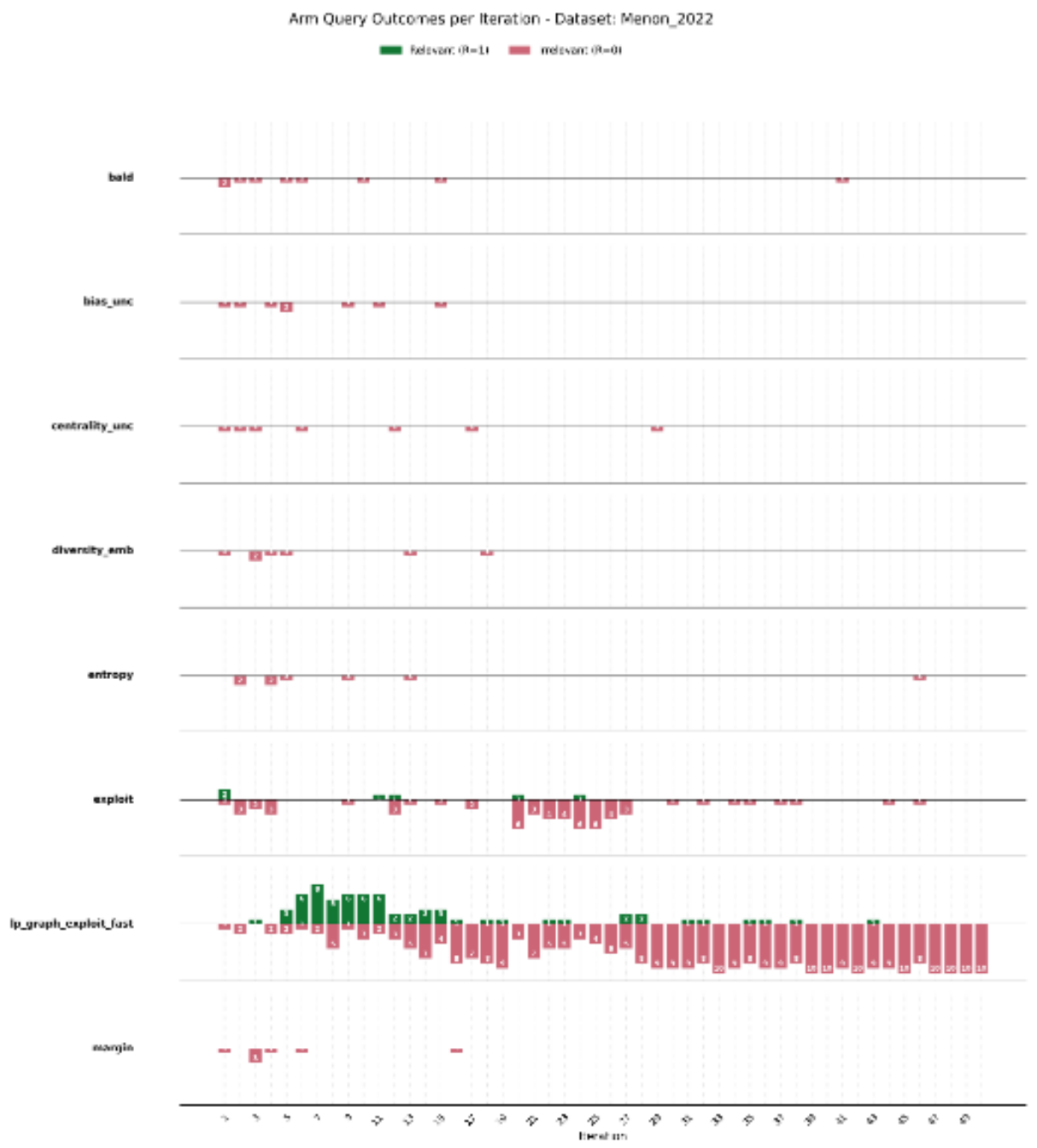

(c) Menon

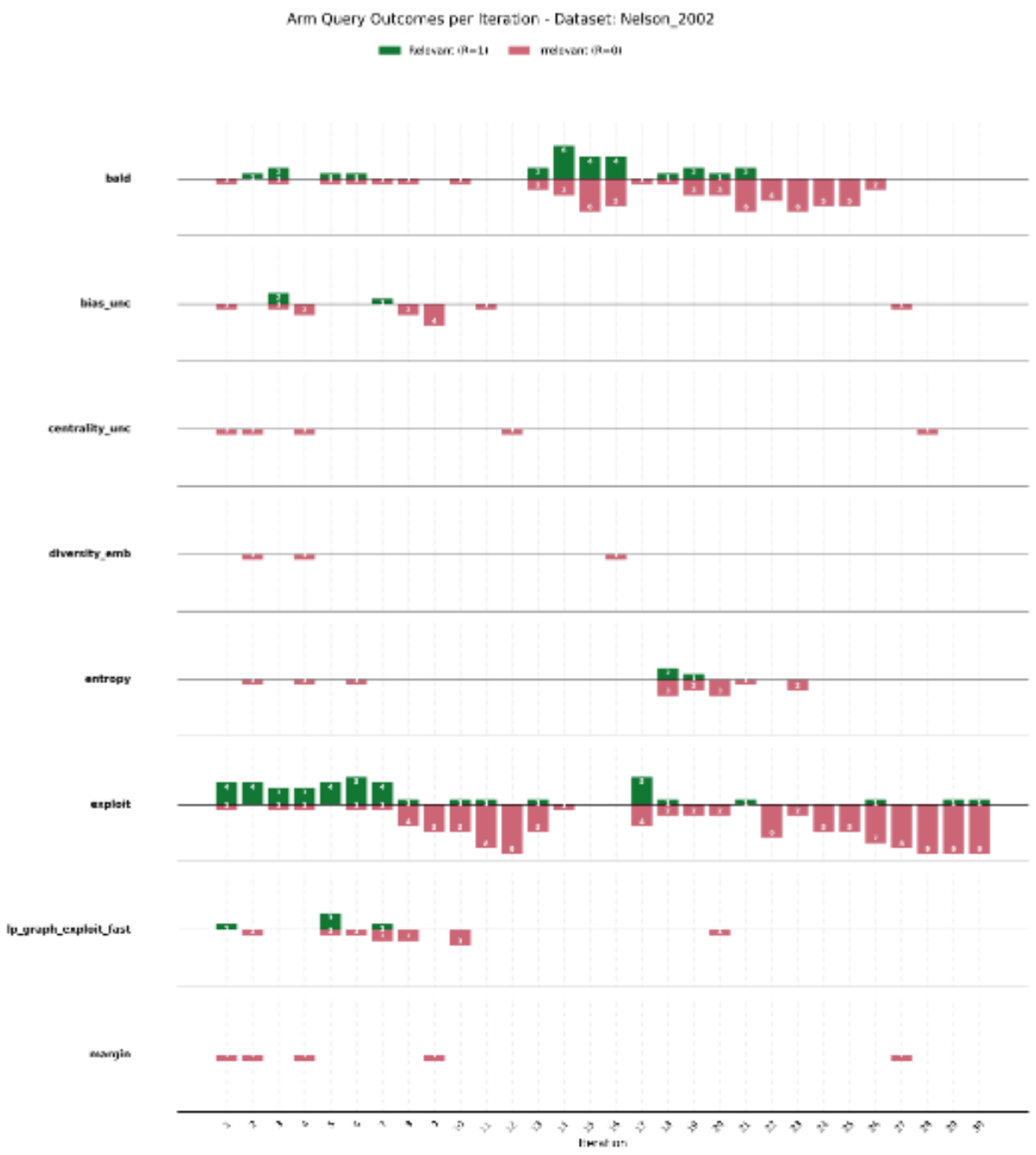

(d) Nelson

Figure 31: Arm query outcomes for the second set of datasets with positive WSS@95 scores.

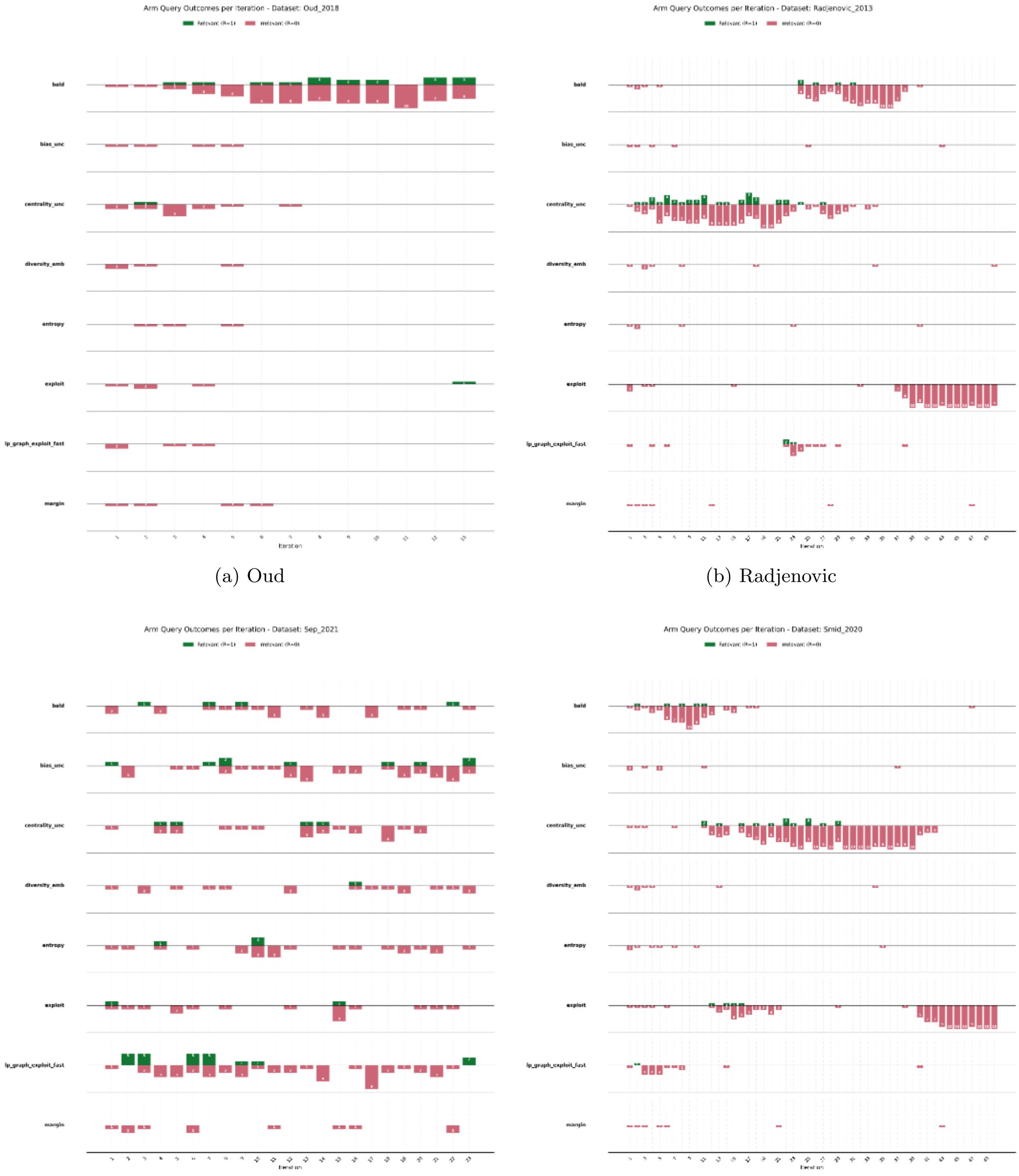

(a) Oud

(b) Radjenovic

(c) Sep

(d) Smid

Figure 32: Arm query outcomes for the third set of datasets with positive WSS@95 scores.

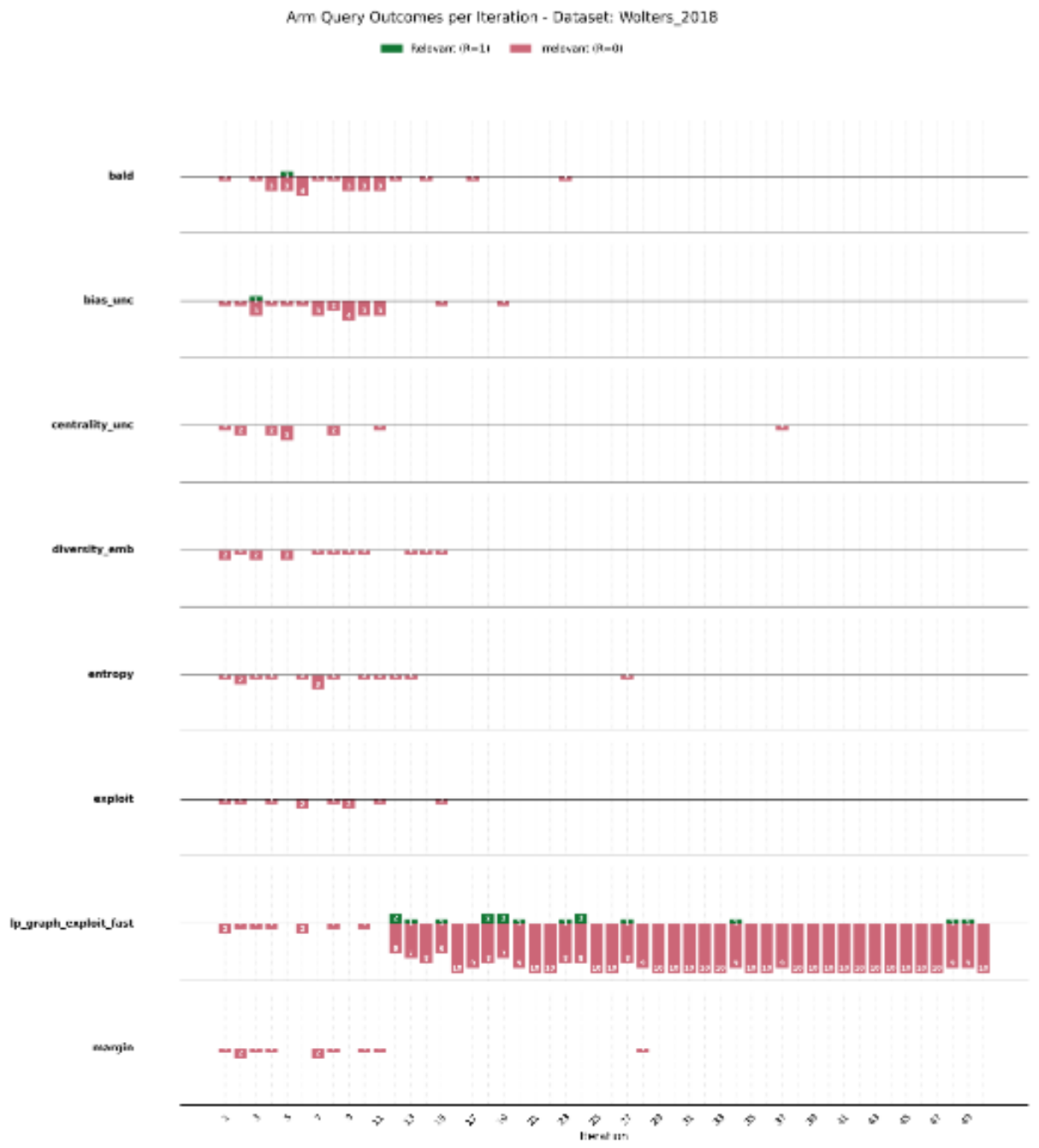

(a) Wolters

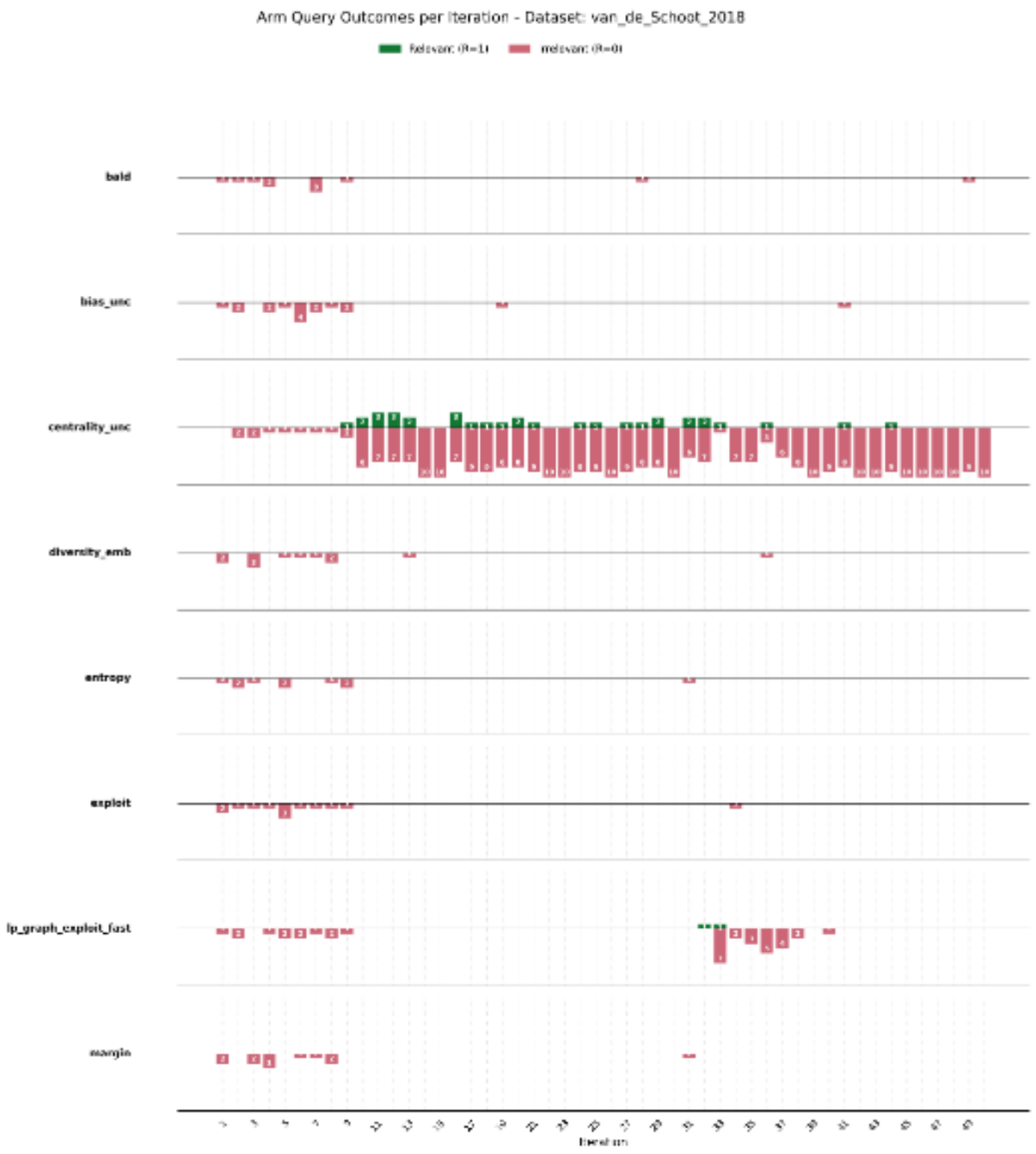

(b) van de Schoot

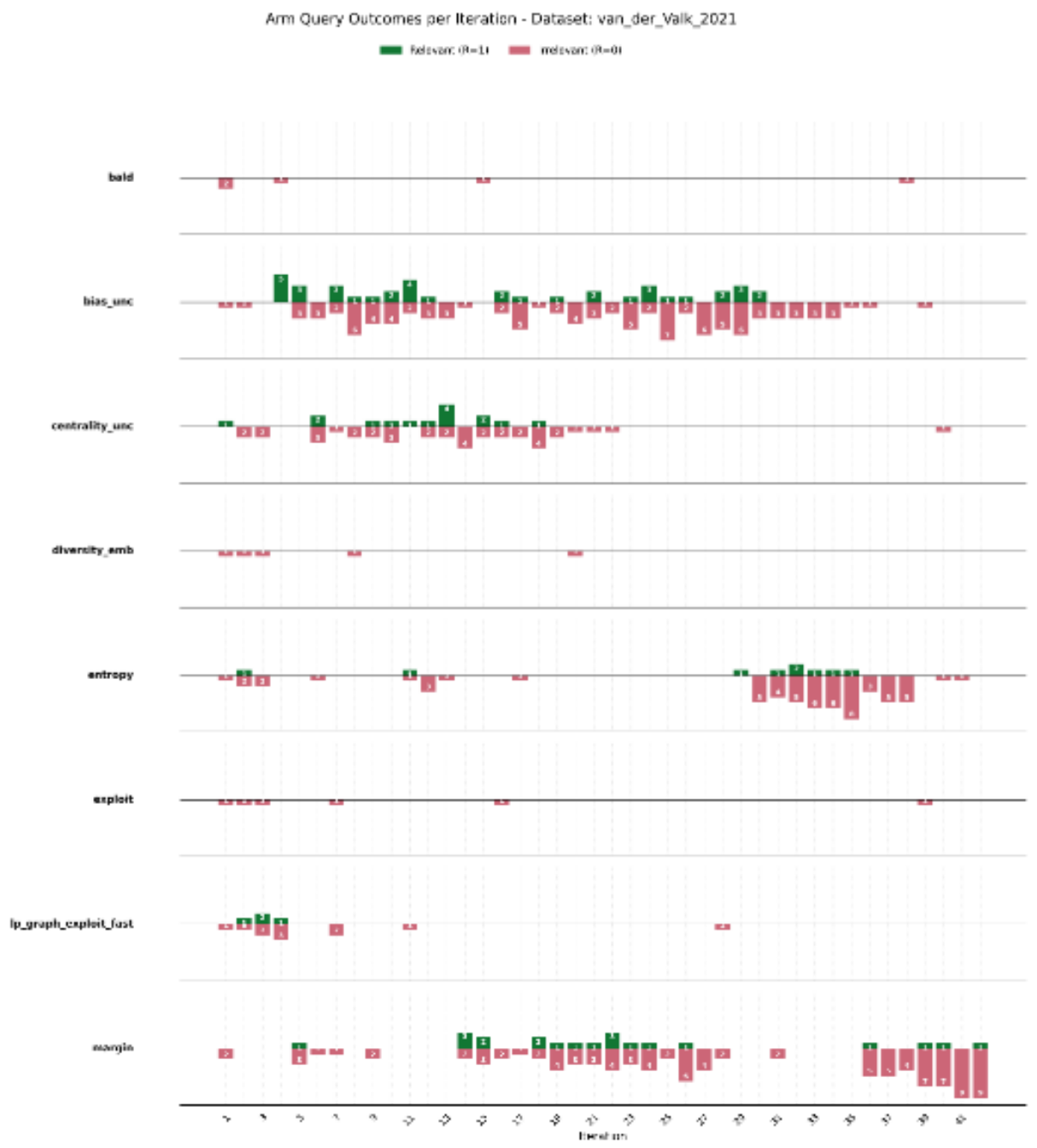

(c) van der Valk

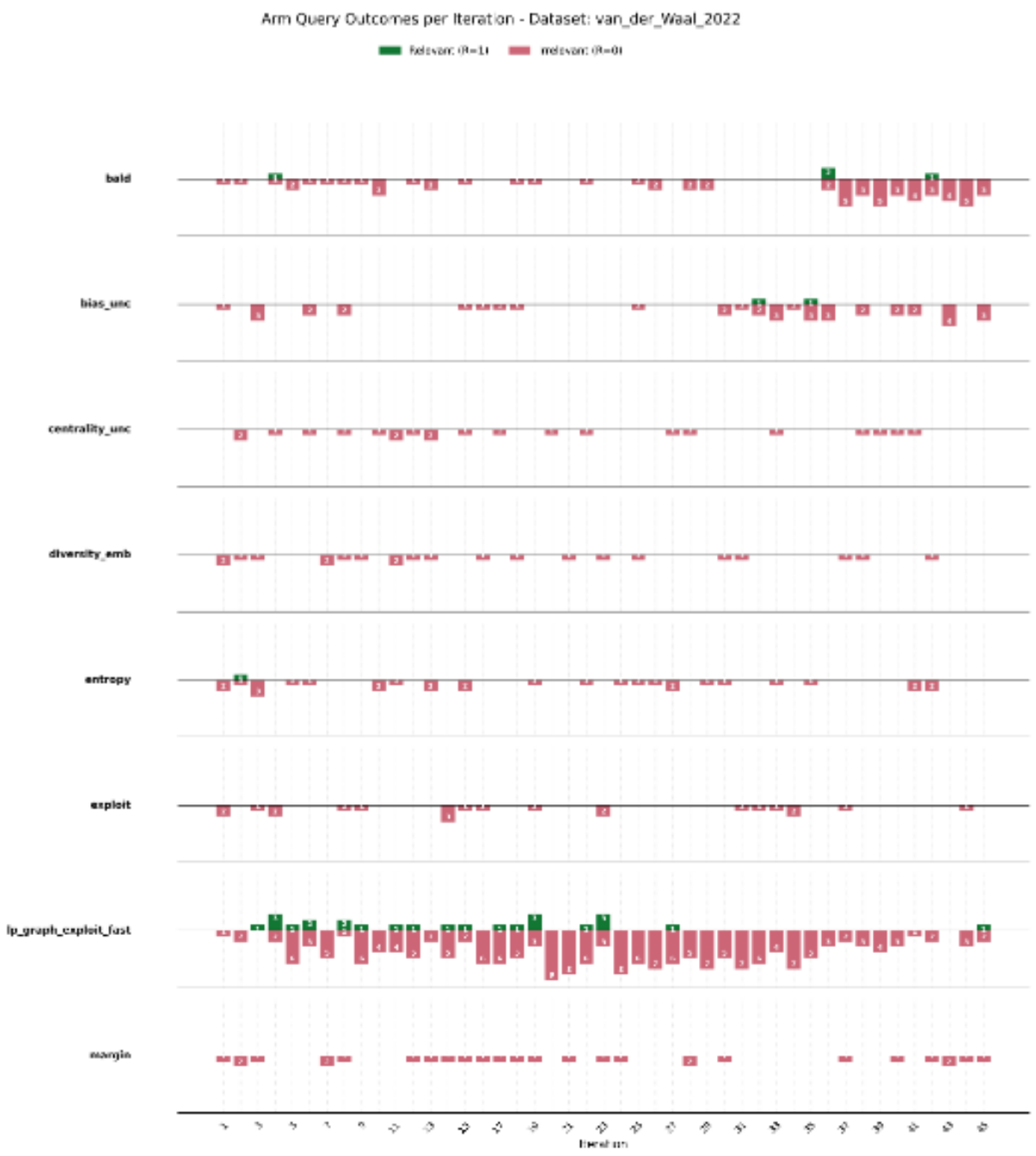

(d) van der Waal

Figure 33: Arm query outcomes for the final set of datasets with positive WSS@95 scores.

# F Diagram: Comprehensive Data preprocessing & parsing

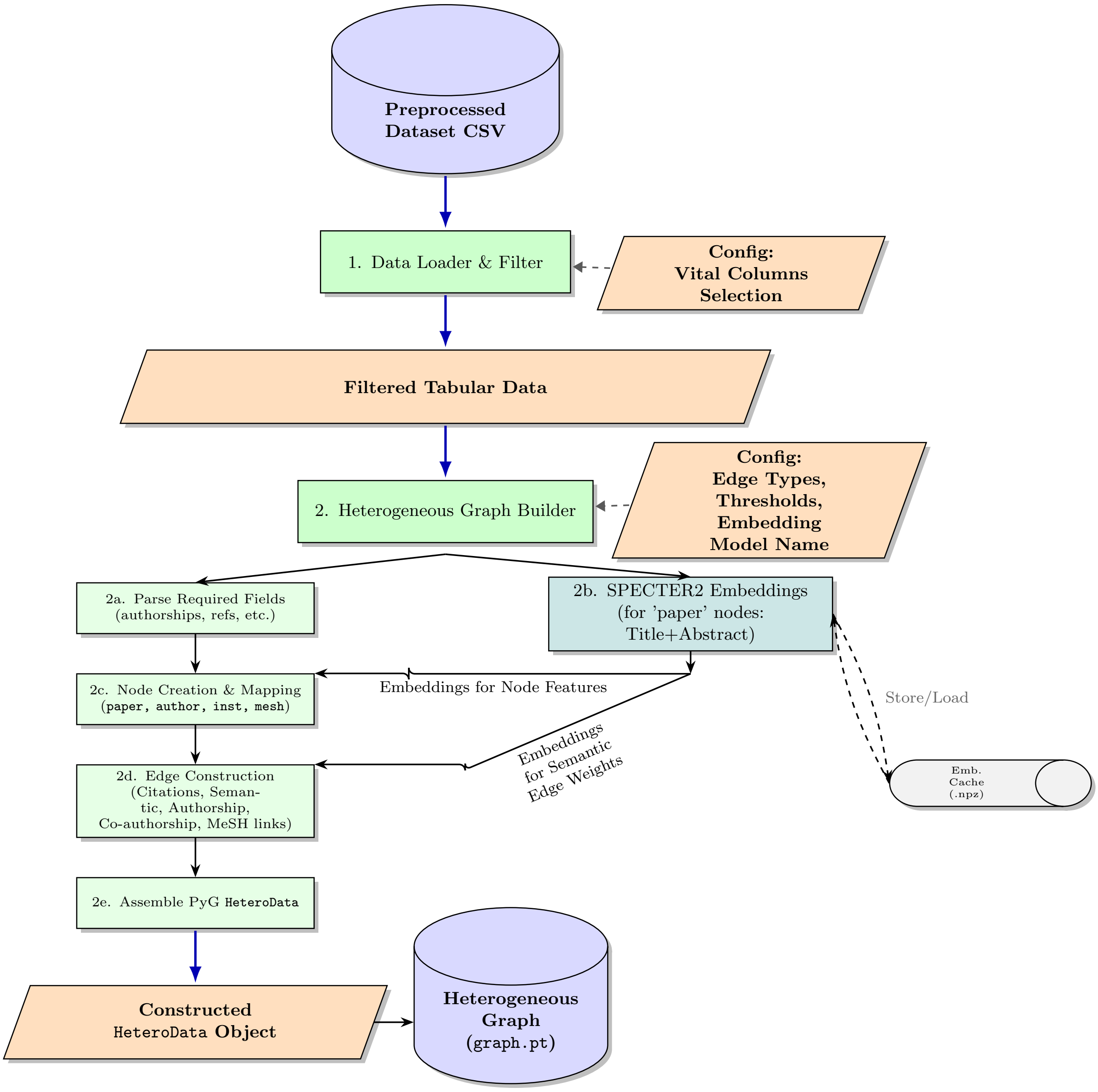

Figure 34: Data Ingestion and Heterogeneous Graph Construction Pipeline. Raw pre-normalized dataset CSVs are processed by the Data Loader to filter essential columns. The Heterogeneous Graph Builder then parses required fields, generates SPECTER2 embeddings for paper nodes (utilizing an optional cache), creates node types (paper, author, institution, MeSH), and constructs edges representing citations, semantic similarity, authorship, co-authorship, and MeSH term associations. These components are finally assembled into a PyTorch Geometric `HeteroData` object, which is serialized to a `.pt` file for downstream use in the active learning loop.

[33]

[33] Legend is on the next page

Table 14: Legend for Fig. 34: Diagram Elements in the Ingestion & Graph Construction Pipeline

| **Shape/Style** | **Description** |
|---|---|
| `Cylinder` (InputFile) | Raw dataset CSV file input (pre-normalized tabular data) |
| `Rounded Rectangle` (ProcessBlock) | Primary pipeline component (e.g., loader, graph builder) |
| `Small Rounded Rectangle` (SubProc) | Sub-component of the graph builder (e.g., node/edge parsing) |
| `Rectangle` (TransformerModel) | Embedding model (e.g., SPECTER2 for paper nodes) |
| `Rectangle` (DataStage) | Intermediate tabular data file (after filtering) |
| `Trapezoid` (OutputFile) | Graph output files (PyG HeteroData, `.pt`) |
| `Document` (Config) | YAML configuration file for parameterizing processing logic |
| `Cylinder` (CacheFile) | Optional cache file for storing/loading embeddings |
| `Solid Arrow` | Control or sequential operation flow |
| `Blue Arrow` | Data flow across pipeline stages |
| `Dashed Gray Arrow` | Configuration dependency |
| `Dashed Arrow with Loop` | Bidirectional cache I/O between model and storage |

# G Design rationale summary

Table 15: The Argumentative Funnel: Justifying the AutoDiscover Architecture by Systematically Eliminating Simpler Alternatives. Each challenge identified in a baseline approach directly motivates a core design principle of the final system.

| challenge | Question Investigated | Key Finding / Evidence | Conclusion & Architectural Implication |
|---|---|---|---|
| ***Part 1: Justifying the Core Feature — Pre-computed Semantic Embeddings*** | | | |
| **C1: Prohibitive LLM Cost** | Why not fine-tune a powerful LLM like SciBERT in each active learning SLR study? | A single retraining cycle takes nearly 3 hours on a high-performance GPU. | **Infeasible for an interactive system.** This establishes the need for a lightweight core model and motivates the use of pre-computed, static embeddings. |
| **C2: TF-IDF Semantic Failure** | If LLMs are too slow, why not use a fast, traditional model like TF-IDF? | Creates structurally disconnected graphs (only 44 edges) and relies on non-generalizable keywords (e.g., `microfracture`). | **Fails to capture semantic meaning.** This reinforces the decision to use powerful semantic embeddings like SPECTERv2, which provide semantic richness without the retraining cost. |
| ***Part 2: Justifying the Model Architecture — A Graph Neural Network*** | | | |
| **C3: Sparsity of Explicit Graphs** | Can we build a graph from explicit metadata like citations or concepts? | Explicit citation networks are extremely sparse (0.48% density). Concept graphs are denser but rely on non-standardized metadata. | **Explicit graphs are unreliable alone.** This motivates building the primary graph on the dense, *implicit* relationships within the SPECTERv2 embeddings. |
| **C4: Duality of Citation Patterns** | What does the citation structure tell us about the required strategy? | The network contains both highly interconnected communities of relevant papers and isolated "islands" reachable only via irrelevant "bridges." | **The data demands both exploitation and exploration.** This transitions the argument from choosing a model to defining how that model must behave. |
| **C5: Non-Linear Separability** | Are SPECTERv2 embeddings sufficient on their own with a simple classifier? | The UMAP visualization shows relevant papers are not in one clean cluster but in multiple, non-linearly separable groups. | **A simple linear classifier is insufficient.** This provides the ultimate justification for using a **Graph Neural Network (GNN)**, which excels at learning in such complex, neighborhood-based spaces. |
| ***Part 3: Justifying the Final Component — The Adaptive Agent*** | | | |
| **C6: Greedy GNN Cold-Start Failure** | Is a GNN enough on its own if we use a simple "greedy" strategy? | The ablation study proves a GNN-only system fails to discover any new relevant papers from a single starting seed. | **The GNN is powerful but needs intelligent guidance.** This proves that the query strategy is as critical as the model itself, especially in the cold-start phase. |
| **C7: Inadequacy of a Fixed Strategy** | Is one "smart" but static query strategy (e.g., uncertainty) good enough? | The comparison of discovery curves shows that no single strategy is consistently optimal; the best strategy changes over time. | **The optimal strategy is non-stationary.** The only remaining solution is a model guided by a **dynamically adaptive agent**. |